\documentclass[preprint,12pt,authoryear]{elsarticle}

\usepackage{amssymb}
\usepackage{lineno}

\usepackage{amsfonts}
\usepackage[toc,page]{appendix}
\usepackage{array}
\usepackage{bm}
\usepackage{booktabs}
\usepackage{color, colortbl}
\usepackage{enumitem}
\usepackage{float}
\usepackage[acronym]{glossaries}
\usepackage[hidelinks]{hyperref}
\usepackage{makecell}
\usepackage{microtype}
\usepackage{multirow}
\usepackage{subcaption}
\usepackage{tablefootnote}
\usepackage{tabularx}
\usepackage{xcolor}   
\usepackage{xspace}

\glsdisablehyper
\usepackage{amsmath}
\DeclareMathOperator*{\argmax}{arg\,max}

\newcolumntype{C}{>{\centering\arraybackslash}X}
\newcolumntype{L}{>{\raggedright\arraybackslash}X}

\newcommand{\tsac}{\ensuremath{\theta_{\text{SAC}}}\xspace}
\newcommand{\tppo}{\ensuremath{\theta_{\text{PPO}}}\space}
\newcommand{\buff}{\ensuremath{\mathcal{B}}\xspace}
\newcommand{\mem}{\ensuremath{\mathcal{M}}\xspace}
\newcommand{\ppo}{\gls{ppo}\xspace}
\newcommand{\ppos}{{\gls{ppo}'s}\xspace}
\newcommand{\sac}{\gls{sac}\xspace}
\newcommand{\sacs}{{\gls{sac}'s}\xspace}
\newcommand{\bul}{Broken, Unsevered Limb\xspace}
\newcommand{\bsl}{Broken, Severed Limb\xspace}
\newcommand{\hrom}{Hip ROM Restriction\xspace}
\newcommand{\arom}{Ankle ROM Restriction\xspace}

\newcommand{\sens}{Frozen Shoulder Lift Position Sensor\xspace}
\newcommand{\slip}{Slippery Elbow Flex Joint\xspace}
\newcommand{\rl}{\gls{rl}\xspace}
\newcommand{\ai}{\gls{ai}\xspace}

\newacronym{ai}{AI}{artificial intelligence}
\newacronym{crl}{CRL}{continual reinforcement learning}
\newacronym{gae}{GAE}{generalized advantage estimator}
\newacronym{grbal}{GrBAL}{Gradient-Based Adaptive Learner}
\newacronym{iot}{IoT}{Internet of Things}
\newacronym{kl}{KL}{Kullback–Leibler divergence}
\newacronym{lb}{LB}{lower bound}
\newacronym{maml}{MAML}{Model-Agnostic Meta-Learning}
\newacronym{mdp}{MDP}{Markov decision process}
\newacronym{ood}{OOD}{out-of-distribution}
\newacronym{pdvf}{PD-VF}{Policy-Dynamics Value Function}
\newacronym{ppo}{PPO}{Proximal Policy Optimization}
\newacronym{rebal}{ReBAL}{Recurrence-Based Adaptive Learner}
\newacronym{relu}{ReLU}{rectified linear unit}
\newacronym{rl}{RL}{reinforcement learning}
\newacronym{rom}{ROM}{range of motion}
\newacronym{sac}{SAC}{Soft Actor-Critic}
\newacronym{se}{SE}{standard error}
\newacronym{smp}{SMP}{Shared Modular Policy}
\newacronym{td}{TD}{temporal difference}
\newacronym{td3}{TD3}{Twin Delayed DDPG}
\newacronym{trpo}{TRPO}{Trust Region Policy Optimization}
\newacronym{ugvs}{UGVs}{unmanned ground vehicles}
\newacronym{ub}{UB}{upper bound}
\newacronym{xml}{XML}{Extensible Markup Language}

\makeatletter
\def\ps@pprintTitle{%
  \let\@oddhead\@empty
  \let\@evenhead\@empty
  \let\@oddfoot\@empty
  \let\@evenfoot\@empty
}
\makeatother

\journal{}

\begin{document}

\begin{frontmatter}


\title{Enhancing Hardware Fault Tolerance in Machines with Reinforcement Learning Policy Gradient Algorithms}

\author[a]{Sheila Schoepp}
\ead{sschoepp@ualberta.ca}
\author[a]{Mehran Taghian}
\ead{taghianj@ualberta.ca}
\author[c]{Shotaro Miwa}
\ead{Miwa.Shotaro@bc.mitsubishielectric.co.jp}
\author[d]{Yoshihiro Mitsuka}
\ead{Mitsuka.Yoshihiro@bp.mitsubishielectric.co.jp}
\author[a]{Shadan Golestan}
\ead{golestan@ualberta.ca}
\author[a,b]{Osmar R. Za\"{i}ane}
\ead{zaiane@ualberta.ca}

\affiliation[a]{
    organization={Department of Computing Science, University of Alberta},
    addressline={8900 114 St NW}, 
    city={Edmonton}, 
    state={Alberta},
    postcode={T6G 2S4},
    country={Canada}
}

\affiliation[b]{
    organization={Alberta Machine Intelligence Institute},
    addressline={10065 Jasper Ave \#1101}, 
    city={Edmonton},
    state={Alberta},
    postcode={T5J 3B1}, 
    country={Canada}
}

\affiliation[c]{
    organization={Advanced Technology R\&D Center, Mitsubishi Electric Corporation},
    addressline={8-1-1,Tsukaguchi-honmachi}, 
    city={Amagasaki-shi}, 
    state={Hyogo},
    postcode={661-8661},
    country={Japan}
}

\affiliation[d]{
    organization={Information Technology R\&D Center, Mitsubishi Electric Corporation},
    addressline={5-1-1, Ofuna}, 
    city={Kamakura-shi}, 
    state={Kanagawa},
    postcode={247-8501},
    country={Japan}
}

\begin{abstract}

Industry is moving toward autonomous, network-connected machines that detect and adapt to changing conditions, including hardware faults.
Conventional fault-tolerant design duplicates hardware and reroutes control logic; reinforcement learning (RL) offers a learning-based alternative.
This paper presents the first systematic comparison of two RL algorithms---Proximal Policy Optimization (PPO) and Soft Actor-Critic (SAC)---for integrating fault tolerance into control.
Beyond algorithm choice, we investigate four knowledge-transfer strategies: retaining or discarding model parameters, and retaining or discarding storage contents.
Performance is evaluated in two Gymnasium environments: Ant-v5 and FetchReachDense-v3.
Results show rapid, fault-specific recovery with clear trade-offs.
In Ant-v5, retaining PPO's parameters boosts early returns and remains the safest choice across all faults, while retaining SAC's parameters yields mixed outcomes.
SAC's early performance further depends on whether the replay buffer is retained: beneficial when prior experiences match current dynamics, but harmful when they diverge.
In FetchReachDense-v3, discarding both PPO's and SAC's parameters was most effective under sensor corruption.
Across tasks, both algorithms recover near-normal performance within minutes in low-dimensional settings and within days in high-dimensional settings, highlighting a clear trade-off between adaptation speed and asymptotic performance.
These findings demonstrate that RL can deliver robust fault tolerance and offer practical guidelines.

\end{abstract}


\end{frontmatter}

\section{Introduction}
\label{introduction}

Automation is revolutionizing industries, enhancing productivity and efficiency \citep{haleem2021hyperautomation,javaid2021substantial,vaidya2018industry}. 
This transformation is achieved by integrating machines equipped with sensors, such as robots, with \gls{ai}-powered analytical systems (agents), enabling real-time data collection and analysis. 
These agents make the critical decisions necessary for automation, including facilitating real-time fault detection and diagnosis of hardware faults in machines \citep{leite2024fault,neupane2025data,riazi2019detecting,zhang2017new}.
While fault detection and diagnosis are essential for identifying and understanding faults, they do not address adaptation to hardware faults.
To succeed in real-world deployments, it is crucial for machines to adapt to unexpected events such as hardware failures \citep{chen2023adapt,schrum2020your}.
As such, enhancing hardware fault tolerance in machines is imperative to ensure robust, autonomous operation in real-world applications.

An established method of improving hardware fault tolerance is through redundancy, where critical hardware components are duplicated to mitigate the risk of failure \citep{guiochet2017safety,qi2024fault,urrea2025hybrid}. 
However, this approach has significant drawbacks, including increased machine size, weight, power consumption, and financial costs \citep{dubrova2013fault}. 
Moreover, retrofitting existing machines with redundant components is often impossible. 
Therefore, pursuing alternative approaches that do not rely on redundancy offers significant advantages for enhancing hardware fault tolerance in machines.

Taking inspiration from nature, one can imagine an agent adapting a machine's behaviour in response to a hardware fault. 
Animals, for example, demonstrate an extraordinary ability to adapt; they can modify their gait in the presence of an injured limb, using non-injured limbs in a compensatory manner \citep{fuchs2014ground,hooper2024bio,hu2025egocentric,jarvis2013kinematic}. 
An established approach to incorporate such adaptable, compensatory behaviours into machines is through algorithmic reconfiguration. 
In algorithmic reconfiguration, an agent adjusts the underlying algorithms that govern hardware usage within a machine, enabling adaptation to changes in hardware conditions \citep{guiochet2017safety,qi2024fault,urrea2025hybrid}.
Algorithmic reconfiguration may involve altering hyperparameter settings, adjusting the model architecture, or even switching to an entirely different algorithm.

In essence, algorithmic reconfiguration is closely linked to continual learning; when a machine faces evolving hardware conditions throughout its lifetime, it must continuously adapt to succeed under changed conditions.
Adaptation in continual learning involves leveraging the continuous flow of data to modify aspects of the algorithm, such as the model parameters and/or storage contents.
To accelerate this adaptation process, previously acquired knowledge can be strategically transferred to new tasks \citep{chen2018lifelong}.

In nature, adaptation to physical changes is often achieved through trial-and-error, aided by knowledge from past experiences. 
This trial-and-error approach is mirrored in continual \gls{rl} \citep{abel2023definition,khetarpal2022towards,sutton2018reinforcement,xu2018reinforced}, making continual \gls{rl} an ideal strategy for an agent tasked with adapting a machine's behaviour to unexpected events such as hardware faults.
Continual \gls{rl} entails an agent continuously interacting with a non-stationary environment, learning from feedback (trial-and-error), and strategically transferring acquired knowledge.

\begin{figure}[t]
    \centering
    \captionsetup{justification=justified}
    \includegraphics[width=0.96\linewidth]{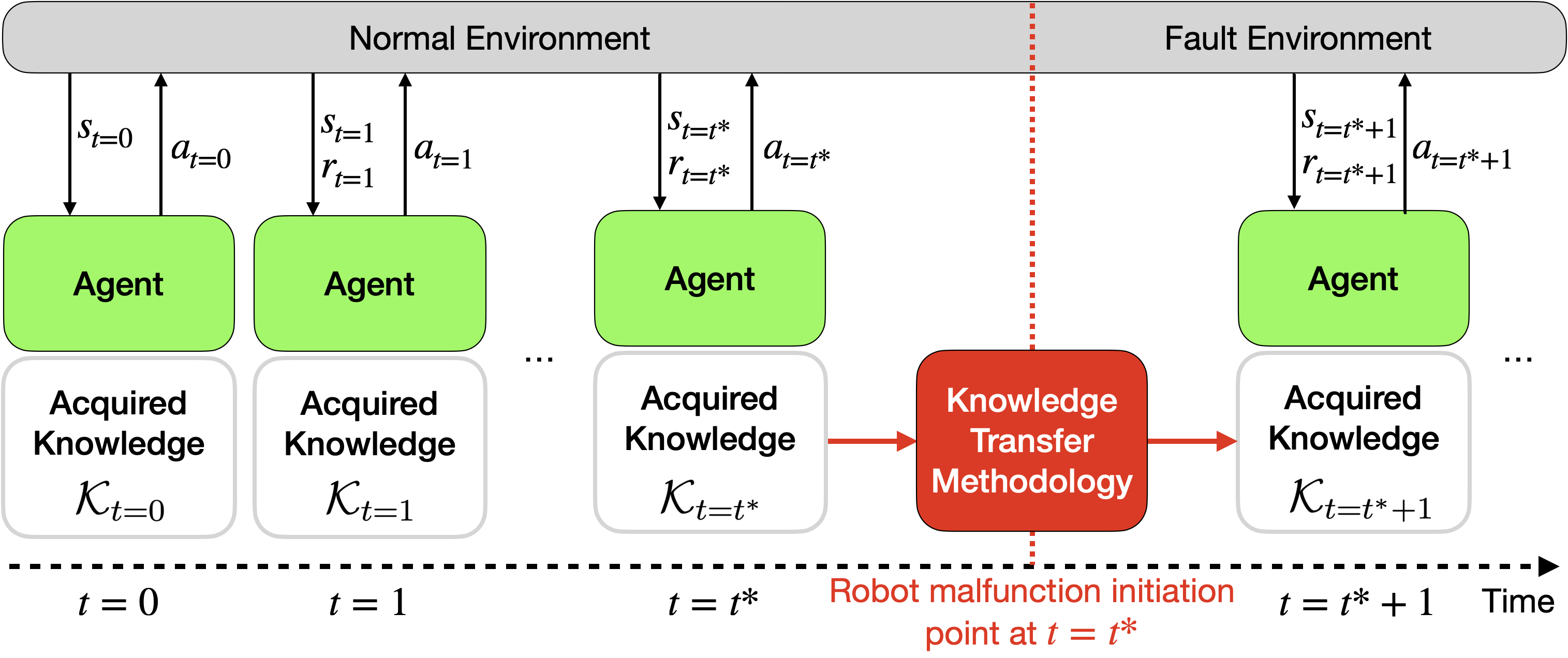}
    \caption{Overview of the proposed methodology. At $t=t^*$, the robot encounters a fault. Knowledge acquired by the agent in the normal environment up to $t=t^*$ (${\mathcal{K}}_{t=t^*}$) is transferred to construct a prior for subsequent learning in the fault environment.}
    \label{fig:study_overview}
\end{figure}

Despite its potential, continual \gls{rl} remains relatively unexplored as a means of improving hardware fault tolerance in machines.
This study investigates how continual \gls{rl} can enhance hardware fault tolerance in simulated machines.
Figure~\ref{fig:study_overview} illustrates the approach: a machine (robot) encounters a fault at time $t=t^*$; knowledge acquired in the normal environment up to $t=t^*$ ($K_{t=t^*}$) is leveraged through different transfer strategies.
Specifically, the analysis considers the impact of retaining or discarding components of the agent's acquired knowledge, including both the learned model parameters and the agent's stored experiences.
The role of retaining and fine-tuning model parameters learned in a normal environment, versus discarding them, in facilitating adaptation is evaluated.
The model parameters, if retained, initially serve as a prior in the fault environment.
In addition, the effect of retaining or discarding the storage contents  (i.e., stored experiences) is explored. If retained, the storage contents initially serve as a proxy for a model of the normal environment.

The hypothesis is that ${\mathcal{K}}_{t{=}t^*}$ contains useful information that can accelerate adaptation to hardware faults. 
To test this hypothesis, the effectiveness of two state-of-the-art \gls{rl} algorithms, \gls{ppo} and \gls{sac}, in improving hardware fault tolerance is empirically evaluated.
For the evaluation, widely accepted benchmarks from the Gymnasium toolkit \citep{towers2024gymnasium}, namely Ant-v5 and FetchReachDense-v3, are used.
To the best of the authors' knowledge, this study is the first to comprehensively examine the performance of \gls{ppo} and \gls{sac} in the context of hardware fault tolerance in machines.
The findings provide valuable insights into optimizing knowledge transfer in a continual \gls{rl} fault adaptation setting, thereby accelerating machine adaptation. While the evaluation is conducted in simulation, the use of standard Gymnasium benchmarks enables controlled fault injection and reproducible comparison across diverse fault conditions, providing a practical proxy for real-world fault adaptation scenarios.

{
\setlist[enumerate]{leftmargin=*,label=\arabic*.,labelsep=0.5em,itemsep=0.25\baselineskip,topsep=0.25\baselineskip}

\paragraph{Contributions} This work makes the following contributions:
\begin{enumerate}
    \item \textit{Problem framing:} \gls{crl} is introduced as a software-based approach to enhancing hardware fault tolerance in machines, providing an alternative to traditional redundancy-based design.
    \item \textit{Comparative analysis:} Two state-of-the-art policy gradient algorithms, \glsentryfull{ppo} and \glsentryfull{sac}, are systematically evaluated under diverse simulated hardware faults.
    \item \textit{Knowledge transfer strategies:} An ablation study investigates four strategies for transferring knowledge at the onset of faults, focusing on retaining or discarding model parameters and storage contents (i.e., stored experiences).
    \item \textit{Empirical insights:} Experimental results across multiple fault environments reveal practical guidelines for adaptation---for example, retaining \gls{ppo} parameters is generally advantageous, while \gls{sac} benefits from selective reinitialization depending on fault severity.
\end{enumerate}
}

\section{Background and Related Work}
\label{sec:background_and_related_work}

This section establishes the research context by examining relevant literature.
Section \ref{sec:reinforcement_learning} outlines key concepts and techniques in \gls{rl}, while Section \ref{sec:related_works} reviews notable contributions to the research landscape.
Together, these discussions highlight the unique contributions of this study within the field.

\begin{table}[!ht]
    \centering
    \caption{Summary of notation.}
    \label{tab:notation}
    \begin{tabular}{ll}
    \toprule
    Symbol & Description \\
    \midrule
    $\mathcal{S}$ & set of states \\
    $\mathcal{A}$ & set of actions \\
    $\mathcal{R}$ & set of possible rewards \\
    $\mathbb{R}$ & set of real numbers \\
    $p(s', r \mid s, a)$ & transition probability function \\
    $t$ & discrete time step \\
    $S_t$ & state at time $t$ \\
    $A_t$ & action at time $t$ \\
    $R_{t}$ & reward at time $t$ \\
    $\tau$ & trajectory (sequence of states, actions, and rewards) \\
    $G_t$ & return following time $t$ \\
    $\gamma$ & discount factor \\
    \(\pi\) & (stochastic) policy; maps states to (distributions over) actions \\
    $\pi(a \mid s)$ & probability of selecting action \(a\) in state \(s\) under policy \(\pi\) \\
    $J(\pi)$ & performance objective for policy $\pi$ \\
    $\mathbb{E}_{\pi}[\cdot]$ & expectation under trajectories generated by $\pi$ \\
    $\pi^*$ & optimal policy \\
    $\theta$ & parameters \\
    $\pi_{\theta}$ & parameterized policy \\
    $\pi_{\theta}(a \mid s)$ & probability of selecting action $a$ in state $s$ under policy $\pi_{\theta}$ \\
    $J(\pi_{\theta})$ & performance objective for policy $\pi_{\theta}$ \\
    $\theta^*$ & optimal parameters \\
    \midrule
    $\mathcal{M}$ & \ppo memory \\
    $L^{\text{CLIP}}(\theta)$ & \ppo clipped surrogate objective \\
    $\pi_{\theta_{\text{old}}}$ & parameterized policy used to collect trajectories with \ppo \\
    $r_t(\theta)$ & \ppo importance ratio \\
    $\theta_{\text{PPO}}$ & \ppo solution at convergence in the normal environment \\
    \midrule
    $\mathcal{B}$ & \sac replay buffer \\
    $\alpha$ & temperature coefficient scaling the SAC entropy term \\
    $\theta_{\text{SAC}}$ & \sac solution at convergence in the normal environment \\
    \bottomrule
    \end{tabular}
\end{table}

\subsection{Reinforcement Learning}
\label{sec:reinforcement_learning}

\gls{rl} is a subfield of machine learning that enables agents to autonomously learn by making sequential decisions through interactions with their environment. 
Sequential decision-making problems are typically modelled as a \gls{mdp}. 
An \gls{mdp} is defined by the tuple \((\mathcal{S}, \mathcal{A}, \mathcal{R}, p)\), where \(\mathcal{S}\) denotes the set of states; \(\mathcal{A}\) denotes the set of actions; \(\mathcal{R} \subset \mathbb{R}\) is the set of possible rewards; \(p(s', r | s, a)\) represents the \gls{mdp} dynamics, which is the probability of transitioning to state \(s'\) and receiving reward \(r\) after taking action \(a\) in state \(s\) \citep{sutton2018reinforcement}.

Interaction between the agent and the environment occurs over a series of discrete time steps, represented by \(t = 0, 1, 2, \dots \).
During each interaction, the environment provides the agent with information on its current state \(S_t \in \mathcal{S}\). 
The agent selects and executes an action \(A_t \in \mathcal{A}\), guided by its policy \(\pi\), where \(\pi(a \mid s) \doteq \Pr\{A_t = a \mid S_t = s\}\).
Subsequently, the agent receives feedback in the form of a numerical reward \(R_{t+1} \in \mathcal{R}\), and the environment updates the agent with its new state \(S_{t+1} \in \mathcal{S}\). 
The sequence of states, actions, and rewards forms a trajectory, denoted by \(\tau\). 
The primary objective of the agent is to maximize the expected discounted return.
For a policy \(\pi\), the return following time \(t\) is defined as \(G_t \doteq \sum_{k=0}^{\infty} \gamma^{k} R_{t+k+1}\), where \(\gamma \in [0,1]\) is the discount factor, and the performance of \(\pi\) is \(J(\pi) \doteq \mathbb{E}_\pi[G_0]\), with \(\mathbb{E}_\pi[\cdot]\) denoting the expectation under trajectories generated by \(\pi\).

This objective guides the agent's learning, as it continually seeks to select actions that result in higher rewards. 
The optimization problem that aims to maximize the objective function \(J(\pi)\) can be expressed as
\begin{equation} \label{eq_policy}
    \pi^* = \argmax_{\pi} J(\pi),
\end{equation}
where \(\pi^*\) is the optimal policy.

\subsubsection{Policy Gradient Methods}

Policy gradient methods optimize the policy directly, as described by Equation~\ref{eq_policy}. 
The policy is represented as a parameterized function \(\pi_{\theta}(a \mid s)\), where \(\theta \in \mathbb{R}^{d}\) is a \(d\)-dimensional vector. 
The policy is represented as a parameterized function $\pi_{\theta}(a \mid s)$, 
where $\theta$ denotes the policy parameters.
For a parameterized policy, we write the objective function as \(J(\pi_{\theta})\). 
In policy gradient methods, the parameters \(\theta\) are evaluated using this performance measure and are iteratively updated using gradient ascent, improving the policy by identifying parameter values that yield higher expected returns.

\subsubsection{Proximal Policy Optimization (PPO)}

\gls{ppo} is an on-policy algorithm designed to limit the magnitude of policy updates, thus ensuring stable and gradual changes during training \citep{schulman2017proximal}. 
\gls{ppo} has proven to be an effective algorithm for robotic applications \citep{kuo2023two}. 

A \gls{ppo} agent collects experiences---namely, a collection of trajectories---through interactions with the environment while following the current policy. 
After collecting a batch of trajectories, these experiences are stored in the agent's memory \(\mathcal{M}\) and used to compute the clipped surrogate objective \(L^{\text{CLIP}}(\theta)\), which approximates policy improvement and guides the policy update from the current policy \(\pi_{\theta_{\text{old}}}\), whose parameters 
\(\theta_{\text{old}}\) were used to collect the trajectories, to an updated policy \(\pi_{\theta}\). 
To prevent drastic policy changes that could destabilize learning, \(L^{\text{CLIP}}(\theta)\) applies a clipping operation to the probability ratio \(r_t(\theta) = \pi_\theta(a_t \mid s_t) / \pi_{\theta_{\text{old}}}(a_t \mid s_t)\), constraining the updated policy to remain close to the current policy. 
The parameters \(\theta\) are optimized over one or more epochs, using minibatches of experiences sampled from \(\mathcal{M}\), according to
\begin{equation} \label{eq_theta}
    \theta^* = \argmax_{\theta} L^{\text{CLIP}}(\theta).
\end{equation}

\noindent After completing the update, the experiences stored in \(\mathcal{M}\) are discarded. 
The algorithm then collects new experiences using the updated policy and stores them in \(\mathcal{M}\). 
This iterative process enables \gls{ppo} to refine the policy gradually and converge towards an optimal or near-optimal solution. The solution at convergence in the normal environment is denoted as \(\theta_{\text{PPO}}\). 
For further details on \gls{ppo}, see \citet{schulman2017proximal}.

\subsubsection{Soft Actor-Critic (SAC)}

\gls{sac} is an off-policy algorithm that is well-suited for real-world problems, owing to its low sample complexity and robustness to various hyperparameter settings \citep{haarnoja2018softa,haarnoja2018softb}.
A distinguishing feature of \gls{sac} is its use of an entropy regularization term in its objective function \(J(\pi_{\theta})\).
The entropy regularization term encourages exploration by favouring stochastic policies.
The entropy term is scaled by a temperature coefficient $\alpha > 0$, controlling the balance between policy stochasticity and expected returns. 
\gls{sac} incorporates an automatic adjustment mechanism for this coefficient, achieved by imposing a constraint that sets a lower bound on the average policy entropy during the optimization process.
This mechanism enables \gls{sac} to strike a delicate balance between exploration and exploitation, allowing the agent to learn a policy that effectively adapts to diverse environmental conditions while maximizing expected returns.

To facilitate the learning process, \gls{sac} uses a replay buffer, denoted as $\mathcal{B}$, to retain past experiences. 
Once $\mathcal{B}$ reaches a specified minimum capacity, batches of experiences---specifically, past agent-environment interactions---are uniformly sampled from $\mathcal{B}$ at each time step and used to update the parameters. 
The parameters are optimized using:

\begin{equation} \label{eq_sac_policy}
     \theta^* = \argmax_{\theta} J(\pi_{\theta})
\end{equation}

\noindent Following each update, a new experience is generated under the updated policy and added to the replay buffer $\mathcal{B}$, replacing the oldest stored experience. 

It is worth noting that each experience stored in the replay buffer is generated under a different policy. 
The ability to reuse past experiences, even when generated under different policies, is what enables \gls{sac} to be highly sample-efficient, requiring fewer interactions with the environment compared to on-policy algorithms, such as \gls{ppo}.

The \gls{sac} solution at convergence in the normal environment is denoted as $\theta_{\text{SAC}}$.
For further details on \gls{sac}, see \citet{haarnoja2018softa}.

\subsection{Related Work}
\label{sec:related_works}

\subsubsection{Policy-Dynamics Embedding for Adaptation}
The algorithm developed by \cite{raileanu2020fast}, known as \gls{pdvf}, demonstrated rapid adaptability to previously unseen environments with changed dynamics. The methodology involved training an ensemble of policies across a set of environments with different dynamics, then learning latent embeddings of both policies and dynamics via self-supervised encoder-decoder networks. A value function over the joint policy-dynamics embedding space was trained in a supervised fashion. At test time, when faced with an unseen environment, a small number of transitions were used to infer the dynamics embedding, and the policy embedding that maximized the learned value function was computed in closed form via singular value decomposition; the resulting embedding was then decoded into actions. To evaluate the effectiveness of their approach, \cite{raileanu2020fast} applied \gls{pdvf} to four \textit{Ant-Legs} environments, where modifications were made to the length of the Ant-v5 robot model's linkages.

A limitation of this work is that, unlike online learning algorithms such as \gls{sac} and \gls{ppo}, \gls{pdvf} does not learn or update a policy in real time; rather, it computes a policy embedding from a learned value function without any parameter updates in the test environment.
This approach deprives the resulting policy of further improvement through continued interaction with the environment.
In contrast, \gls{sac} and \gls{ppo} are online learning algorithms capable of adapting a transferred policy. 
Furthermore, a competing algorithm, \gls{ppo}all, was trained in a multi-task setting, where negative transfer can hinder performance if the training tasks are too dissimilar \citep{rosenstein2005transfer}, rendering this method of limited practical value in the fault adaptation domain, where fault tasks may differ drastically.

In this study, \gls{ppo} is trained on a single task, demonstrating that the learned, high-performing policy is both transferable and quickly adaptable to unseen tasks with similar structure, and it outperforms the reported performance of \gls{pdvf} on comparable tasks.

\subsubsection{Meta-Reinforcement Learning} 

\cite{nagabandi2018learning} introduced two variants of meta-\gls{rl} algorithms, namely a \gls{grbal} and a \gls{rebal}, which demonstrated superior speed and sample efficiency in adapting to changing dynamics, including hardware faults, novel terrains, and unexpected perturbations. These two algorithms learn a model of the environment dynamics, enabling rapid adaptation with as little as 1.5--3 hours of real-world experience. The evaluation of the algorithms was conducted across several simulated MuJoCo tasks, including a quadrupedal Ant model with a single crippled leg.

A drawback of the algorithms proposed by \cite{nagabandi2018learning} is that machine faults can arise unexpectedly, exhibiting diverse and unpredictable characteristics. Meta-\gls{rl} algorithms, such as \gls{grbal} and \gls{rebal}, require meta-training over a distribution of tasks, which, although shown to generalize to some out-of-distribution tasks, may still limit their practicality when the test-time task differs substantially from the training distribution. In contrast, \gls{ppo} and \gls{sac} have the advantage of adapting without requiring a pre-specified task distribution. This flexibility enables them to effectively handle a wide range of faults without the need for specialized pretraining. Furthermore, \gls{ppo} and \gls{sac} are shown in our experiments to outperform \gls{grbal} and \gls{rebal} on comparable tasks, consistent with the tendency of model-based approaches to achieve suboptimal asymptotic performance compared to model-free approaches---a limitation acknowledged by \cite{nagabandi2018learning} themselves.

\subsubsection{Preprocessing in Simulation} 

\cite{cully2015robots} used a robot simulator to construct a low-dimensional behaviour-performance map. 
In the event of a fault, a real-world robot would take a trial-and-error approach to fault adaptation, iteratively selecting and executing the projected best-performing behaviour from the map, and subsequently updating the map with the real-world performance. 
This iterative procedure would continue until a high-performing behaviour, surpassing a predefined threshold, was identified. Notably, this approach demonstrated significant success, achieving recovery within minutes in a low-dimensional parameter search space.

This approach, however, had certain limitations. 
First, it relied on the availability of a simulator to generate a behaviour-performance map, which limited its practicality for robots lacking simulators.
Second, it faced challenges when confronted with a high-dimensional parameter search space. 
This study addresses these limitations by offering broad applicability to a wide range of robots without the need for prior computation in simulation. 
In addition, the effectiveness of adaptation using the \gls{ppo} and \gls{sac} algorithms is demonstrated within high-dimensional parameter spaces, utilizing raw sensor output data to select continuous actions.

\subsubsection{A Policy for Each Actuator}

\cite{huang2020one} introduced a decentralized approach to learning called \gls{smp}, where each component of a robot, e.g., an actuator, was treated as an individual module. Each module had its own policy, which was instantiated from a global policy (i.e., a single, reusable neural network). The input to each policy included raw data from the module's local sensors. A system of message passing allowed for communication between modules. \gls{smp} was trained in a multi-task learning setting, where each task involved a robot model with a unique arrangement of linkages and/or joints.

\cite{huang2020one} evaluated \gls{smp} in the OpenAI Gym environments Walker2D-v2, Humanoid-v2, Hopper-v2, and HalfCheetah-v2 \cite{brockman2016openai}. Although \gls{smp} was shown to perform well on simple tasks, it faced difficulties on more complex tasks (e.g., Humanoid-v2). The performance of \gls{smp} was compared to a baseline, where a centralized policy was trained using \gls{td3}~\citep{fujimoto2018addressing} in a multi-task setting, highlighting the effectiveness of \gls{smp}. Notably, previous work has shown that training a centralized policy in a multi-task setting, as was done with \gls{td3}, can lead to performance issues due to negative transfer from vastly different tasks \citep{rosenstein2005transfer}.

This study employs a centralized approach to learning, using a single policy to control all actuators in the robot model within a single-task setting. 
This approach demonstrates high performance in a high-dimensional state space, effectively capturing interactions and interdependencies among actuators, and is shown to be transferable and quickly adaptable to new, unseen environments.
Unlike \gls{smp}, the transferred policy does not require significant pretraining in a multi-task setting; instead, it is trained using experiences collected under normal conditions, emphasizing its practicality and efficiency.

\section{Methodology}
\label{sec:methodology}

This section describes the experimental procedures used in this study, which explore the potential of PPO and SAC in enhancing hardware fault tolerance in simulated machines. Experiments are conducted in two Gymnasium environments, Ant-v5 and FetchReachDense-v3, following a three-phase methodology: (1) learning a task in a normal (pre-fault) environment, (2) introducing a hardware fault, and (3) continuing task learning from Phase 1 in a (post-)fault environment. The experimental design spans multiple dimensions of variation, including distinct environments, diverse fault types affecting both dynamics and observations, two reinforcement learning paradigms, and four knowledge-transfer strategies. Taken together, these dimensions support a comprehensive evaluation of adaptation behaviour across varied conditions.

\subsection{Experimental Phases}
\label{sec:experimental_phases}

This section describes the three phases of the experiments.

\subsubsection{Phase 1}
\label{sec:phase_1}

Phase 1 involves having an agent learn a task in a standard Gymnasium environment for $t=t^*$ time steps, using either \gls{ppo} or \gls{sac}. In the Gymnasium Ant-v5 environment, the task requires an agent to learn a gait that maximizes the forward propulsion speed of a four-legged ant. In the Gymnasium FetchReachDense-v3 environment, the task involves an agent learning to move the end effector (or closed gripper) of a robot arm to a randomly generated, three-dimensional goal position as quickly and accurately as possible. The standard Gymnasium environment is referred to as the normal (pre-fault) environment. 

At the end of Phase 1, the knowledge acquired by the agent in the normal environment can be represented as  ${\mathcal{K}}_{t=t^*}{=}\{\theta_{PPO}, {\mathcal{M}}\}$ or  ${\mathcal{K}}_{t=t^*}{=}\{\theta_{SAC}, {\mathcal{B}}\}$, respectively.

\subsubsection{Phase 2}
\label{sec:phase_2}

In Phase 2, a hardware fault is introduced into the environment. This modification aims to replicate a real-world hardware fault for experimental purposes. The hardware faults replicated in this study are described in detail in Section \ref{sec:hardware_faults}.

\subsubsection{Phase 3}
\label{sec:phase_3}

Phase 3 involves the continuation of learning the task with a robot model experiencing the hardware fault introduced in Phase 2. The modified Gymnasium environment is referred to as the fault environment. 

At the start of Phase 3, the agent's knowledge from Phase 1, ${\mathcal{K}}_{t=t^*}$, is transferred.\footnote{It is assumed that a fault has been detected.} 
To investigate knowledge transfer in the context of continual learning, a study is conducted with four different knowledge transfer approaches, as shown in Table~\ref{tab:knowledge_transfer_approaches}.

\begin{table}[t]
  \centering
  \scriptsize
  \begin{tabularx}{\linewidth}{C C C}
    \toprule
    & \multicolumn{2}{c}{RL Algorithm} \\ \midrule
    Knowledge Transfer Approach & PPO & SAC \\ 
    
    \midrule
    
    Approach 1 & $\text{retain}\,\theta_{\text{PPO}},\ \text{retain}\,\mathcal{M}$ &
                 $\text{retain}\,\theta_{\text{SAC}},\ \text{retain}\,\mathcal{B}$ \\

    \addlinespace
    
    Approach 2 & $\text{retain}\,\theta_{\text{PPO}},\ \text{discard}\,\mathcal{M}$ &
                 $\text{retain}\,\theta_{\text{SAC}},\ \text{discard}\,\mathcal{B}$ \\

    \addlinespace
    
    Approach 3 & $\text{discard}\,\theta_{\text{PPO}},\ \text{retain}\,\mathcal{M}$ &
                 $\text{discard}\,\theta_{\text{SAC}},\ \text{retain}\,\mathcal{B}$\\

    \addlinespace
    
    Approach 4 (baseline) & $\text{discard}\,\theta_{\text{PPO}},\ \text{discard}\,\mathcal{M}$ &
                           $\text{discard}\,\theta_{\text{SAC}},\ \text{discard}\,\mathcal{B}$ \\ 
    
    \bottomrule
  \end{tabularx}
  \captionsetup{justification=justified}
  \caption{The four knowledge-transfer approaches in this study.}
  \label{tab:knowledge_transfer_approaches}
\end{table}

\subsection{Hardware Faults}
\label{sec:hardware_faults}

This section details the specific hardware faults introduced in the experiments.

\subsubsection{Ant-v5}
\label{sec:ant_v2}

In the Ant-v5 environment, a fault is introduced to the right rear leg of the ant. This leg is one of two legs primarily responsible for generating the forward pushing motion of the ant, while the other two legs primarily contribute to stability, as evidenced by numerous runs conducted within the normal environment. By introducing a fault to the right rear leg, the agent's learned policy is no longer effective---its ability to effectively propel the ant forward is impeded. Consequently, the agent experiences difficulty in achieving the desired task of maximizing the ant's forward speed. 

\begin{figure*}[!ht]
    \centering
    \begin{subfigure}[t]{0.37\linewidth}
    \includegraphics[width=\linewidth,height=31.9mm,
                trim=0 0 0 120, clip]{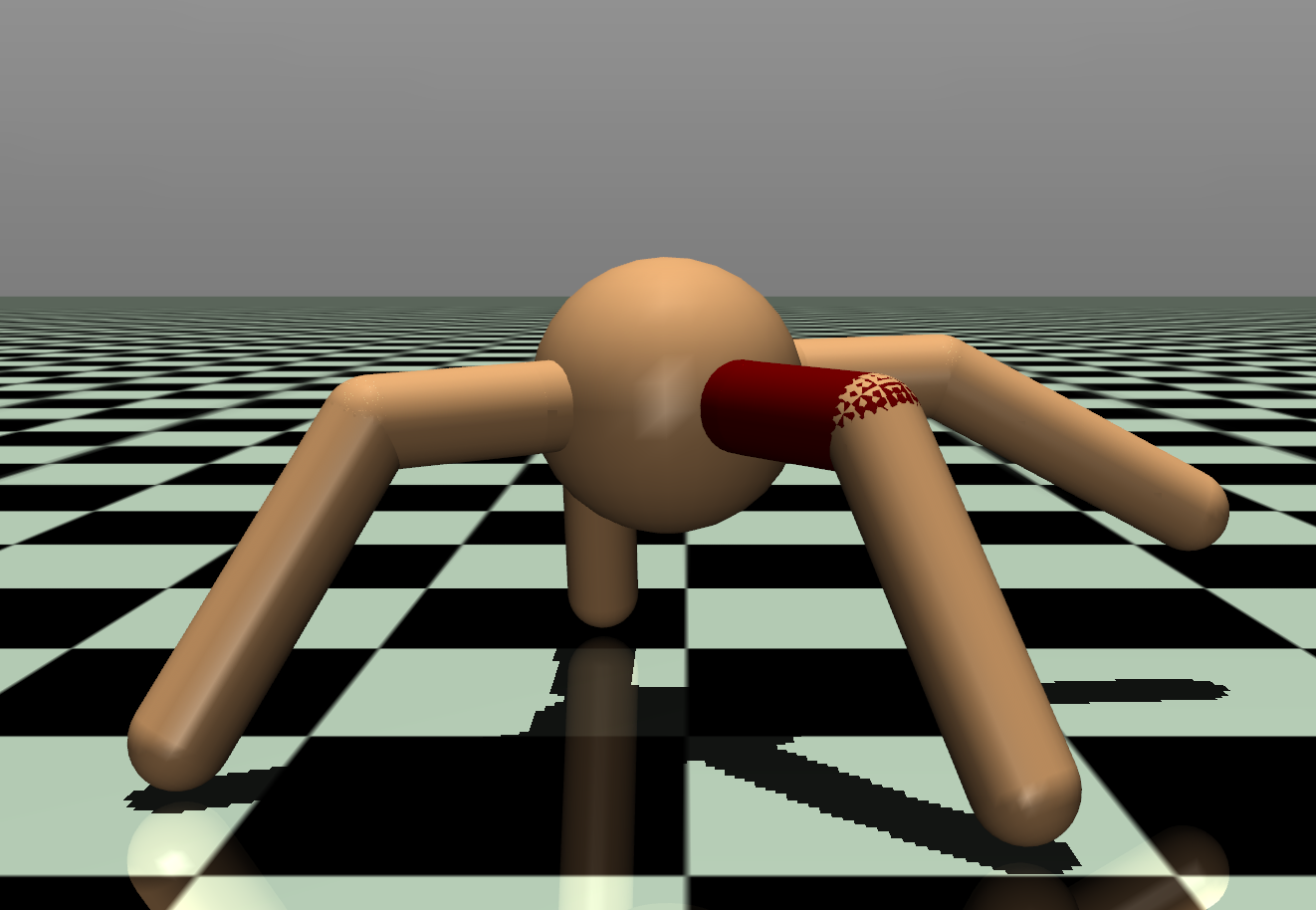}
    \caption{Hip ROM Restriction}
    \label{fig:antv2_hip_rom_restriction}
    \end{subfigure}
    \begin{subfigure}[t]{0.37\linewidth}
    \includegraphics[width=\linewidth,keepaspectratio]{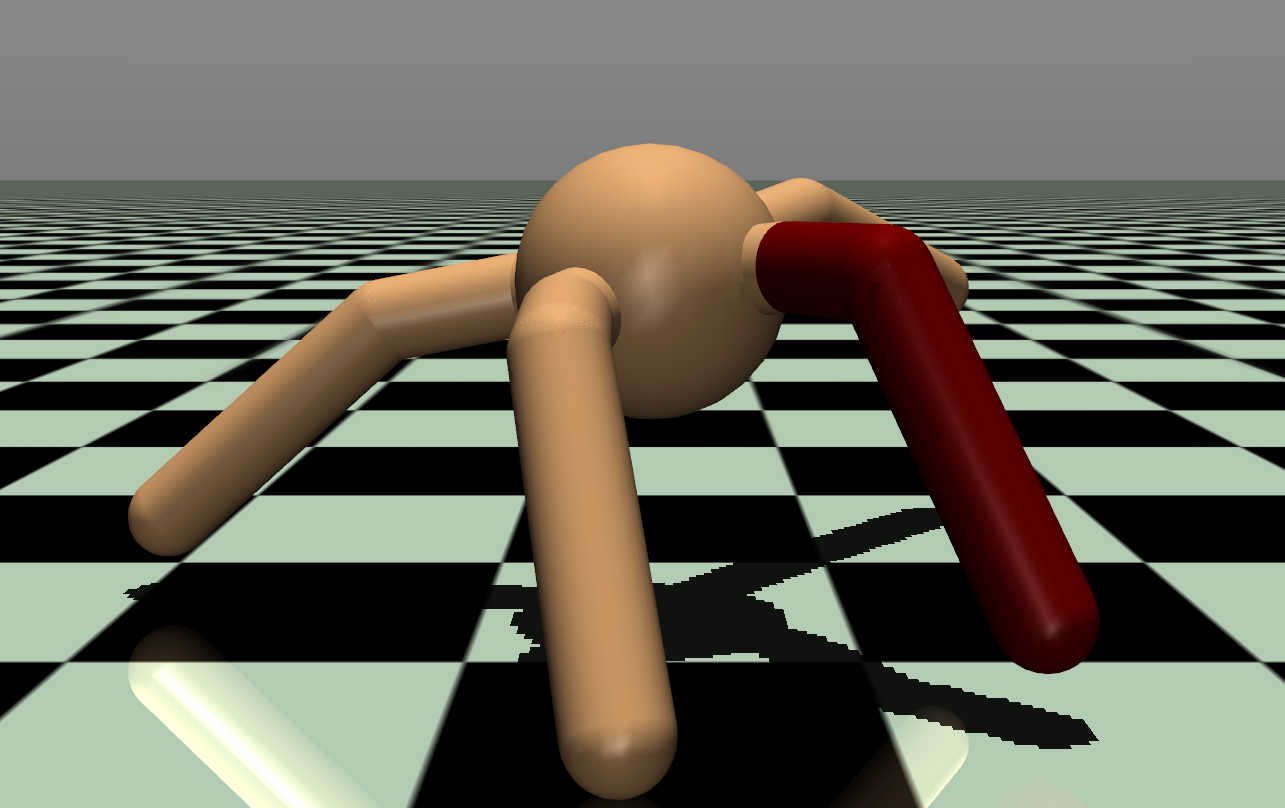}
    \caption{Ankle ROM Restriction}
    \label{fig:antv3_ankle_rom_restriction}
    \end{subfigure}
    
    \begin{subfigure}[t]{0.37\linewidth}
    \includegraphics[width=\linewidth, keepaspectratio,
                trim=0 38 0 56, clip]{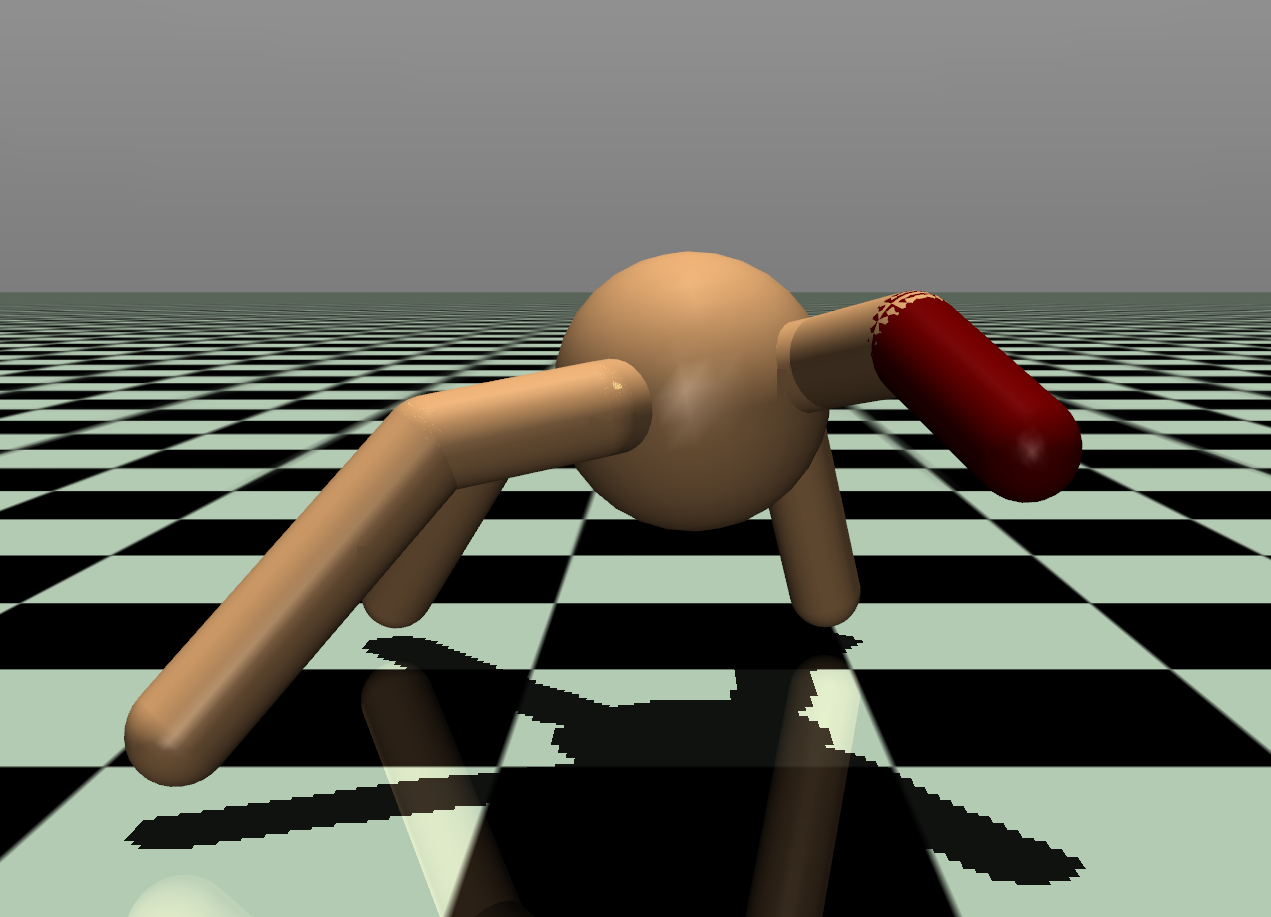}
    \caption{Broken, Severed Linkage}
    \label{fig:antv1_broken_severed_limb}
    \end{subfigure}
    \begin{subfigure}[t]{0.37\linewidth}
    \includegraphics[width=\linewidth,keepaspectratio]{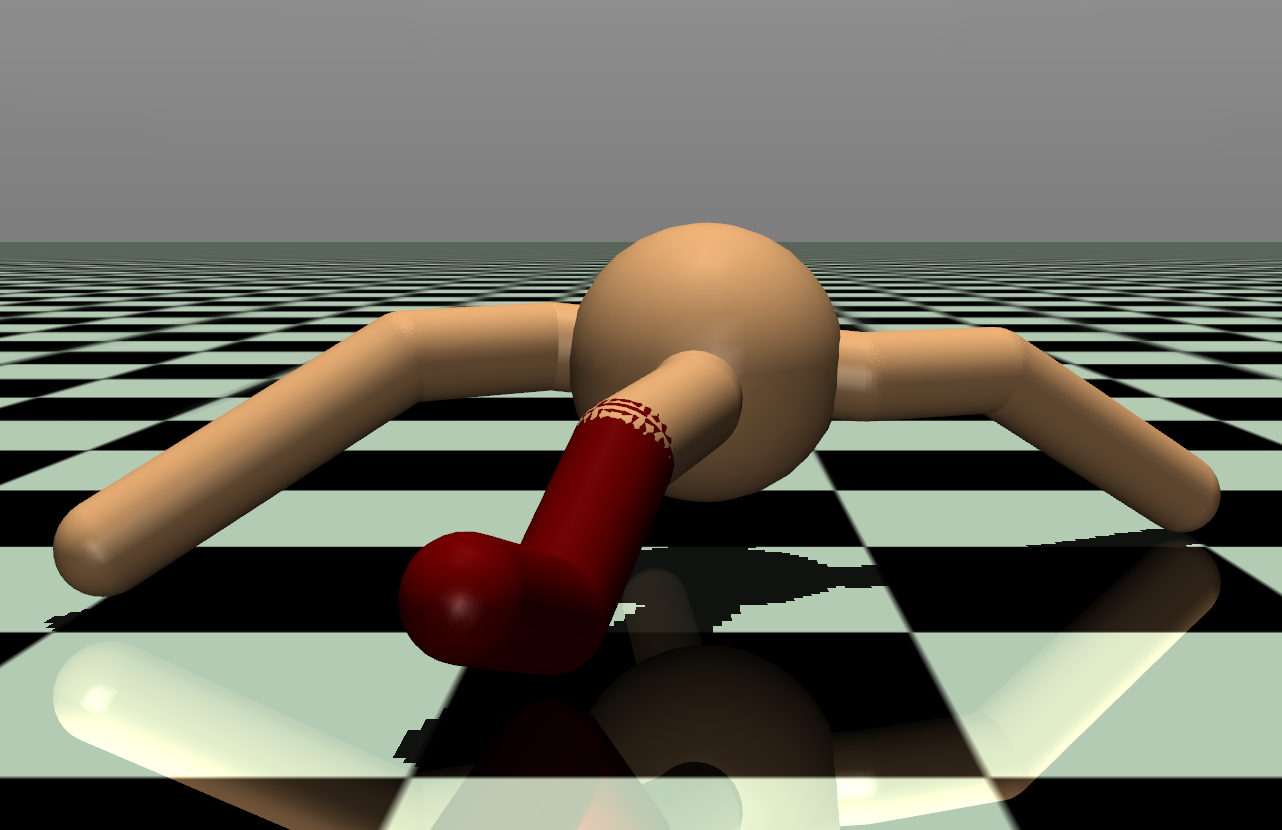}
    \caption{Broken, Unsevered Linkage}
    \label{fig:antv4_broken_unsevered_limb}
    \end{subfigure}
    \captionsetup{justification=justified}
    \caption{Visualization of four simulated hardware faults introduced to the right rear leg of the Ant-v5 robot. Affected joints and links are highlighted in red. (a) Hip joint range of motion is restricted to \([ -5^\circ, 5^\circ ]\). (b) Ankle joint range of motion is limited to \([ 65^\circ, 70^\circ ]\). (c) Lower leg link is shortened by 50\% to simulate a broken and severed linkage. (d) Lower leg link is similarly shortened and connected via an unactuated ball joint to simulate a broken but unsevered linkage.}
    
    \label{fig:antv2_faults}
\end{figure*}

Four distinct faults are implemented as separate instances of the Ant-v5 environment, as shown in Figure \ref{fig:antv2_faults}. Each fault independently alters the underlying dynamics of the environment, thereby impacting the ant's locomotion capabilities. The following paragraphs describe each fault type in detail.

\paragraph{\textbf{Reduced Range of Motion}}
\label{sec:reduced_range_of_motion}

In real-world machines, the presence of external or internal debris can disrupt the gear mechanics within a joint, limiting its \gls{rom}. To simulate this fault, two cases that restrict the \gls{rom} of a joint in the ant's right rear leg are considered: (1) limiting the hip joint motion to the range \([ -5^\circ, 5^\circ ]\) (Figure~\ref{fig:antv2_hip_rom_restriction}), and (2) limiting the ankle joint motion to the range \([ 65^\circ, 70^\circ ]\) (Figure~\ref{fig:antv3_ankle_rom_restriction}).\footnote{The original ranges of hip joint motion and ankle joint motion are \([ -30^\circ, 30^\circ ]\) and \([ 30^\circ, 70^\circ ]\), respectively.}

\paragraph{\textbf{Broken Limb}}
\label{sec:broken_limb}
Real-world robot linkages are susceptible to damage from physical impacts \citep{ambaye2024robot}. A linkage, the rigid section between the joints on a robot, can sustain damage in such a way that it can become broken and (1) dissociated from the robot frame, or (2) partially dissociated from the robot frame, with the linkage held together by a weakened, unbroken section or inner connector cables. In these experiments, two types of faults are considered. To simulate the former (i.e., a broken, severed linkage, as shown in Figure~\ref{fig:antv1_broken_severed_limb}), the length of the lower link in the ant's right rear leg is reduced by half. 
To simulate the latter (i.e., a broken, unsevered linkage, as shown in Figure~\ref{fig:antv4_broken_unsevered_limb}), the same modification is applied, but an unactuated ball joint with three unconstrained rotational degrees of freedom is added at the end of the shortened link, followed by a section identical in shape and length to the removed portion. The observation vector is modified to exclude data from the added joint and link, ensuring its dimensionality matches that of the Ant-v5 normal environment.

\subsubsection{FetchReachDense-v3}
\label{sec:fetchreach_v1}

\begin{figure}[!ht]
    \centering
    \begin{subfigure}[t]{0.37\textwidth}
        \centering
        \includegraphics[width=\linewidth,
                trim=0 3 0 2.5, clip]{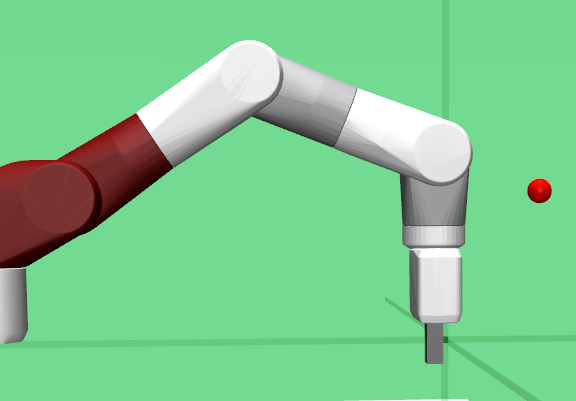}
        \caption{Frozen Shoulder Lift Position Sensor}
        \vspace{0.2cm}
        \label{fig:fetchreachf1_frozen_shoulder_lift_position_sensor}
    \end{subfigure}
    \begin{subfigure}[t]{0.37\textwidth}
        \centering
        \includegraphics[width=\linewidth]{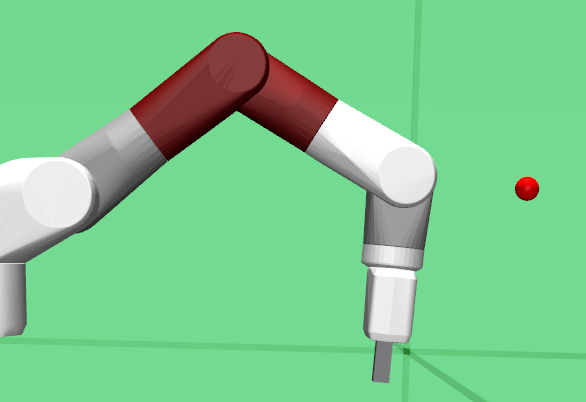}
        \caption{Slippery Elbow Flex Joint}
        \vspace{0.2cm}
        \label{fig:fetchreachf2_slippery_elbow_flex_joint}
    \end{subfigure}
    \captionsetup{justification=justified}
    \caption{Visualization of two simulated hardware faults introduced in the FetchReachDense-v3 environment. Affected joints and links are highlighted in red. (a) The shoulder lift joint's position sensor is frozen, continuously reporting a fixed value of -1.5 radians regardless of the joint's actual configuration, creating a mismatch between observation and ground truth. (b) The elbow flex joint exhibits a slippery fault, consistently overshooting its commanded position by 0.05 radians and mimicking the effect of a damaged or worn gear mechanism.}
    \label{fig:fetchreachv3_faults}
\end{figure}

In the FetchReachDense-v3 environment, a fault is introduced to the shoulder lift joint in one instance and to the elbow flex joint in another, as shown in Figure \ref{fig:fetchreachv3_faults}. 
These joints were chosen because they are crucial for the robot arm to succeed in its task.
A fault in either joint causes the robot arm to experience inaccuracies, resulting in delays in reaching the goal position with its end effector.

\paragraph{\textbf{Frozen Position Sensor}}
\label{sec:frozen_position_sensor}

A frozen sensor is a common type of sensor fault. A frozen sensor consistently reports a constant value that may not accurately reflect the physical property being measured \citep{haro2025failure}. The first fault in FetchReachDense-v3, depicted in Figure \ref{fig:fetchreachf1_frozen_shoulder_lift_position_sensor}, is a frozen shoulder lift position sensor. To simulate this fault, the robot's shoulder lift position sensor is modified so that it continuously reports a position of $-1.5$ radians. This inaccurately reported position subsequently affects the computed position of the robot's end effector, which is calculated using forward kinematics. As a result, this fault impacts the observation obtained in FetchReachDense-v3. Notably, this fault does not alter the underlying dynamics of the environment itself. Instead, it introduces a discrepancy by corrupting the observation---one entry no longer accurately reflects the true state of the robot due to a frozen sensor fault.

\paragraph{\textbf{Position Slippage}}
\label{sec:position_slippage}

In mechanical gears, broken teeth can cause the gear to slip to the next intact tooth, leading to operational issues and inaccurate movements \citep{srikanth2014slippage}. The second fault in FetchReachDense-v3, depicted in Figure \ref{fig:fetchreachf2_slippery_elbow_flex_joint}, is a slippery elbow flex joint. This fault causes the joint to consistently overshoot its commanded position by a fixed offset whenever an action is applied. Specifically, if the joint is commanded to move by \( x \) radians, the fault causes it to move by \( x + c \) radians, where \( c = 0.05 \) radians is a constant offset actuator error.

\subsection{Evaluation}
\label{sec:evaluation}

For each experiment, 30 runs are conducted using seeds ranging from 0 to 29. All experiments assess learning progress through policy evaluation rollouts. At predefined intervals, learning is temporarily paused and the current policy is evaluated for 10 episodes. The reported evaluation metric is the average return in a policy evaluation rollout, computed over 30 runs. Additionally, 95\% confidence intervals, computed using a t-distribution, are shown as shaded regions in the plots.

In the primary experiments, both the normal and fault environments were allowed sufficient learning durations to ensure convergence, although the number of time steps may differ between them. In the Ant-v5 environment, \gls{ppo} trains for 20,000,000 time steps in the normal environment and 30,000,000 time steps in the fault environments, with evaluations every 200,000 time steps. \gls{sac} is trained for 3,000,000 time steps in both the normal and fault environments, with evaluations every 30,000 time steps. In FetchReachDense-v3, \gls{ppo} is trained for 50,000 time steps in both the normal and fault environments, with evaluations every 500 steps. For \gls{sac}, the normal environment is trained for 10,000 time steps and the fault environment for 100,000 time steps, with evaluations every 100 steps. Evaluation frequencies were selected to provide a sufficiently detailed view of learning dynamics without incurring excessive computational overhead.

To highlight the early adaptation capabilities of each algorithm under each fault condition, performance data is required at 300,000 time steps for Ant-v5 and 30,000 time steps for FetchReachDense-v3. This data already exists for all algorithm-fault pairings except \gls{ppo} in each Ant-v5 fault environment. To address this, an additional evaluation for \gls{ppo} is conducted in each Ant-v5 fault environment using a reduced training duration of 300,000 time steps, with evaluations at 0, 200,000 and 300,000 time steps.

\subsection{Algorithm Implementation}
\label{sec:algorithm_implementation}

This section presents a concise overview of the implementation of two \gls{rl} algorithms---\glsentryfull{ppo} and \glsentryfull{sac}.
The neural network architectures for each algorithm are described, including the number of layers and nodes.
Specific implementation choices, such as code-level optimizations, are also highlighted.

\subsubsection{\glsentrylong{ppo}}
The implementation of \gls{ppo} uses a single neural network 
comprising a policy network and a value network. 
Both internal networks are designed as feedforward networks with two hidden layers, 
each consisting of 64 nodes and using hyperbolic tangent activation functions. 
The policy network incorporates a learnable parameter known as the log standard deviation,
which is initially set to $0$. 
The Adam optimizer \citep{kingma2014adam} is used for optimization.
To improve the performance of \gls{ppo}, code-level optimizations recommended by~\cite{engstrom2020implementation} are incorporated, including linear learning rate decay, use of a generalized advantage estimator \citep{mnih2016asynchronous}, orthogonal initializations \citep{saxe2013exact}, and hyperbolic tangent activation functions.

Notably, prior to learning in the fault environment, \gls{ppo}'s learning rate is reset to its initial value. This reset allows the algorithm to adapt to the changed environment.

\subsubsection{\glsentrylong{sac}} The implementation of \gls{sac} uses two twinned soft Q-networks and a policy network. All networks are feedforward networks with two hidden layers, each consisting of 256 nodes and using \gls{relu} activations. Each layer is initialized using Xavier uniform initialization \citep{glorot2010understanding}. Optimization is performed using the Adam optimizer \citep{kingma2014adam}.

\subsubsection{Hyperparameter Optimization}
\label{sec:hyperparameter_optimization}
Real-world machines are typically optimized to perform effectively under normal conditions. When a fault occurs, conducting a new hyperparameter search tailored to the changed conditions can be time-consuming and computationally demanding. Therefore, hyperparameter optimization is restricted to learning in the Ant-v5 and FetchReachDense-v3 normal environments.

To optimize the hyperparameters of each algorithm, the Optuna framework \citep{akiba2019optuna} is used to perform automated search.
Hyperparameter values are sampled from either specified continuous intervals \([a, b]\) or discrete sets \(\{x_1, x_2, \dots\, x_n\}\). 
For each environment-algorithm pairing, 200 unique hyperparameter configurations are evaluated.
For each configuration, 5 runs are conducted using NumPy~\cite{harris2020array} random seeds 0 to 4. 
The best configuration is selected based on the highest average return across the entire training process, aggregated over all five seeds.
Tables \ref{tab:ppo_hps} and \ref{tab:sac_hps} summarize the best-performing hyperparameters for each standard Gymnasium environment (i.e., normal environment).

\begin{table}[!ht]
    \centering
    \scriptsize
    \begin{tabularx}{\linewidth}{@{\extracolsep{\fill}} 
    p{0.35\linewidth} 
    >{\centering\arraybackslash}p{0.25\linewidth} 
    C 
    C
    }
        \toprule
        \textbf{Hyperparameter} & 
        \textbf{Possible Values} &
        \textbf{Ant} & 
        \textbf{FetchReach} \\
        
        \midrule
        
        learning rate & 
        $[0.00001, 0.001]$&
        $0.0001672$ &
        $0.0008641$ \\
        
        linear learning rate decay &
        \{True, False\} &
        True &
        True\\
        
        gamma ($\gamma$) &  
        $[0.8, 0.9999]$ &
        $0.9960$ &
        $0.8301$ \\
        
        number of samples & 
        $\{1024, 2048, \dots, 8192\}$ &
        $4096$ &
        $256$ \\
        
        mini-batch size & 
        $\{32, 64, \dots, 512\}$ &
        $32$ &
        $32$ \\
        
        epochs &  
        $\{3, 4, \dots, 10\}$ &
        $5$ &
        $10$ \\
        
        epsilon ($\epsilon$) & 
        $[0.1, 0.3]$ &
        $0.2458$ &
        $0.2887$ \\
        
        value function loss coefficient ($c_1$) & 
        $[0.1, 1.0]$ &
        $0.4853$&
        $0.1410$\\
        
        policy entropy coefficient ($c_2$) &
        $[0.0001, 0.1]$ &
        $0.003953$ &
        $0.01380$\\
        
        clipped value function & 
        \{True, False\} &
        False &
        False\\
        
        max grad norm &
        \{0.5, 1.0\} & 
        0.5 &
        0.5\\
        
        use GAE & 
        \{True, False\} &
        True &
        True\\
        
        GAE lambda & 
        $[0.9, 1.0]$ &
        $0.9006$&
        $0.9039$\\
        
        normalize rewards &
        \{True, False\} &
        True &
        True\\
        
        \bottomrule
    \end{tabularx}
    \captionsetup{justification=justified}
    \caption{Search spaces and selected values for \gls{ppo} hyperparameters. The 'Possible Values' column lists the range or set of values explored during hyperparameter optimization using Optuna. The final two columns show the best-performing configuration (based on average return across five seeds) for the Ant-v5 and FetchReachDense-v3 environments.}
    \label{tab:ppo_hps}
 \end{table}

\begin{table}[!ht]
    \centering
    \scriptsize
    \begin{tabularx}{\linewidth}{@{\extracolsep{\fill}} 
    p{0.3\linewidth} 
    >{\centering\arraybackslash}p{0.3\linewidth} 
    C 
    C
    }
        \toprule
        \textbf{Hyperparameter} & 
        \textbf{Possible Values} &
        \textbf{Ant} & 
        \textbf{FetchReach} \\
        
        \midrule
        
        learning rate & 
        $[0.00001, 0.001]$ &
        0.0002225 &
        0.0008507 \\
        
        gamma ($\gamma$) &  
        $[0.8, 0.9999]$ &
        0.9815 &
        0.8504 \\
        
        replay buffer size ($\times10^{6}$) & 
        \{0.10, 0.25, 0.50, 0.75, 1.00\}  &
        1.00 &
        0.10\\
        
        batch size & 
        $\{64, 128, \dots, 512\}$ &
        512&
        256\\
        
        automatic entropy tuning &
        \{True, False\} &
        False &
        False\\
        
        alpha $(\alpha)$ &
        $[0.0001, 0.2]$ &
        0.07461 &
        0.1336\\
        
        target update interval &
        $\{1, 2, \dots, 10\}$ &
        8 &
        5\\
        
        target smoothing coefficient ($\tau$) & 
        $[0.001, 0.1]$ &
        0.05151 &
        0.003237\\
        
        normalize rewards &
        \{True, False\} &
        False &
        True\\
        
        \bottomrule
    \end{tabularx}
    \captionsetup{justification=justified}
    \caption{Search spaces and selected values for \gls{sac} hyperparameters. The 'Possible Values' column lists the range or set of values explored during hyperparameter optimization using Optuna. The final two columns show the best-performing configuration (based on average return across five seeds) for the Ant-v5 and FetchReachDense-v3 environments.}
    \label{tab:sac_hps}
\end{table}

\subsubsection{Computational Footprint}
\label{sec:computational_footprint}

In our implementation, \sac has a larger footprint than \ppo due to both model size and storage.
\sac uses a policy network and two twinned soft Q-networks, each with two hidden layers of 256 units, and trains with a replay buffer that contains up to 1.00 million experiences in Ant-v5 and 0.10 million in FetchReachDense-v3.
\ppo uses a policy network and a value network with two hidden layers of 64 units and a small memory that holds at most 4096 and 256 experiences in Ant-v5 and FetchReachDense-v3, respectively.
At inference, only the policy network of each algorithm is used.
Since SAC's policy network has more parameters than PPO's, its inference-time footprint is larger.

\section{Results}
\label{sec:results}

In this section, three distinct focal points of investigation are addressed. 
In Section \ref{sec:performance_sample_efficiency_and_speed}, the differences between \gls{ppo} and \gls{sac} in performance, sample efficiency, and adaptation speed in a fault environment are analyzed. In Section \ref{sec:transfer_of_task_knowledge}, the performance disparities among the four knowledge transfer approaches in a continual \gls{rl} and hardware fault setting are assessed. Finally, in Section \ref{sec:comparison_with_prior_fault_adaptation_approaches}, \gls{ppo} and \gls{sac} are compared with prior methods for hardware fault adaptation.

\subsection{Performance, Sample Efficiency and Real-Time Speed}
\label{sec:performance_sample_efficiency_and_speed}

\begin{figure*}[!ht]
    \centering
    \captionsetup{justification=centering}
    
    \begin{subfigure}{\linewidth}
        \centering
        \includegraphics[width=0.7\linewidth]{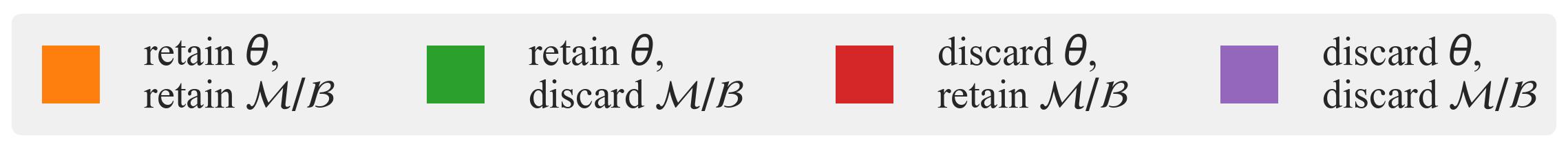}
        \vspace{0.2cm}
    \end{subfigure}

    \begin{subfigure}{\linewidth}
        \centering
        \includegraphics[width=0.48\linewidth]{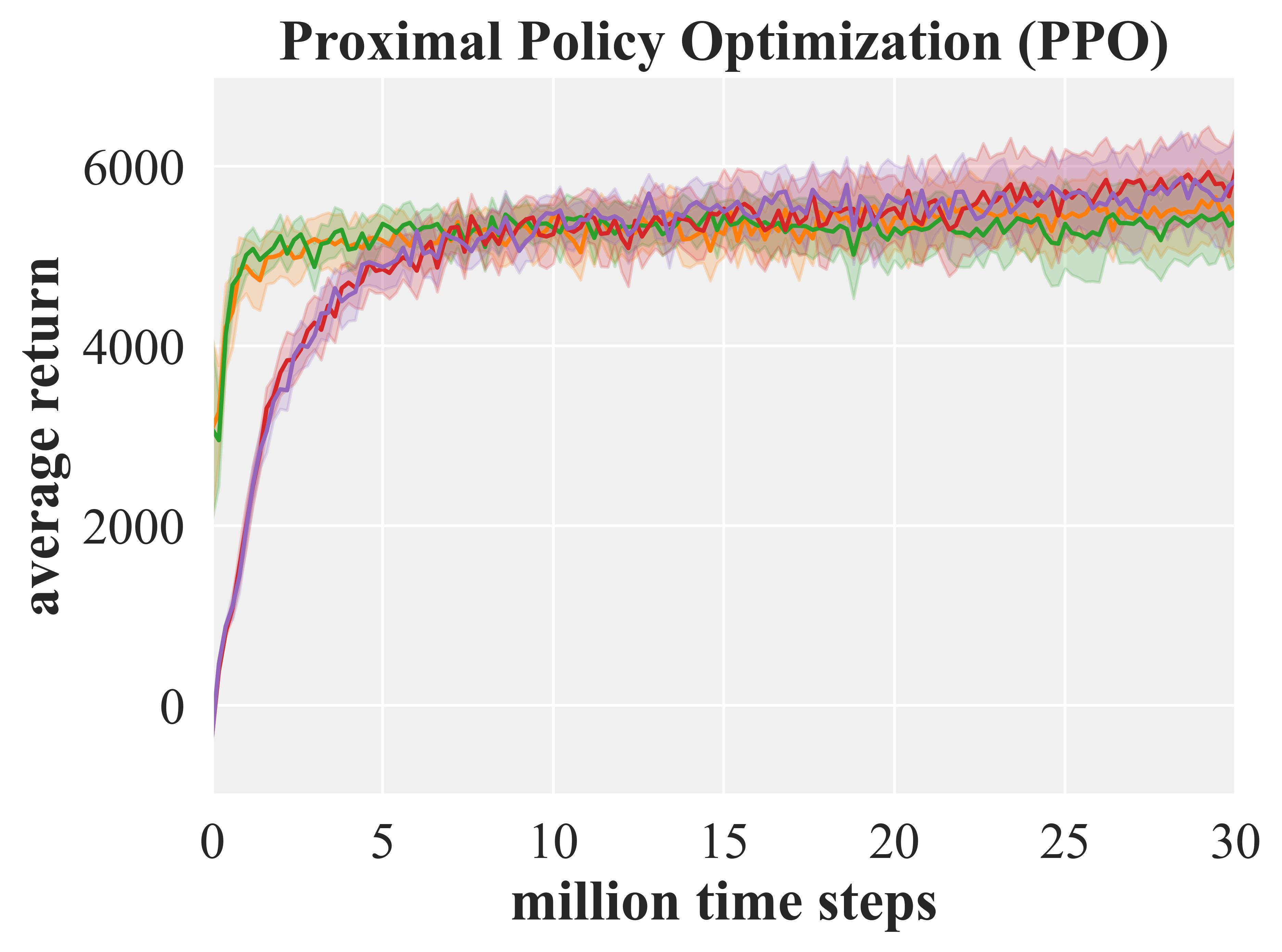}
        \includegraphics[width=0.48\linewidth]{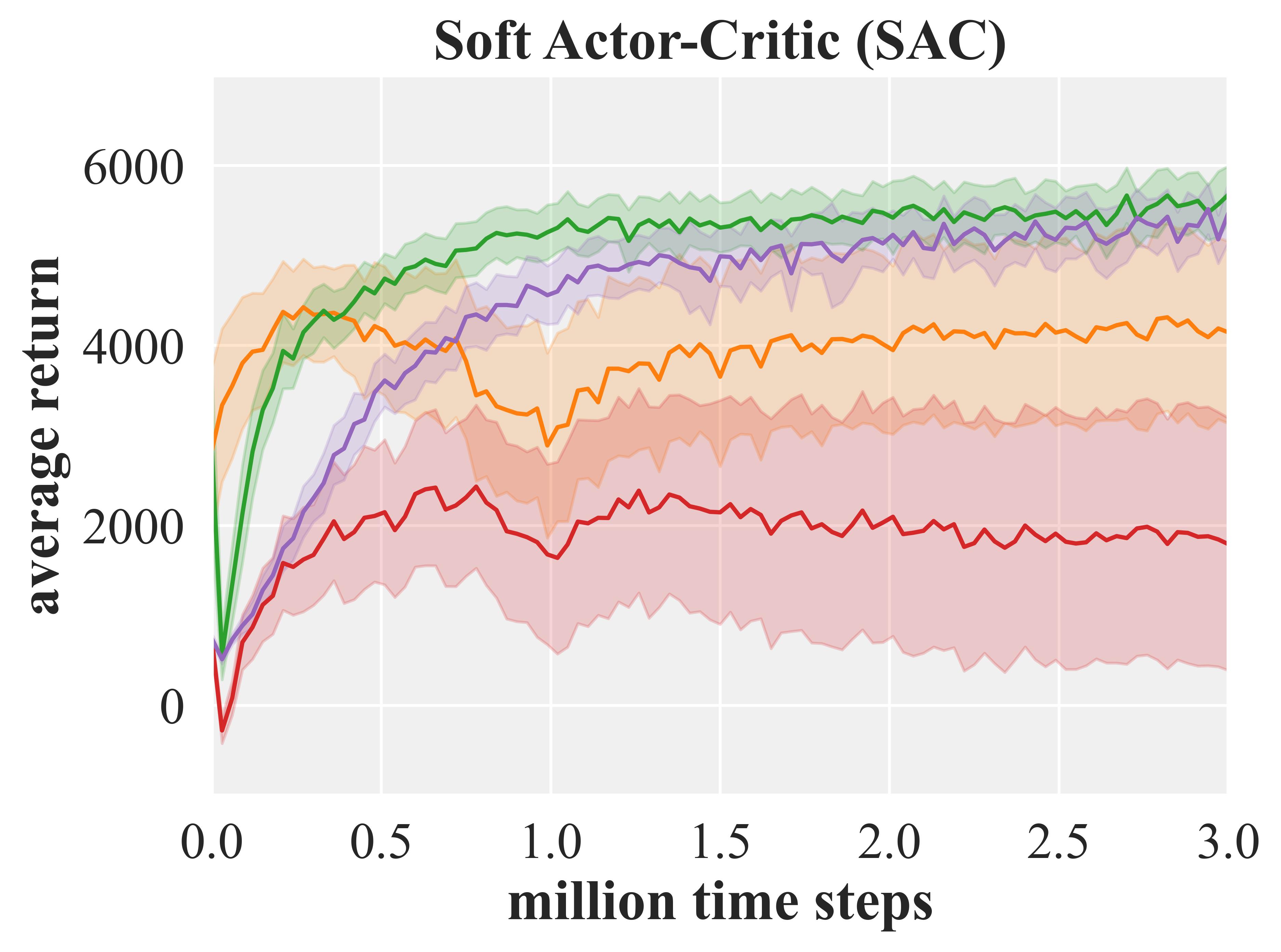}
        \caption{Hip ROM Restriction}
        \vspace{0.2cm}
        \label{fig:antf2}
    \end{subfigure}
    \begin{subfigure}{\linewidth}
        \centering
        \includegraphics[width=0.48\linewidth]{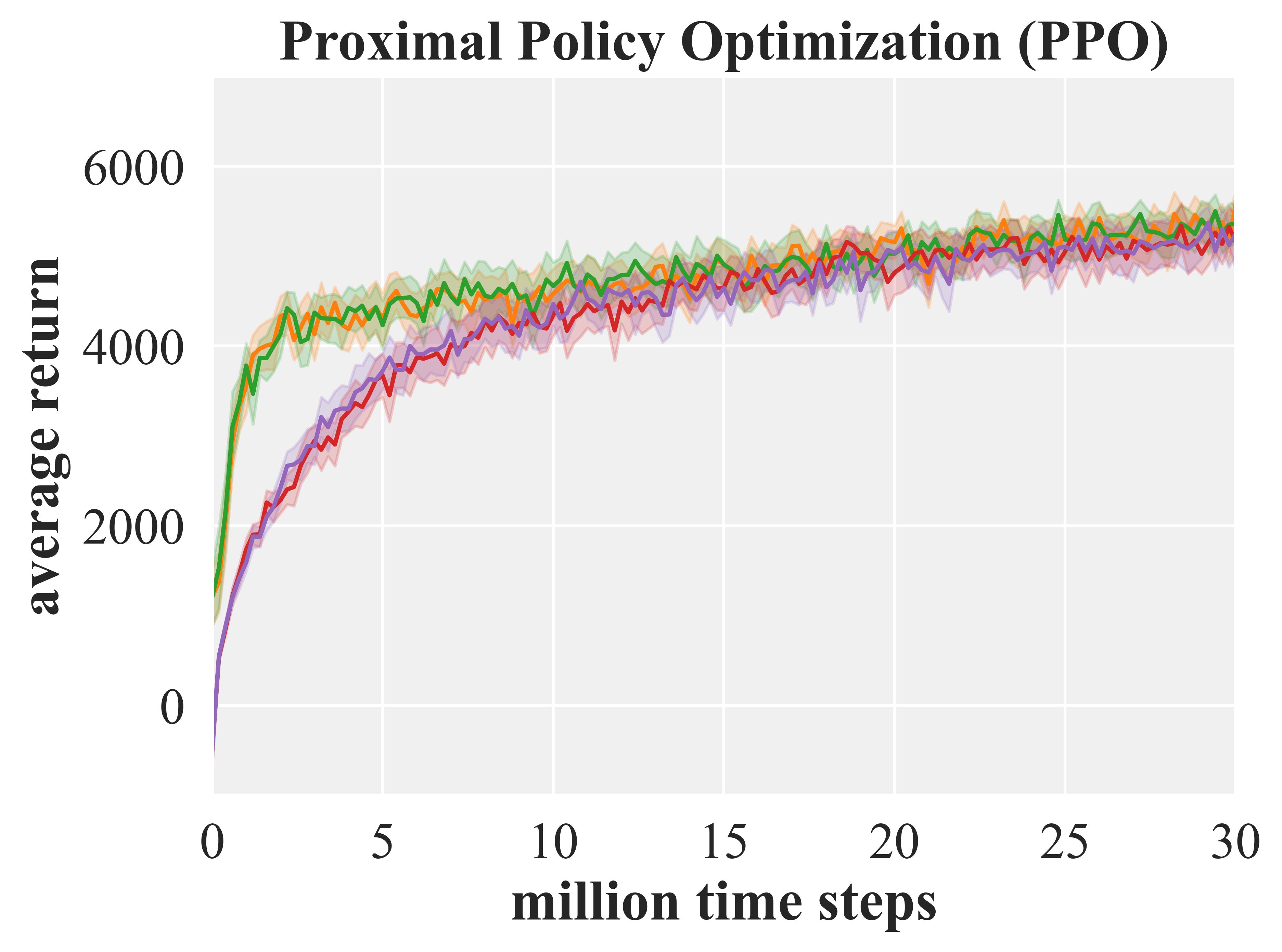}
        \includegraphics[width=0.48\linewidth]{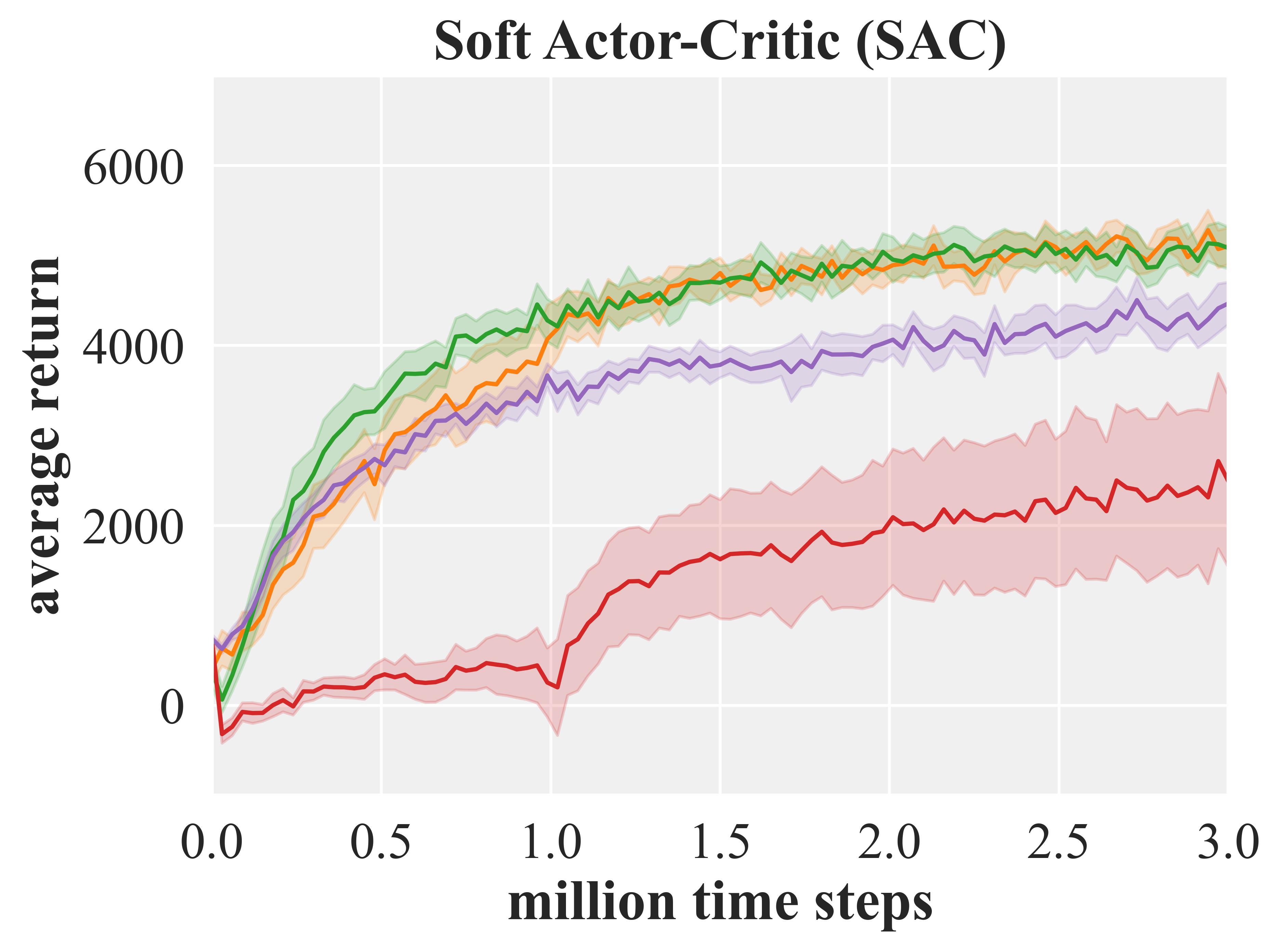}
        \caption{Ankle ROM Restriction}
        \vspace{0.2cm}
        \label{fig:antf1}
    \end{subfigure}
    
    \captionsetup{justification=justified}
    \caption{Learning curves showing the average return of \gls{ppo} and \gls{sac} in the Ant-v5 Hip ROM Restriction and Ankle ROM Restriction fault environments, evaluated under four knowledge transfer approaches. Results are averaged over 30 runs, with shaded regions indicating 95\% confidence intervals. Each subplot shows average return versus training steps (in millions), revealing differences in sample efficiency, adaptation speed, and robustness across algorithms and knowledge transfer approaches.}
    \label{fig:ant_plot_1}
\end{figure*}

\begin{figure*}[!ht]
    \centering
    \captionsetup{justification=centering}
    
    \begin{subfigure}{\linewidth}
        \centering
        \includegraphics[width=0.7\linewidth]{images/legend2.jpg}
        \vspace{0.2cm}
    \end{subfigure}
    
    \begin{subfigure}{\linewidth}
        \centering
        \includegraphics[width=0.48\linewidth]{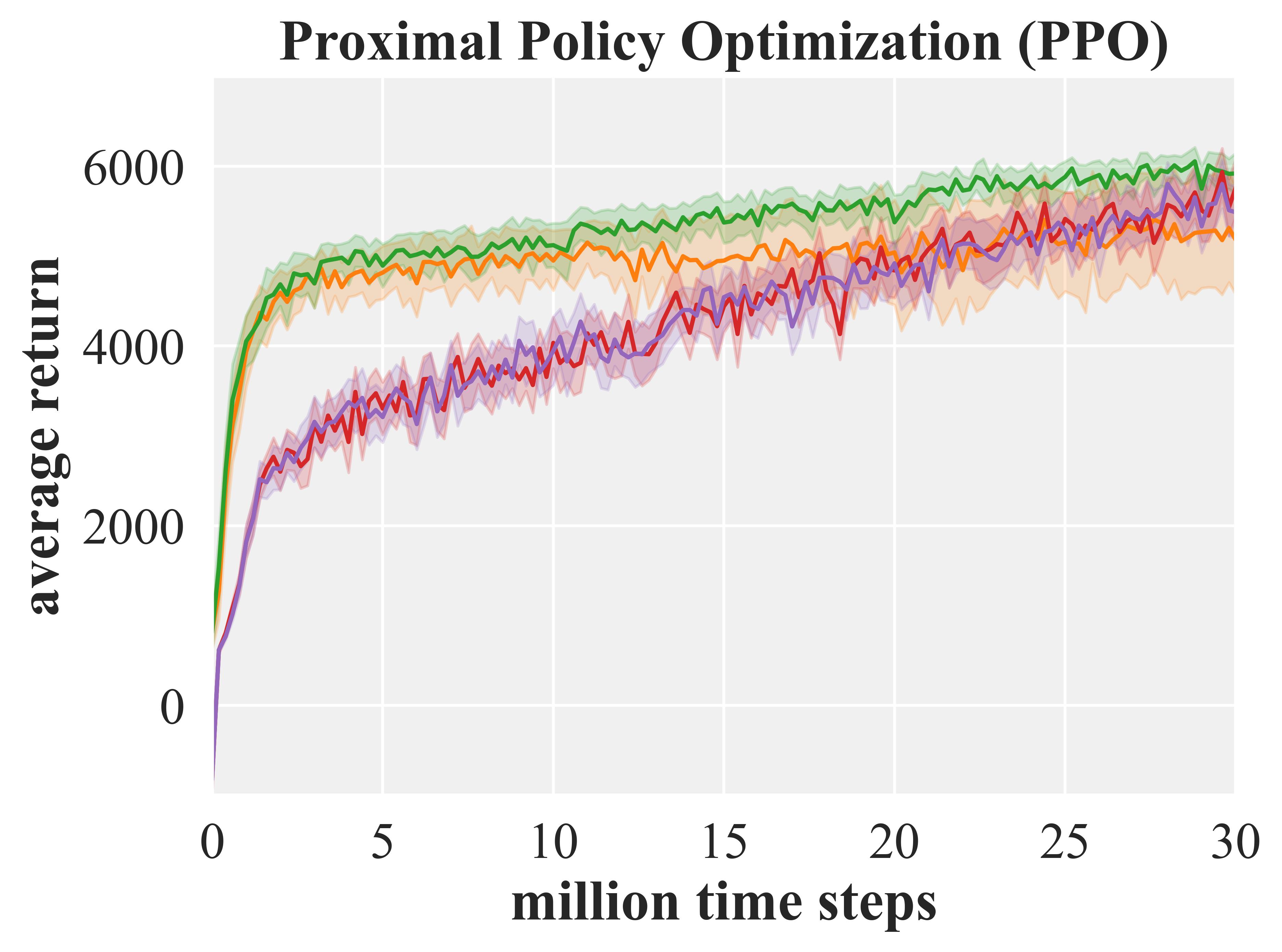}
        \includegraphics[width=0.48\linewidth]{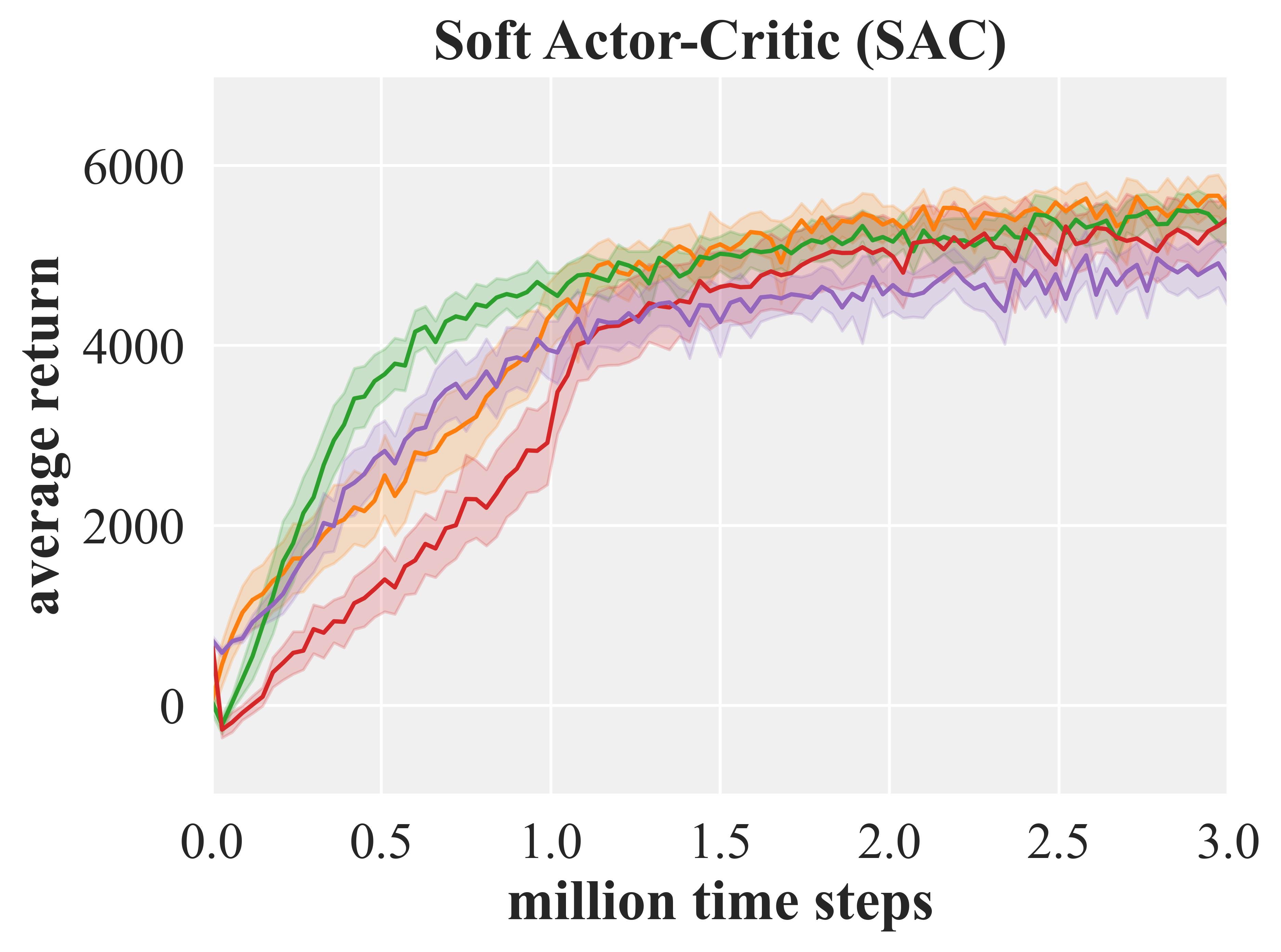}
        \caption{Broken, Severed Limb}
        \vspace{0.2cm}
        \label{fig:antf3}
    \end{subfigure}
    \begin{subfigure}{\linewidth}
        \centering
        \includegraphics[width=0.48\linewidth]{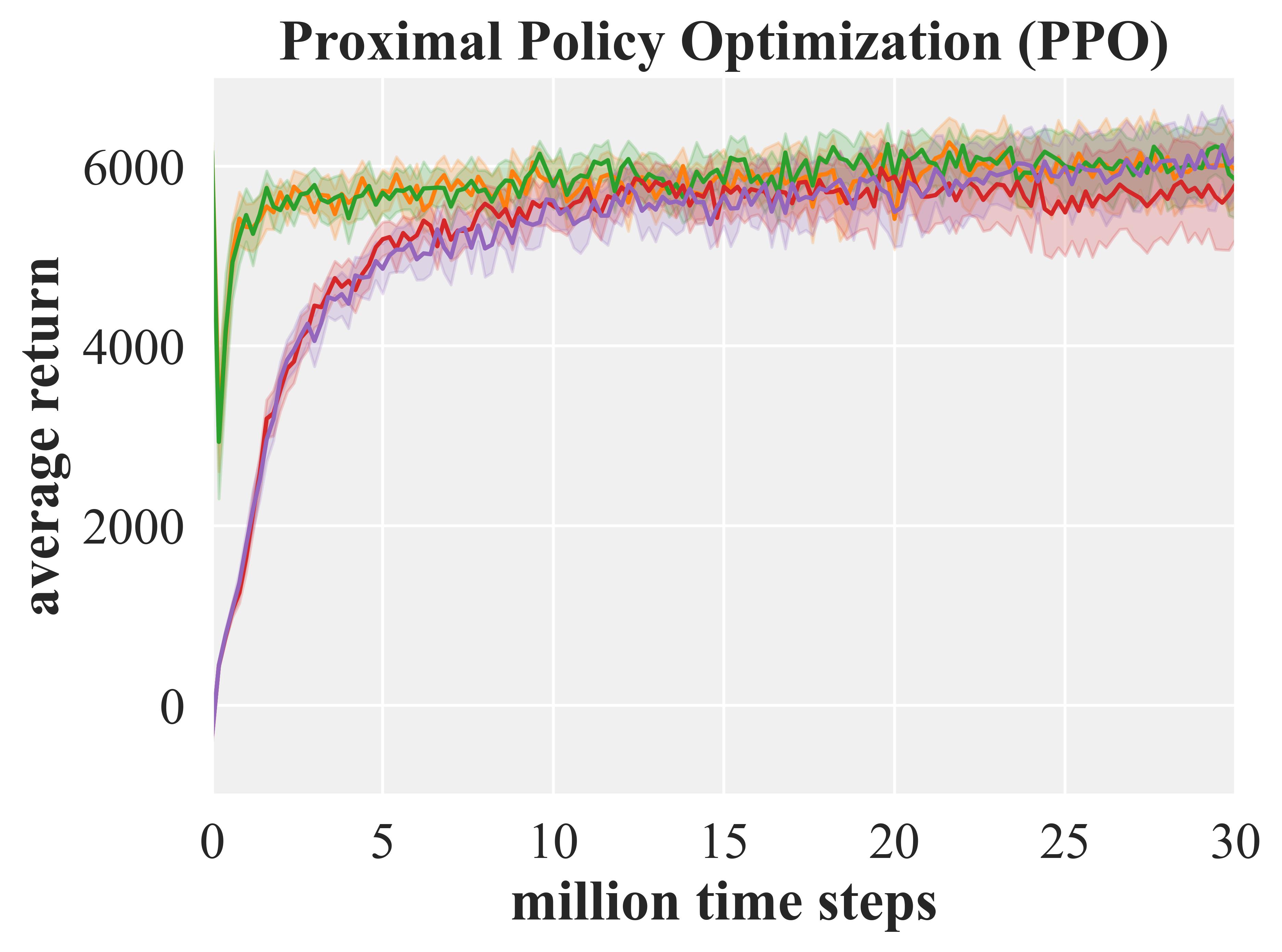}
        \includegraphics[width=0.48\linewidth]{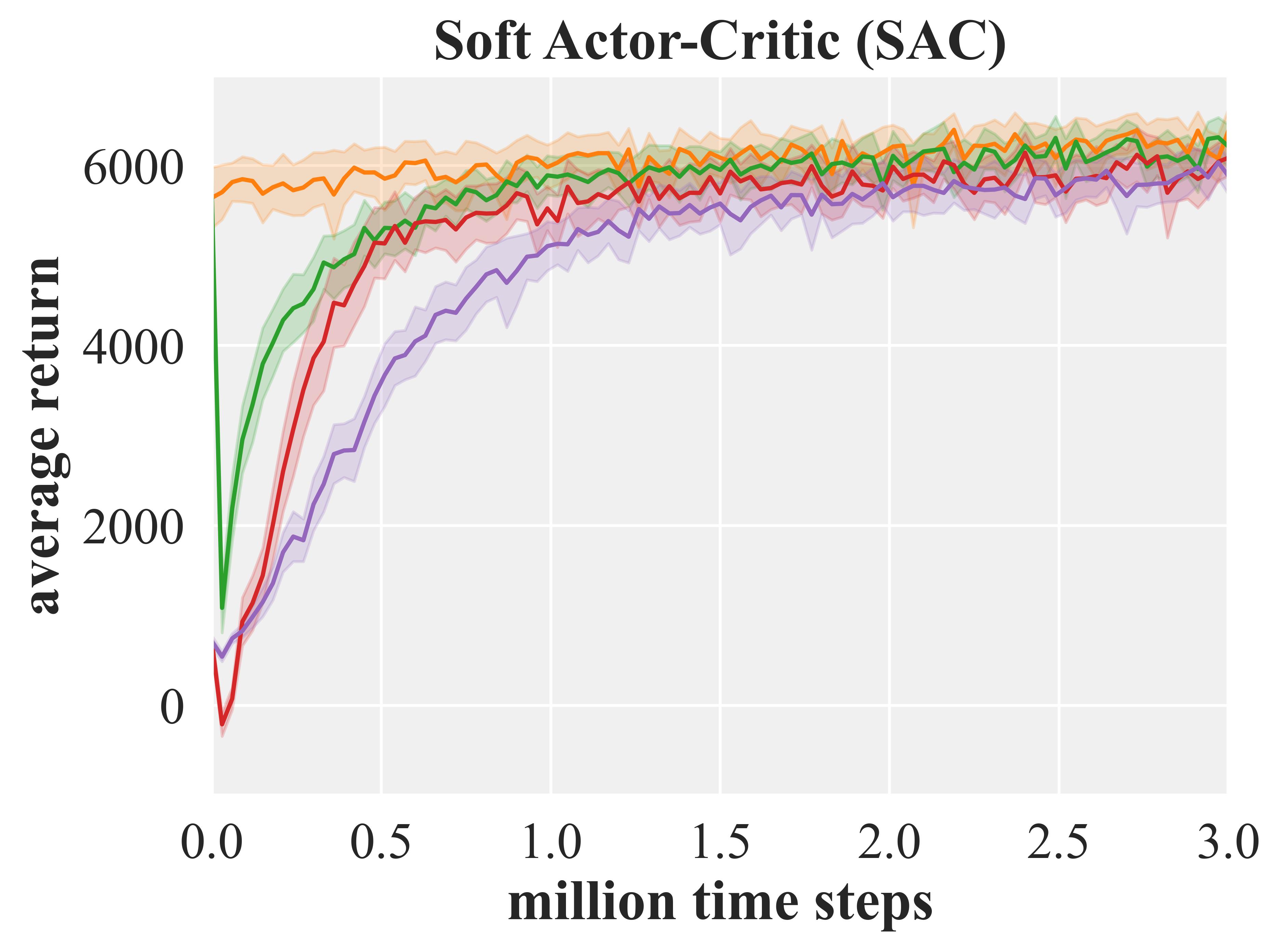}
        \caption{Broken, Unsevered Limb}
        \vspace{0.2cm}
        \label{fig:antf4}
    \end{subfigure}
    
    \captionsetup{justification=justified}
    \caption{Learning curves showing the average return of \gls{ppo} and \gls{sac} in the Ant-v5 Broken, Severed Limb and Broken, Unsevered Limb fault environments, evaluated under four knowledge transfer approaches. Results are averaged over 30 runs, with shaded regions indicating 95\% confidence intervals. Each subplot shows average return versus training steps (in millions), revealing differences in sample efficiency, adaptation speed, and robustness across algorithms and knowledge transfer approaches.}
    \label{fig:ant_plot_2}
\end{figure*}

\begin{figure*}[!ht]
    \centering
    \captionsetup{justification=centering}
    \begin{subfigure}{\linewidth}
        \centering
        \includegraphics[width=0.7\linewidth]{images/legend2.jpg}
        \vspace{0.2cm}
    \end{subfigure}
    \begin{subfigure}{\linewidth}
        \centering
        \includegraphics[width=0.48\linewidth]{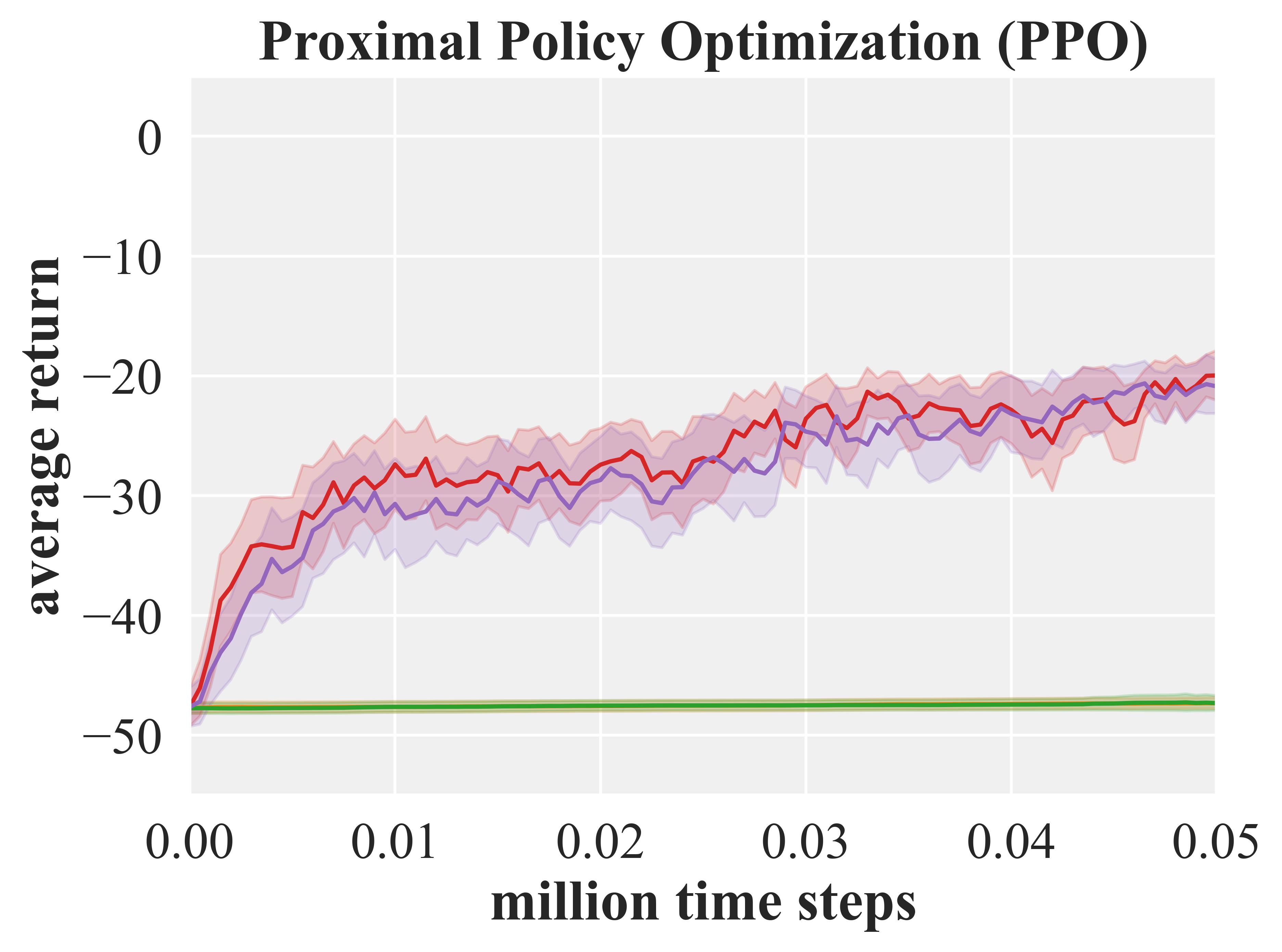}
        \includegraphics[width=0.48\linewidth]{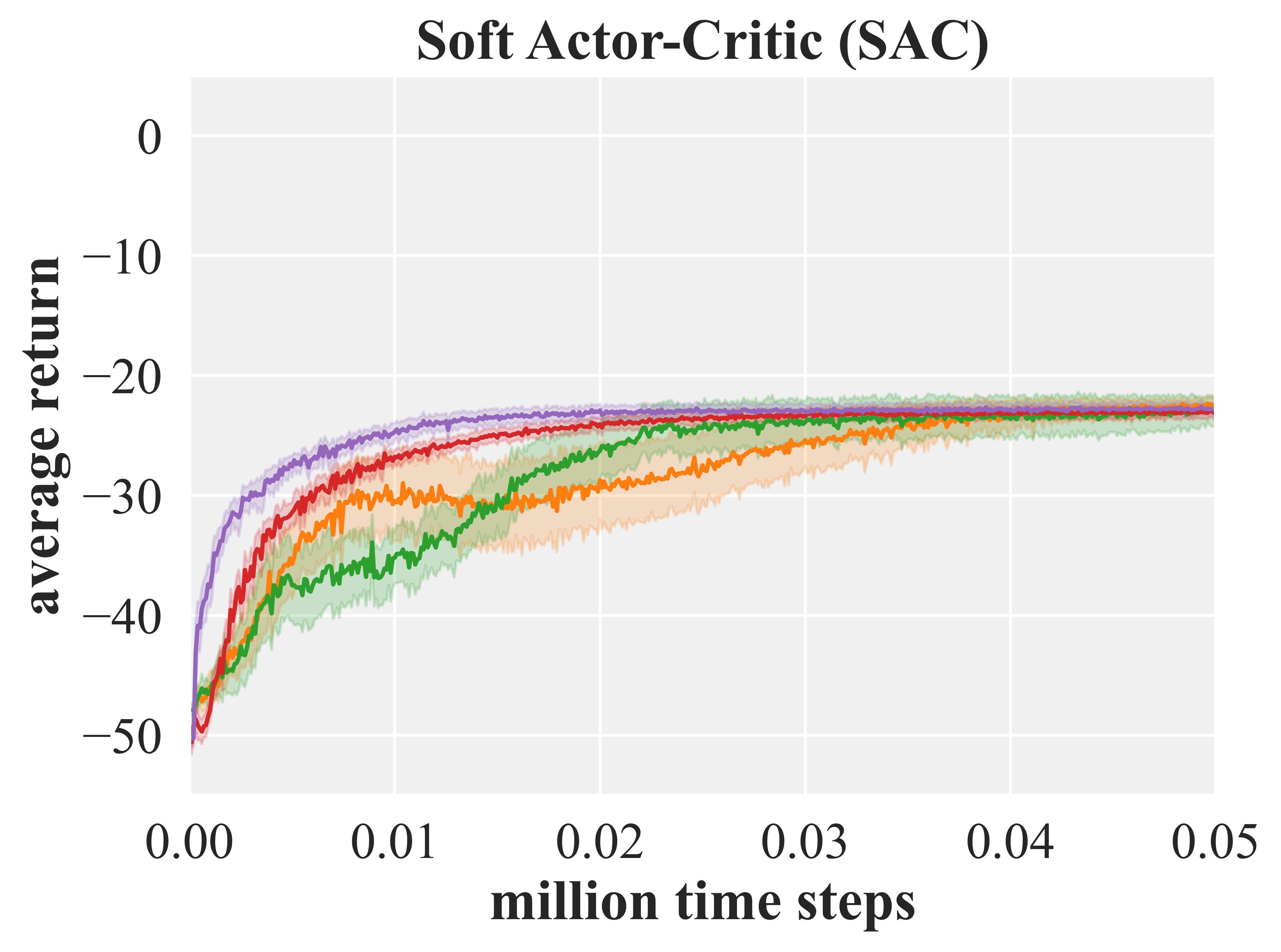}        
        \caption{Frozen Shoulder Lift Position Sensor}
        \vspace{0.2cm}
        \label{fig:fetchreachf1}
    \end{subfigure}
    \begin{subfigure}{\linewidth}
        \centering
        \includegraphics[width=0.48\linewidth]{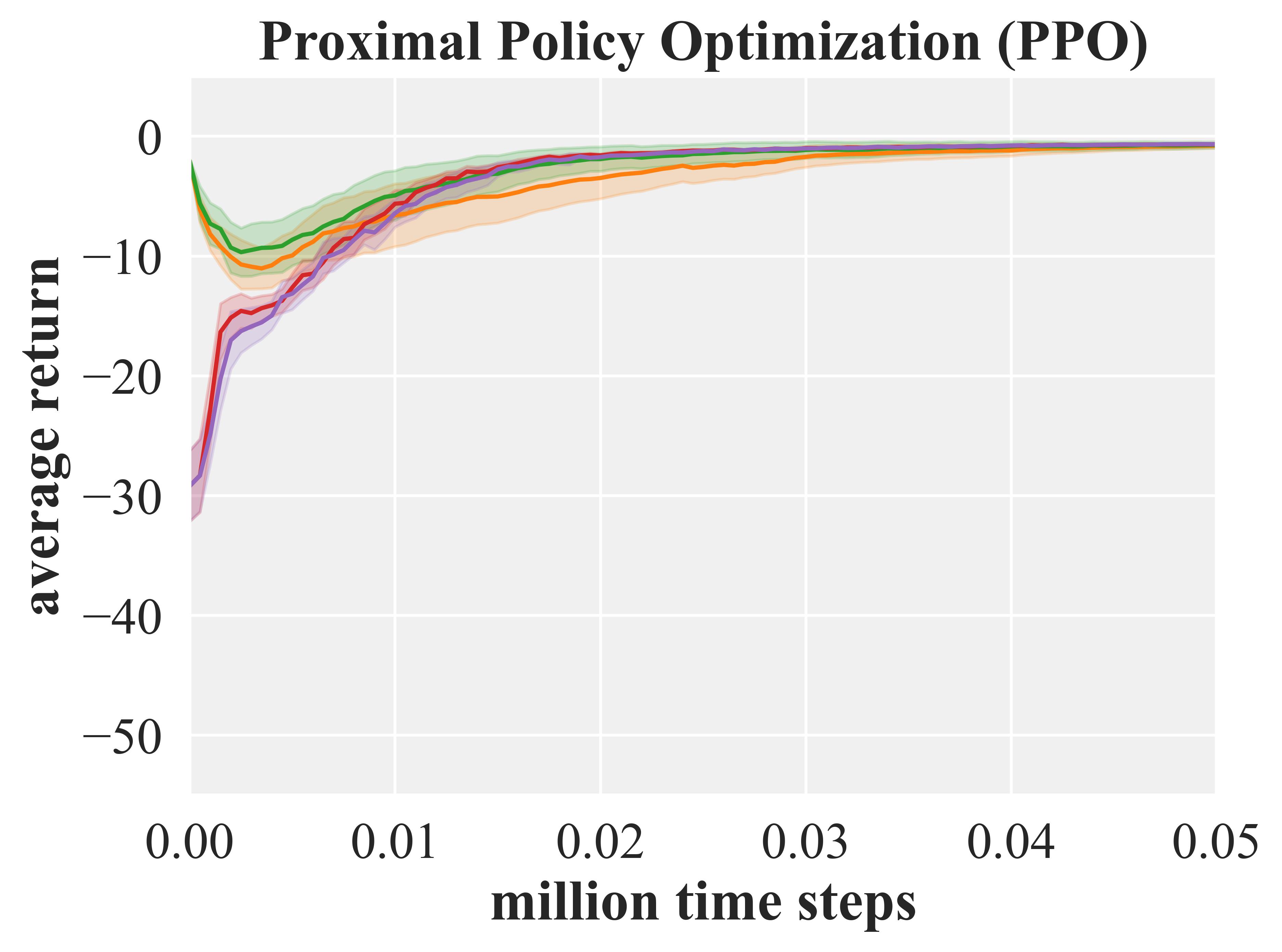}
        \includegraphics[width=0.48\linewidth]{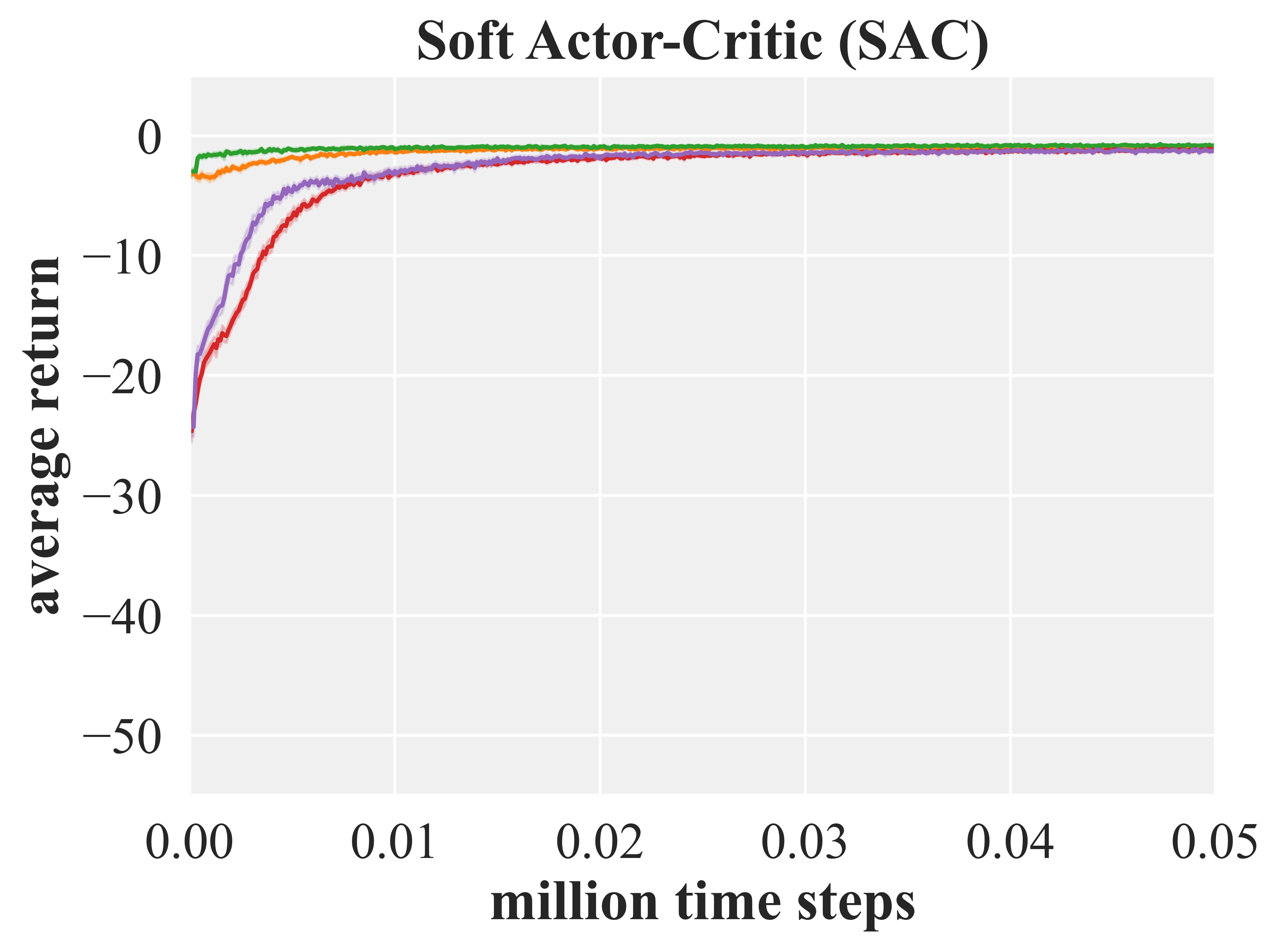}
        \caption{Slippery Elbow Flex Joint}
        \vspace{0.2cm}
        \label{fig:fetchreachf2}
    \end{subfigure}
    
    \captionsetup{justification=justified}
    \caption{Learning curves showing the average return of \gls{ppo} and \gls{sac} in the FetchReachDense-v3 Frozen Shoulder Lift Position Sensor and Slippery Elbow Flex Joint fault environments, evaluated under four knowledge transfer approaches. Results are averaged over 30 runs, with shaded regions indicating 95\% confidence intervals. Each subplot shows average return versus training steps (in millions), revealing differences in sample efficiency, adaptation speed, and robustness across algorithms and knowledge transfer approaches.}
    \label{fig:fetchreach_plot}
\end{figure*}

In the first comparative evaluation, the performance, sample efficiency, and real-time speed of the adaptation process with \gls{ppo} and \gls{sac} in the six fault environments are examined. Figures \ref{fig:ant_plot_1} and \ref{fig:ant_plot_2} illustrate the average return achieved by each of the four knowledge transfer options with respect to the number of real experiences (i.e., time steps) in the Ant-v5 fault environments. Figure \ref{fig:fetchreach_plot} presents analogous results for the FetchReachDense-v3 environments.

Figures \ref{fig:ant_plot_1} and \ref{fig:ant_plot_2} depict the real-time equivalent of 17.4 days (\gls{ppo}) and 1.7 days (\gls{sac}) of learning in the four Ant-v5 fault environments, while Figure \ref{fig:fetchreach_plot} depicts the real-time equivalent of 0.6 hours (\gls{ppo} and \gls{sac}) of learning in the two FetchReachDense-v3 fault environments. These real-time equivalents reflect simulation time and do not include additional computational overhead, such as policy inference, environment resets, logging, etc., which can increase the wall-clock runtime.

For the evaluation, the focus is on each algorithm's best knowledge transfer approach in each fault environment.
The best knowledge transfer approach for each algorithm is defined as the one exhibiting the highest performance throughout the entire learning process (depicted in Figures \ref{fig:ant_plot_1}--\ref{fig:fetchreach_plot}).
For example, in the Broken, Unsevered Limb fault environment shown in Figure \ref{fig:antf4}, the best knowledge transfer approach for \gls{ppo} is to retain the model parameters \(\theta_{PPO}\), noting that discarding or retaining the memory \(\mathcal{M}\) has little impact on performance.
In contrast, for \gls{sac}, the best knowledge transfer approach is to retain both the model parameters \(\theta_{SAC}\) and the replay buffer \(\mathcal{B}\). 

When comparing each algorithm's best knowledge transfer approach, \gls{ppo} generally achieves higher asymptotic returns than \gls{sac} in the four Ant-v5 fault environments (Figures \ref{fig:ant_plot_1} and \ref{fig:ant_plot_2}).
\gls{sac}, however, adapts more quickly than \gls{ppo} in real-time.
For instance, in the Ankle ROM Restriction fault environment (Figure \ref{fig:antf1}), \gls{sac}'s best knowledge transfer approach reaches a performance comparable to that of \gls{ppo} in 2,000,000 time steps (1.2 days), whereas \gls{ppo}'s best approach requires approximately 17,500,000 time steps (10.1 days) to achieve the same performance. 
These results highlight a key trade-off: \gls{ppo} is more effective for maximizing long-term performance, while \gls{sac} offers superior real-time adaptation efficiency.
This distinction reflects their algorithmic properties---\gls{sac}, as an off‑policy method, achieves greater sample efficiency, while \gls{ppo}'s on‑policy nature means it requires substantially more real experience.

In the low-dimensional FetchReachDense-v3 environments (Figure \ref{fig:fetchreach_plot}), the performance trend is mixed.
In the Frozen Shoulder Lift Position Sensor fault environment, \gls{ppo}'s best knowledge transfer approach reaches peak performance at 50,000 time steps (0.6 hours), while \gls{sac}'s best approach converges more quickly---within 20,000 time steps (0.2 hours)---but converges to a lower return.
For the Slippery Elbow Flex Joint fault, both algorithms' best transfer approaches yield similar asymptotic returns; however, \gls{sac} achieves this performance in fewer time steps.

Taken together, these findings suggest that \gls{ppo} excels at achieving strong final performance under most fault conditions, provided sufficient training time and an appropriate transfer strategy. In contrast, \gls{sac} is better suited for rapid adaptation.

\subsection{Transfer of Task Knowledge}
\label{sec:transfer_of_task_knowledge}

In the second comparative evaluation, the early performance of the four knowledge transfer approaches is investigated. The baseline (Approach 4) involves no knowledge transfer; tuples \(\{\theta_{PPO}, {\mathcal{M}}\}\) and \(\{\theta_{SAC}, {\mathcal{B}}\}\) are discarded for \gls{ppo} and \gls{sac}, respectively. Consequently, the policy is trained with data collected solely in each fault environment and is unbiased to the normal environment.

\begin{figure}[!ht]
    \centering
    \begin{subfigure}{\linewidth}
        \centering
        \includegraphics[width=0.7\linewidth]{images/legend2.jpg}
        \vspace{0.2cm}
    \end{subfigure}
    
    \begin{subfigure}{\linewidth}
        \centering
        \includegraphics[width=0.48\linewidth]{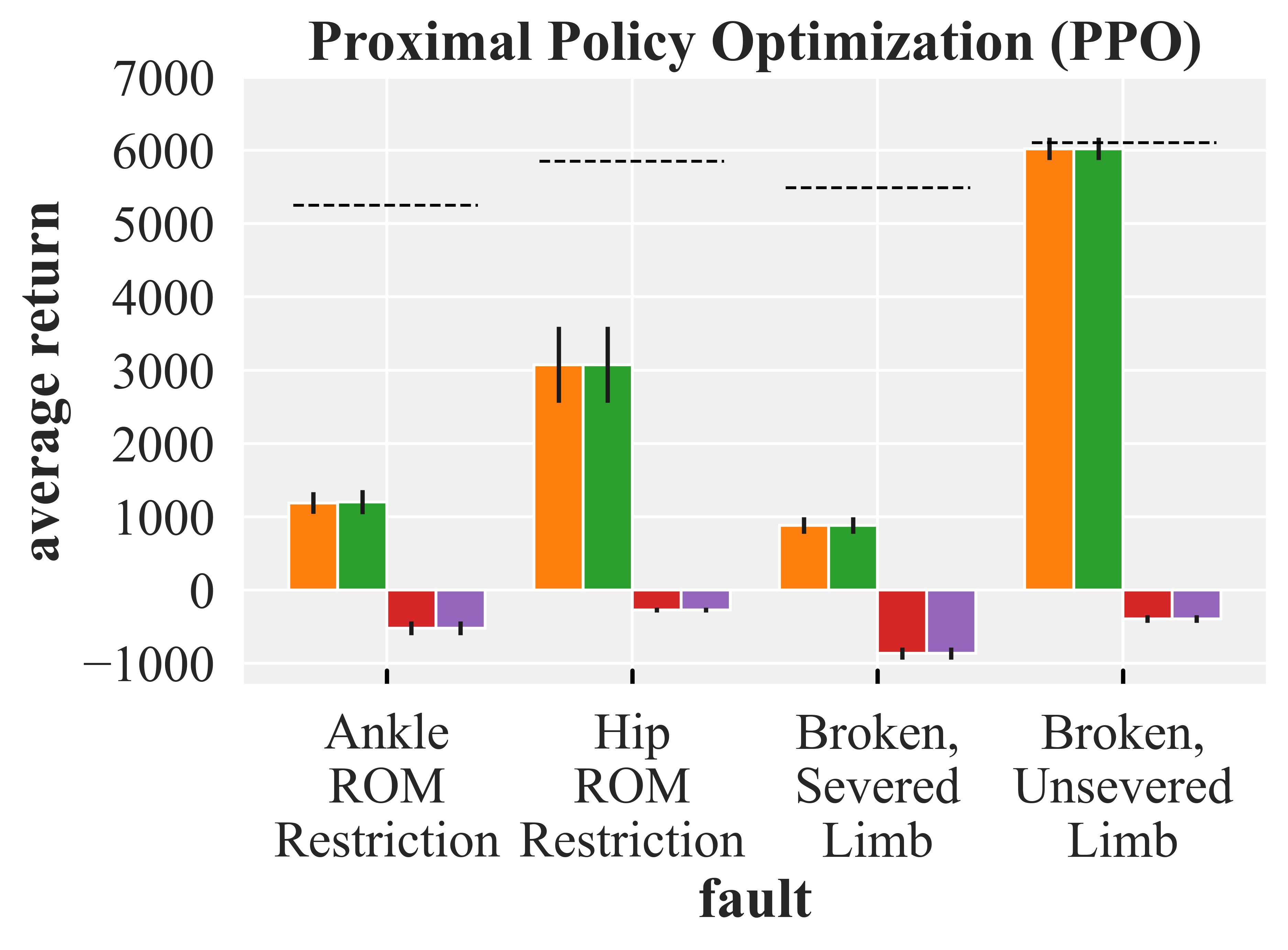}
        \includegraphics[width=0.48\linewidth]{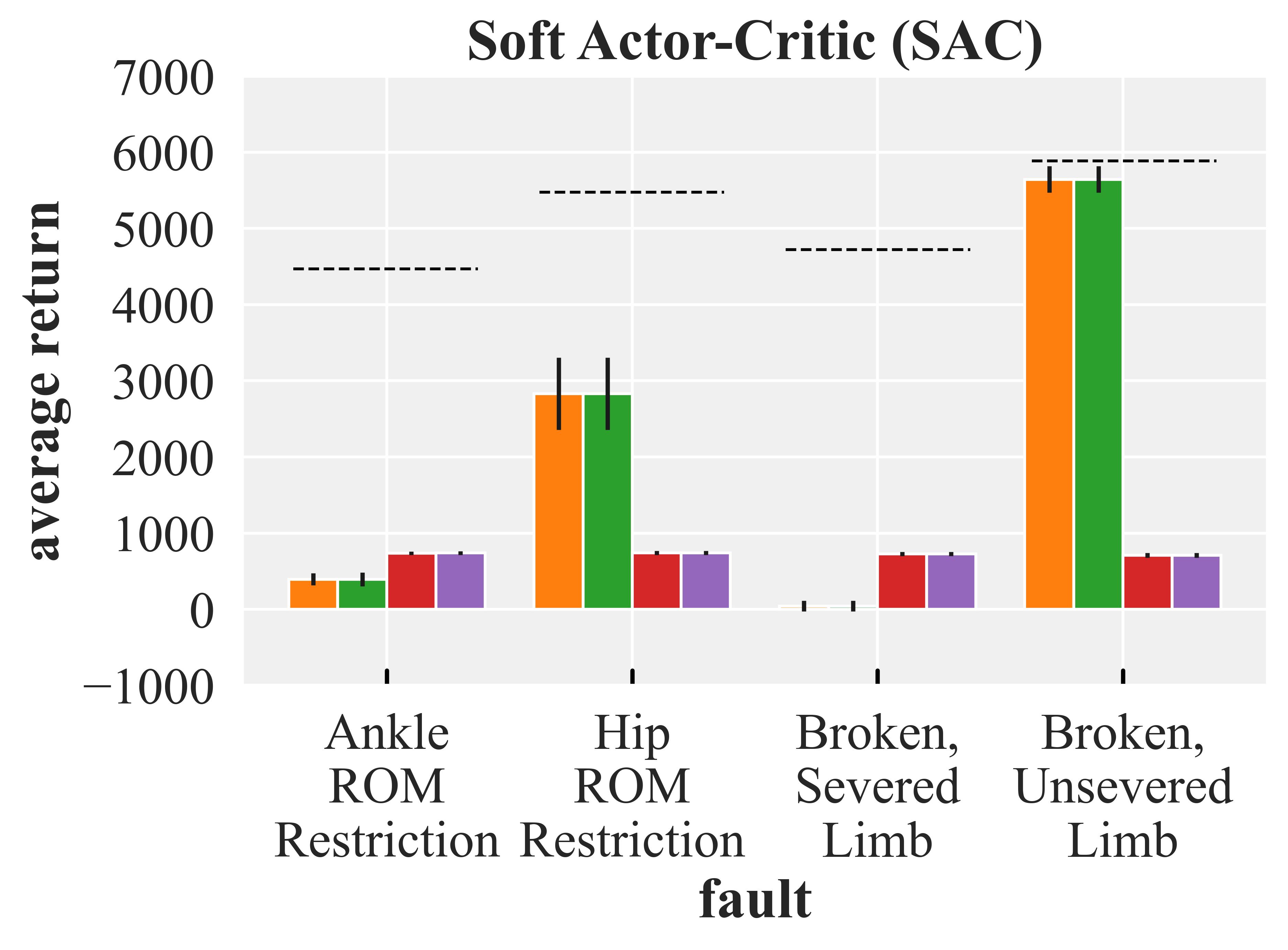}
        \caption{No Adaptation (0 Time Steps)}
        \vspace{0.2cm}
        \label{fig:ant_bar_plot_0k}
    \end{subfigure}
    \begin{subfigure}{\linewidth}
        \centering
        \includegraphics[width=0.48\linewidth]{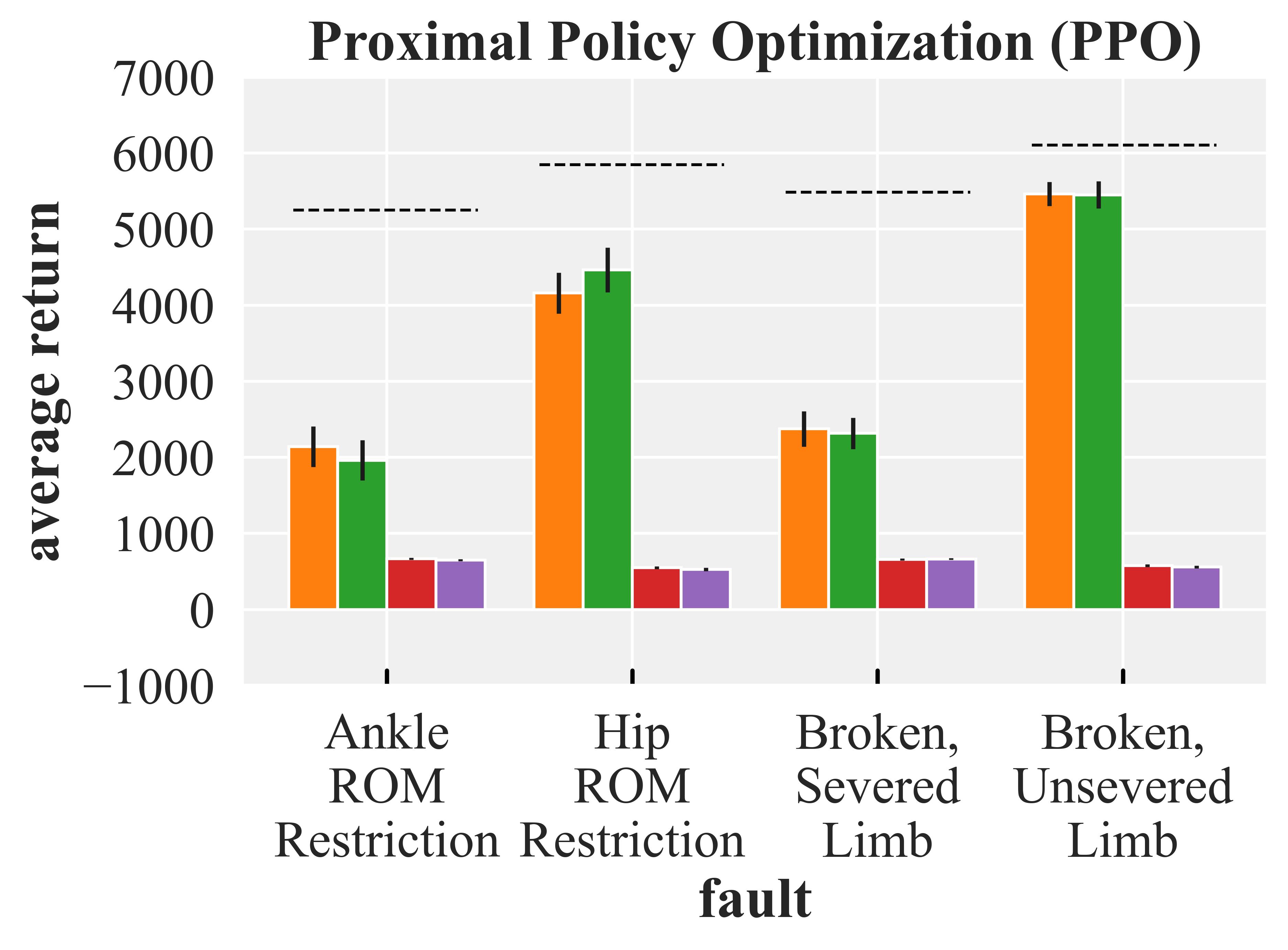}
        \includegraphics[width=0.48\linewidth]{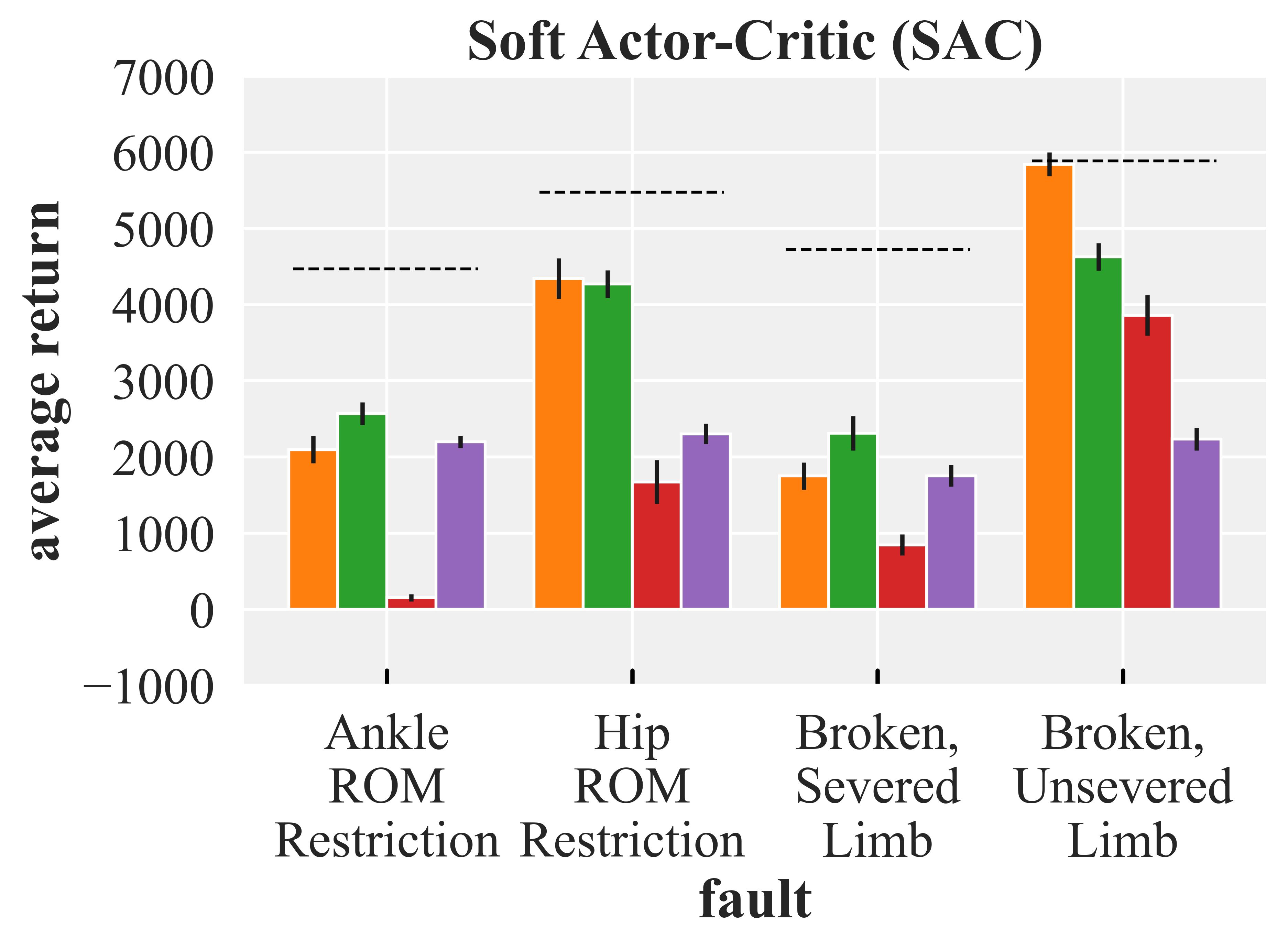}
        \caption{Partial Adaptation (300,000 Time Steps)}
        \vspace{0.2cm}
        \label{fig:ant_bar_plot_300k}
    \end{subfigure}
    
    \captionsetup{justification=justified}
    \caption{Early performance in the four Ant-v5 fault environments at two time points: (a) immediately after fault onset (0 time steps) and (b) after 300,000 time steps of adaptation. Each bar represents a different knowledge transfer approach. The black dashed line indicates the asymptotic performance of the baseline (Approach 4), which discards all prior knowledge. For \gls{ppo}, retaining the model parameters consistently yields the best early performance. For \gls{sac}, the optimal knowledge transfer approach depends on both the fault type and the amount of adaptation. With no adaptation, discarding all knowledge performs better for some faults, while retaining all knowledge is superior in others. With partial adaptation, retaining all knowledge is consistently superior to all other knowledge transfer approaches. These findings underscore that effective knowledge transfer strategies depend not only on the algorithm and fault type, but also on the degree of adaptation—highlighting the interaction between these factors in shaping early performance outcomes.}
    \label{fig:ant_bar_plot}
\end{figure}

\begin{figure}[!ht]
    \centering
    \begin{subfigure}{\linewidth}
        \centering
        \includegraphics[width=0.7\linewidth]{images/legend2.jpg}
        \vspace{0.2cm}
    \end{subfigure}
    
    \begin{subfigure}{\linewidth}
        \centering
        \includegraphics[width=0.48\linewidth]{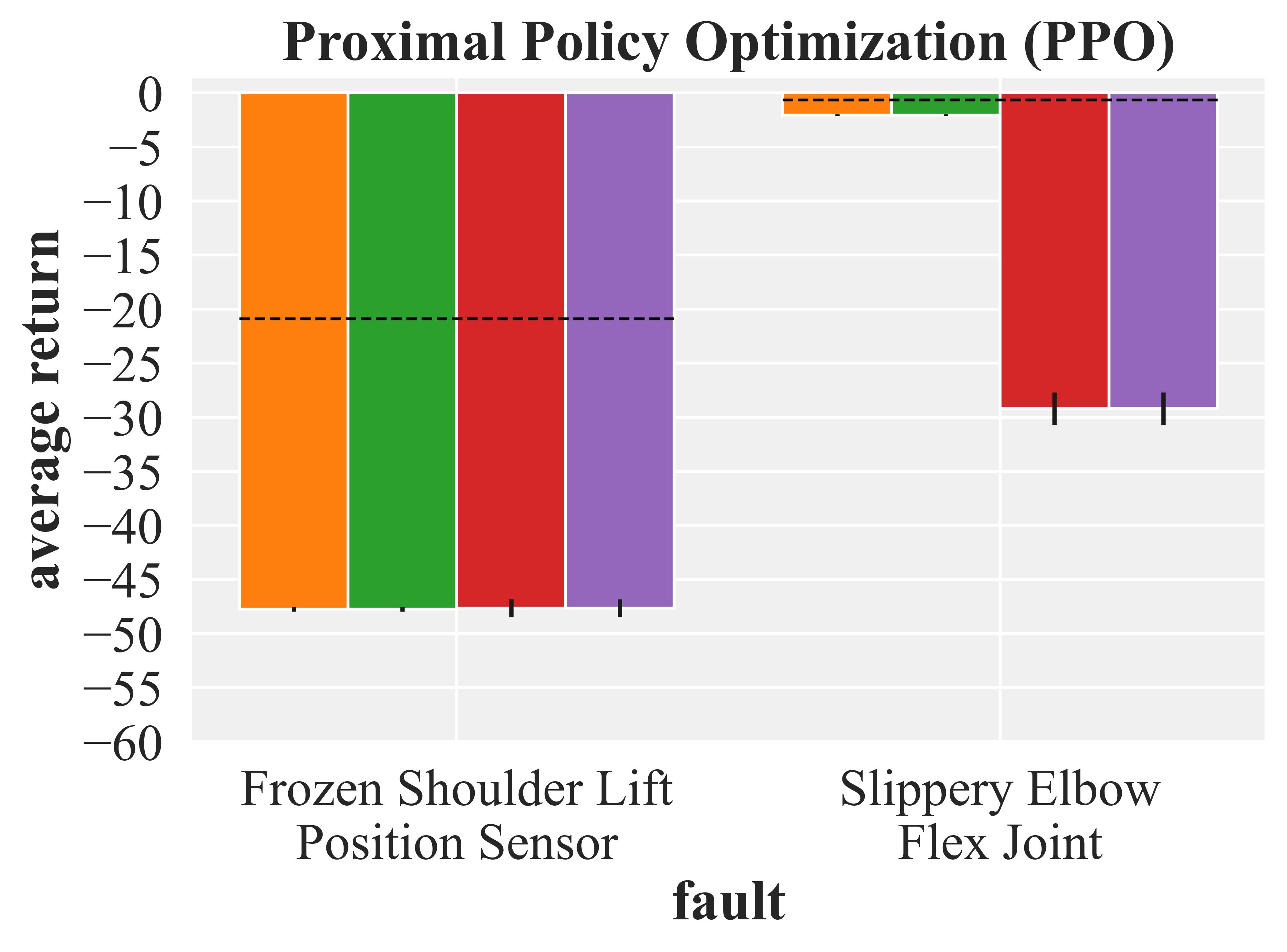}
        \includegraphics[width=0.48\linewidth]{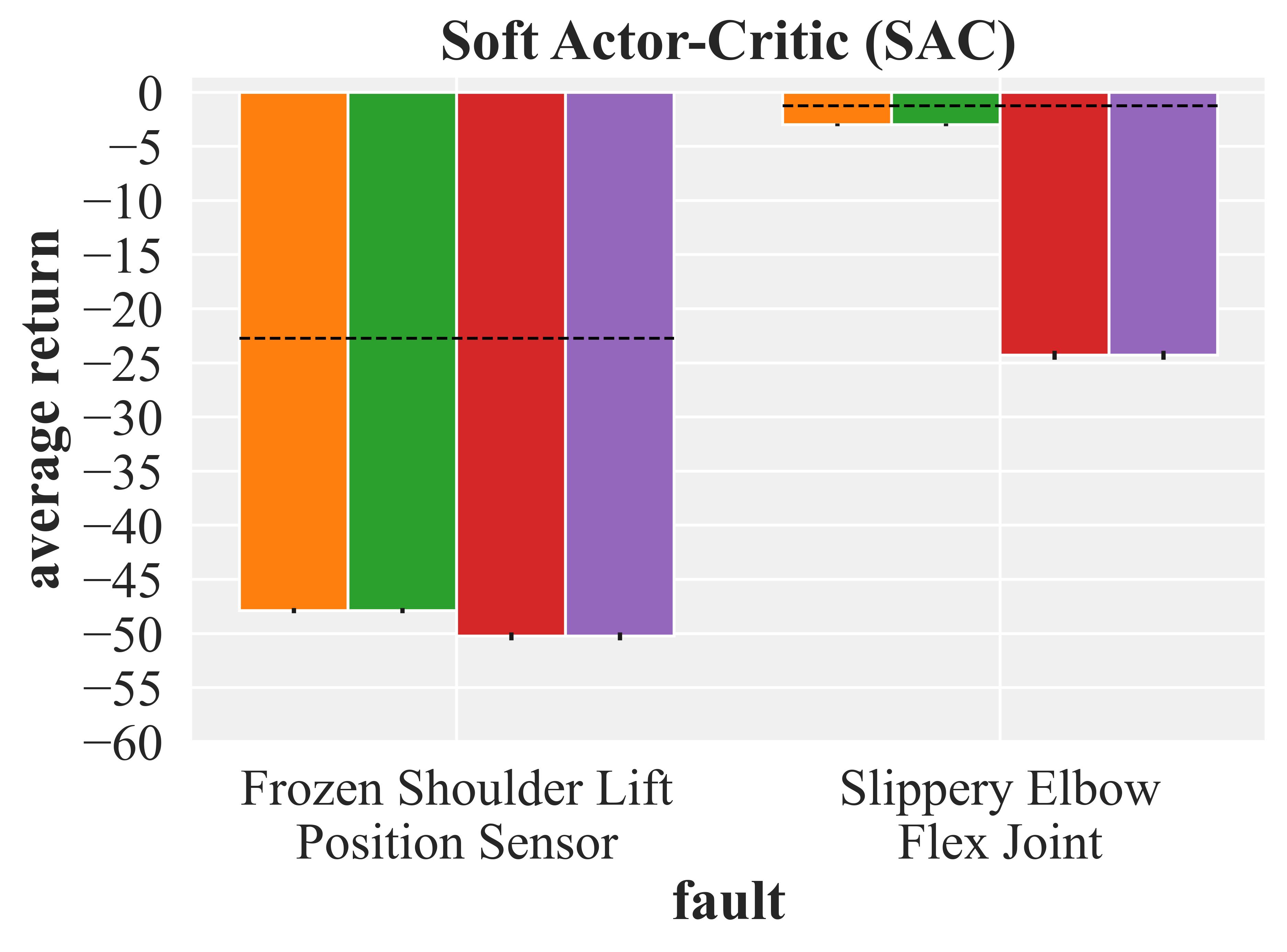}
        \caption{No Adaptation (0 Time Steps)}
        \vspace{0.2cm}
        \label{fig:fetchreach_bar_plot_0k}
    \end{subfigure}
    \begin{subfigure}{\linewidth}
        \centering
        \includegraphics[width=0.48\linewidth]{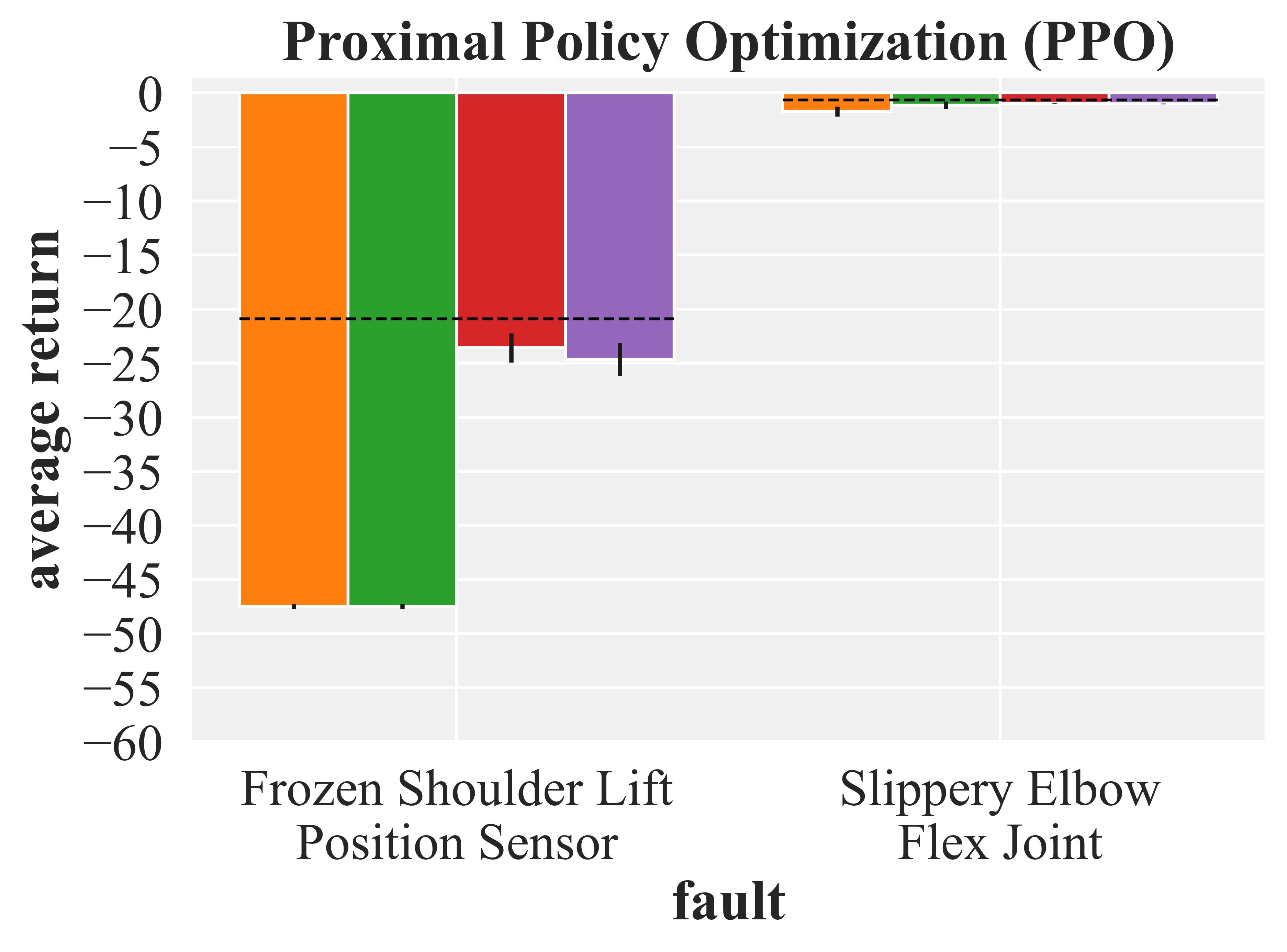}
        \includegraphics[width=0.48\linewidth]{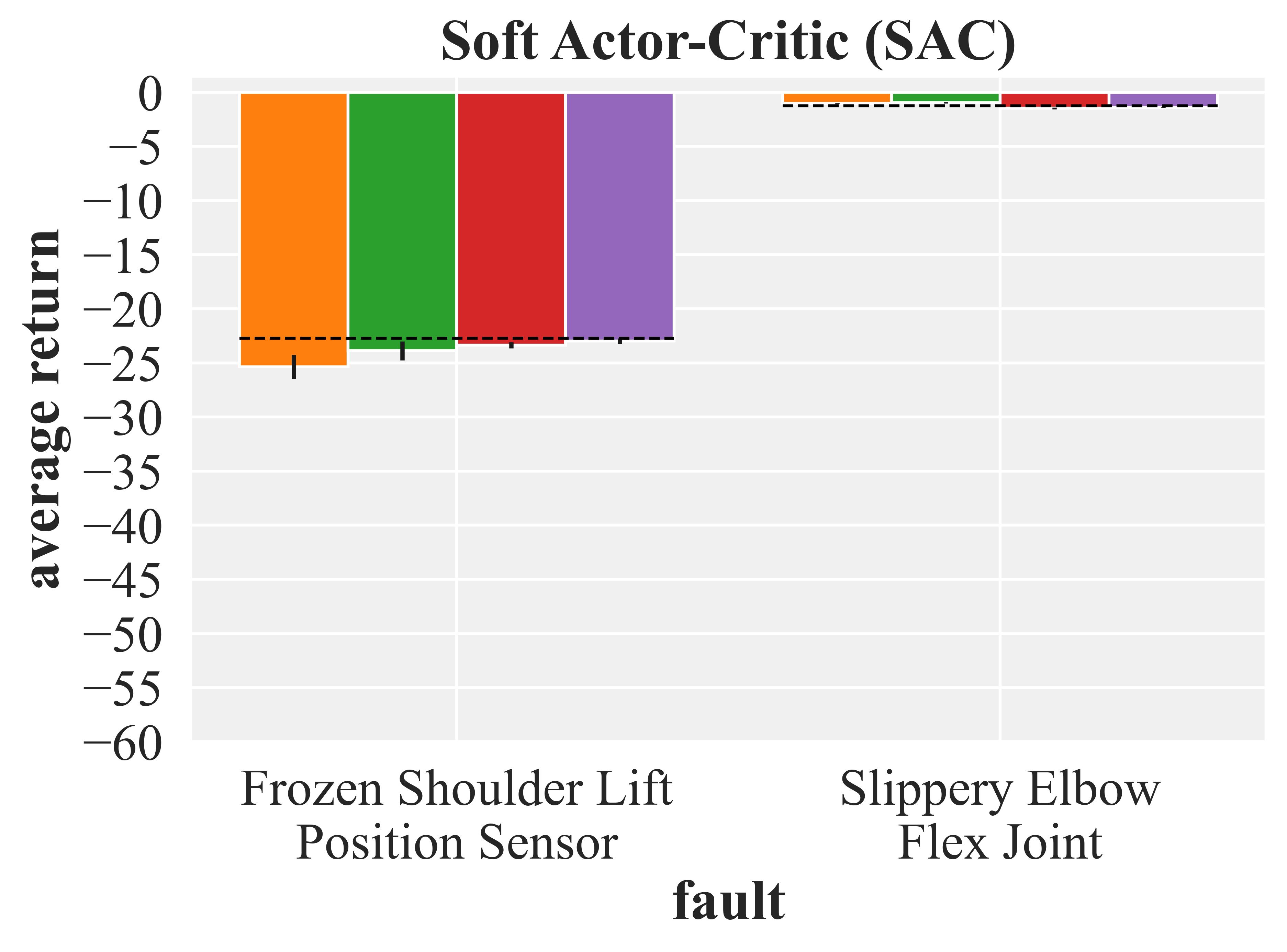}
        \caption{Partial Adaptation (30,000 Time Steps)}
        \vspace{0.2cm}
        \label{fig:fetchreach_bar_plot_30k}
    \end{subfigure}
    
    \captionsetup{justification=justified}
    \caption{Early performance in the two FetchReachDense-v3 fault environments at two time points: (a) immediately after fault onset (0 time steps) and (b) after 30,000 time steps of adaptation. Each bar represents a different knowledge transfer approach. The black dashed line indicates the asymptotic performance of the baseline (Approach 4), which discards all prior knowledge. \gls{ppo} achieves its best early performance in the Frozen Shoulder Lift Position Sensor fault when the model parameters \(\theta_{\text{PPO}}\) are discarded, whereas retaining \(\theta_{\text{PPO}}\) is more effective in the Slippery Elbow Flex Joint fault. In contrast, \gls{sac} consistently performs best when the model parameters \(\theta_{\text{SAC}}\) are retained across both fault types.}
    \label{fig:fetchreach_bar_plot}
\end{figure}

Figures \ref{fig:ant_bar_plot} and \ref{fig:fetchreach_bar_plot} show the early performance of each knowledge transfer approach in the Ant-v5 and FetchReachDense-v3 fault environments at two critical points in the learning process: (1) immediately after the onset of a fault, where knowledge from the normal environment has been transferred (and discarded, if applicable), but no learning in the fault environment has occurred; and (2) 300,000 time steps (Ant-v5) and 30,000 time steps (FetchReachDense-v3) after the onset of a fault, chosen to highlight the notable differences in early performance among the four knowledge transfer approaches in each fault environment. The asymptotic performance of the baseline in each fault environment is indicated by a black dashed line.

The results show that, with \gls{ppo}, retaining (and fine-tuning) the model parameters \(\theta_{PPO}\) generally leads to the best early performance in the fault environments, as the knowledge contained within the models facilitates rapid adaptation. 
However, the Frozen Shoulder Lift Position Sensor fault (depicted in Figure \ref{fig:fetchreach_bar_plot}) presents an exception, where \gls{ppo} initially struggles regardless of the knowledge transfer strategy.
Here, retaining and fine-tuning the \gls{ppo} parameters underperforms relative to approaches that discard and reinitialize the model parameters.
The Frozen Shoulder Lift Position Sensor fault is unique as it is the only fault that does not alter the dynamics of the environment. 
Rather, it introduces an observation error by consistently reporting an incorrect position for the Shoulder Lift joint, resulting in a mismatch between the agent's observation and the true state.

It is also observed that \gls{ppo}'s memory has minimal impact on performance in the fault environments, whether retained or discarded. In the experiments, \gls{ppo}'s memory capacity is small and all experiences are discarded from the memory after a single update. Consequently, the old experiences retained in the memory have minimal influence on adaptation. 

Furthermore, as shown in Figures \ref{fig:ant_plot_1}--\ref{fig:fetchreach_plot}, \gls{ppo} exhibits relatively little variability in asymptotic performance across the four knowledge transfer approaches in most fault environments.

\gls{sac} shows that its optimal early performance strategy depends on the fault type (Figures \ref{fig:ant_bar_plot} and \ref{fig:fetchreach_bar_plot}).
In the Ant-v5 Hip ROM Restriction and Broken, Unsevered Limb faults, retaining both the model parameters \(\theta_{\text{SAC}}\) and the replay buffer \(\mathcal{B}\) yields the highest return. 
In contrast, for the Ant-v5 Ankle ROM Restriction and Broken, Severed Limb faults, retaining both the model parameters \(\theta_{\text{SAC}}\) and discarding the replay buffer \(\mathcal{B}\) leads to the best early performance. 
These patterns indicate that, in the early phase, retaining the SAC parameters \(\theta_{\mathrm{SAC}}\) consistently benefits performance, while the optimal handling of the replay buffer \(\mathcal{B}\) (retain vs.\ discard) varies with the fault type.
An analogous pattern holds for the FetchReachDense‑v3 Slippery Elbow Flex Joint fault, where retaining \(\theta_{SAC}\) remains beneficial, while the choice to retain or discard \(\mathcal{B}\) is less pronounced in modulating early performance.
Similar to \gls{ppo}, the Frozen Shoulder Lift Position Sensor fault is an exception for \gls{sac}, where retaining acquired knowledge yields a slight performance gain with no adaptation (0 time steps) but a slight performance degradation after partial adaptation (30,000 time steps).

Figures~\ref{fig:ant_plot_1} and \ref{fig:ant_plot_2} further show that the asymptotic performance of \gls{sac} varies substantially across the four knowledge transfer approaches in the high-dimensional Ant-v5 fault environments.
Conversely, Figure~\ref{fig:fetchreach_plot} shows that in the low-dimensional FetchReachDense-v3 fault environments, \gls{sac}'s performance is less sensitive to the choice of knowledge transfer approach.
Together, these findings imply that \gls{sac}'s sensitivity to knowledge transfer scales with task complexity.

\subsection{Comparison With Prior Fault Adaptation Approaches}
\label{sec:comparison_with_prior_fault_adaptation_approaches}

In the third comparative evaluation, the performance of \gls{ppo} and \gls{sac} in the four Ant-v5 fault environments (shown in Figure \ref{fig:ant_bar_plot}) is compared to the performance of three meta-\gls{rl} algorithms in an \textit{Ant: crippled leg} environment \citep{nagabandi2018learning} and the performance of \gls{pdvf} in four \textit{Ant-Legs} environments \citep{raileanu2020fast}.

In the meta-\gls{rl} experiments conducted by \cite{nagabandi2018learning}, the early performance of the proposed meta-\gls{rl} algorithms, \gls{grbal} and \gls{rebal}, reached an average return of approximately 430 in the \textit{Ant: crippled leg} environment. Their performance curves appear to plateau around this value, though the authors do not report asymptotic values numerically. They did, however, report the asymptotic performance of a third meta-\gls{rl} algorithm, \gls{maml} \citep{finn2017model}, as approximately 710 in the same environment.

In the \gls{pdvf} experiments by \cite{raileanu2020fast}, \gls{pdvf} achieved returns ranging from approximately 200 to 350 across the four \textit{Ant-Legs} environments, consistently outperforming \gls{ppo}all. The latter, trained in a multi-task setting, reached returns in the range of roughly 120 to 320.
Meanwhile, \gls{ppo}env, trained exclusively in the test environment, achieved a performance in the range of 374 to 862, consistently outperforming \gls{pdvf}.

In the early adaptation phase within the Ant-v5 fault environments, both \gls{ppo} and \gls{sac} outperform previously proposed methods, as summarized in Tables \ref{tab:ant_no_adaptation} and \ref{tab:ant_early_adaptation}.
Performance, however, varies considerably depending on the specific fault and whether storage is retained or discarded.
For each fault, \gls{ppo}'s optimal knowledge transfer approach---retaining \tppo---yields mean returns of 811--6167 with no adaptation (0 time steps) and 2139--5463 after partial adaptation (300,000 time steps). 
Similarly, with partial adaptation (300,000 time steps), \gls{sac}'s best transfer approach---retaining \tsac---achieves returns of 2312--5841.
Notably, \gls{ppo}, with no adaptation, attains a performance approximately 1.9--14.3 times higher than that reported for \gls{grbal} and \gls{rebal}, and approximately 1.1--8.7 times higher than the asymptotic performance of \gls{maml}.
After partial adaptation (300,000 time steps), \gls{ppo} achieves a performance approximately 5.0--12.7 times higher than the performance reported for \gls{grbal} and \gls{rebal}, and approximately 3.0--7.7 times higher than the asymptotic performance reported for \gls{maml}. 
Similarly, after partial adaptation, \gls{sac} achieves an early performance 5.4 to 13.6 times higher than \gls{grbal} and \gls{rebal}, and 3.3 to 8.2 times higher than the asymptotic performance of \gls{maml}. 
Furthermore, \gls{ppo} nearly matches or outperforms \gls{ppo}env.
It is worth noting that other factors, such as the implementation of \gls{ppo} and the disparity between the fault environments, may contribute to these differences in performance.

While a more in-depth comparison of these methods is required, the preliminary comparative evaluation indicates that \gls{ppo} and \gls{sac} achieve competitive performance compared to previous related work,
especially when an appropriate knowledge transfer method is chosen. 
These results underscore the promise of integrating advanced knowledge transfer strategies into off‑the‑shelf \gls{rl} algorithms for more effective machine fault adaptation.

\section{Discussion}
\label{sec:discussion}

This paper investigates the effectiveness of transferring knowledge acquired by an \gls{rl} agent in a normal environment---where the agent learns a task with a fully functioning robot model---to a fault environment, where the agent must adapt to perform the same task as the robot model experiences a hardware fault. The discussion begins with Ant-v5 and later transitions to FetchReachDense-v3.

\subsection{Ant-v5}
\label{sec:discussion_ant}

Ant-v5, introduced in Section \ref{sec:ant_v2}, is a Gymnasium environment where an eight-joint quadruped must maximize its forward velocity. Its high-dimen-\\sional action space and complex joint coordination make it a classic \gls{rl} benchmark.

In Ant-v5, four actuator faults were introduced, each weakening a different joint to a different degree: the \hrom fault restricted hip motion to $10^\circ$, permitting force only within a small range; the \bul fault shortened the lower leg link and inserted a passive ball joint, followed by a section identical to the removed portion, resulting in motors remaining connected to the ground through that link; the \arom fault confined the ankle to $5^\circ$, essentially locking it; and the \bsl fault removed half the leg, preventing ankle torques from reaching the ground.
 
Our results for Ant-v5 are shown in Figures \ref{fig:ant_plot_1}, \ref{fig:ant_plot_2}, and \ref{fig:ant_bar_plot}, as well as Tables \ref{tab:ant_no_adaptation}--\ref{tab:ant_full_adaptation}.

\paragraph{No adaptation} 

At $t=0$, no learning occurs. For \sac, decisions come from the stochastic, high-entropy policy parameterized by \tsac, and the replay buffer \buff has no effect.
Consequently, performance across the four Ant-v5 fault environments is driven primarily by whether \tsac is retained or discarded.
\sac results in better performance (i.e., higher returns) when the actuators on which the policy relies remain usable---\hrom ($2830.99\pm472.29$) and \bul ($5642.99\pm172.58$); discarding \tsac with these faults produces returns below $800$.
When a critical actuator is disabled, the high-entropy policy continues to allocate actions to that joint, yielding returns below $400$. Reinitializing \tsac produces a new, unbiased policy that redistributes effort across all the actuators, resulting in higher returns---\arom ($738.39\pm22.52$) and \bsl ($727.24\pm25.64$).

\ppo is trained with little entropy regularization; \tppo parameterizes a policy that is nearly deterministic. 
At $t = 0$, the replay memory \mem has had no influence on the policy, and performance relies solely on the model parameters $\theta_{\text{PPO}}$.
Retaining $\theta_{\text{PPO}}$ results in higher performance across all four faults---\arom ($1200.10\pm165.28$), \hrom ($3075.30\pm520.16$), \bsl ($811.28\pm98.02$), and \bul ($6167.02\pm128.25$).
Discarding (reinitializing) \tppo produces a Gaussian policy with near-zero means, yielding behaviour that is effectively random and results in returns below $-300$.

In conclusion, at $t=0$, \sacs high-entropy policy fails when critical joints are disabled, recovering only after reinitializing \tsac, whereas \ppos low-entropy, near-deterministic policy proves more robust.

\paragraph{Partial adaptation} 

After 300,000 time steps, some adaptation (retraining) has occurred. As a result, the retained model parameters $\theta$ and retained experiences in \mem/\buff have influenced the policy; retained experiences in \mem/\buff have been partially (\sac) or fully replaced (\ppo).

At this stage, \sac's performance remains strongly dependent on whether the model parameters \tsac are retained or discarded, whereas the effect of the replay buffer \buff is less clear---confidence intervals overlap across conditions, indicating fault-specific differences that are not consistently significant.
Retaining \tsac, while discarding \buff, yields the highest returns in the \arom ($2569.49\pm148.15$) and \bsl ($2312.05\pm225.16$) faults, suggesting that fresh experiences update the policy in a manner that reduces reliance on disabled joints while still leveraging the benefit of the retained model parameters \tsac. 
In contrast, retaining both \tsac and \buff is most effective for the \hrom ($4343.12\pm267.30$) and \bul ($5841.98\pm155.97$) faults, where the stored transitions remain valid and accelerate policy adaptation. 
Dropping \tsac generally harms performance (returns $<2350$); the exceptions are the \arom ($2197.15\pm77.31$) and \bsl ($1754.01\pm141.59$) faults when both \tsac and \buff are reset.

For \ppo, the retained model parameters \tppo remain influential; retaining them yields observed mean returns that are 2--9 times higher than when reinitializing. 
The influence of the memory \mem is small and fault-specific. Discarding \mem slightly helps the \hrom ($4463.93\pm295.02$) fault, likely because outdated trajectories no longer bias the value estimates, whereas retaining \mem yields the best or near-best scores in the \arom ($2139.18\pm266.93$), \bsl ($2374.34\pm231.94$), and \bul ($5463.51\pm158.17$) faults. 
Resetting \tppo consistently leads to poor outcomes (returns below 670), suggesting that the nearly deterministic gait encoded in \tppo appears to provide the strongest starting point, even after partial adaptation.

In summary, early adaptation benefits both algorithms in different ways. \sac gains most by pairing its high-entropy policy with a fresh buffer when old experiences clash with new dynamics. In contrast, \ppos remains competitive as long as the model parameters \tppo are retained, with the memory \mem playing a minor role.

\paragraph{Full adaptation}

After each agent has fully adapted in the fault environments, both the agent's policy and the stored experiences \mem/\buff have been updated for millions of steps; no retained experiences remain in \mem/\buff.

For \sac, retaining the model parameters \tsac corresponds to the highest observed mean returns across all faults, regardless of whether the replay buffer \buff is retained---\arom ($5105.93\pm103.59$), \hrom ($5679.06\pm161.71$ when \buff is discarded), \bsl ($5514.13\pm103.45$), and \bul ($6390.86\pm121.99$). Discarding \tsac still comes with substantial risk. It shows markedly lower mean returns in the \arom and \hrom faults unless \buff is also discarded, and even then the means remain below those obtained with the best knowledge transfer approach (retaining \tsac).

For \ppo, extended training largely removes the influence of retained model parameters $\theta_{PPO}$.
All four transfer options converge to high returns, and in two faults the highest observed mean returns occur after discarding the model parameters $\theta_{PPO}$---\hrom ($5965.3\pm238.6$) and \bul ($6103.0\pm214.6$).
The memory \mem remains a minor, fault-specific influence: discarding it slightly helps the mean return in the \bsl ($5919.9\pm115.9$) fault but has little effect elsewhere.

After millions of training steps, \sac shows the highest observed mean returns when retaining the model parameters \tsac across all four faults. In contrast, \ppo shows only small differences across knowledge transfer options, as all approaches converge to similarly high returns.

\subsection{FetchReachDense-v3}
\label{sec:discussion_fetchreach}

FetchReachDense-v3 is a Gymnasium manipulation task in which a seven-joint robotic arm must guide its gripper to randomly placed 3-D targets, receiving dense rewards for speed and precision. In this environment, two distinct faults were examined: a \sens fault that freezes the shoulder lift position sensor at $-1.5$ radians, providing the agent with incorrect state information while leaving the underlying mechanics intact, and a \slip fault that causes every elbow-flex joint torque command to overshoot by $0.05$ radians, simulating gear slippage.

Our FetchReachDense-v3 results appear in Figures \ref{fig:fetchreach_plot} and \ref{fig:fetchreach_bar_plot}, as well as Tables \ref{tab:fetchreach_no_adaptation}--\ref{tab:fetchreach_full_adaptation}.

\paragraph{No adaptation} 

At $t = 0$, no adaptation (learning) has occurred; each agent executes its policy parameterized by the model weights $\theta$, whether retained or reinitialized; any retained experiences in storage \mem/$\mathcal{B}$ have had no influence on the agent's policy.

With \sac, retaining \tsac yields higher returns for both the \slip fault ($-2.99\pm0.12$) and the \sens fault ($-47.88\pm0.24$), whereas discarding \tsac reduces returns for the \slip fault ($-24.29\pm0.41$) and the \sens fault ($-50.25\pm0.37$).

For \ppo, retaining \tppo yields higher mean returns in the \slip fault ($-2.04\pm0.08$) compared to discarding \tppo ($-29.20\pm1.51$).
In the \sens fault, mean performance stays consistently near $-47.7$ regardless of whether \tppo is retained or discarded.

In summary, at $t=0$ (no adaptation), retained model parameters $\theta$ only help when the fault is an actuation error; if the observations themselves are corrupted, the retained parameters no longer provide an advantage.

\paragraph{Partial Adaptation}

By 30,000 time steps, each agent has begun to update its policy parameters.
If retained, the memory \mem contents have been entirely replaced with new experiences and the replay buffer \buff contents have only partially been replaced ($|\mathcal{B}|$ = 100,000).

For \sac, in the \sens fault, all four knowledge transfer approaches cluster around returns of $-23$, with the highest mean return obtained by discarding both \tsac and \buff ($-22.98\pm0.27$); retaining either element shifts the return by $-1$ to $-2$. 
For the \slip fault, the highest mean return is attained by retaining \tsac and discarding \buff ($-0.97\pm0.07$). 
Discarding \tsac drops the return to $-1.39\pm0.06$ if \buff is discarded.

For \ppo, the effect of retaining \tppo differs by fault. 
For the \sens fault, discarding \tppo improves the return from $-47.50\pm0.23$ (\tppo retained) to $-23.58\pm1.36$ (\tppo discarded with \mem retained), with similar returns when \mem is discarded ($-24.65\pm1.52$).
The small memory \mem continues to have little influence.
For the \slip fault, all four knowledge transfer approaches converge near-optimal returns between $-1.72$ and $-0.98$, with the highest mean return coming from discarding \tppo and retaining \mem ($-0.98\pm0.04$). 
The remaining three approaches fall within $0.7$ of this performance, indicating minimal differences across all four knowledge transfer approaches.

In summary, \ppo gains most by reinitializing \tppo when the sensor is faulty (\sens); however, when under actuator error (\slip), \ppo achieves near-equivalent returns across all four knowledge transfer options.
In contrast, \sac benefits most from retaining $\theta_{SAC}$, with the best mean return for the \slip fault when \buff is discarded, while in the \sens fault, all knowledge transfer approaches yield similar returns.

\paragraph{Full Adaptation}

In the FetchReachDense-v3 fault environments, after 50,000 time steps (full adaptation), both agents have rebuilt their experience stores---\mem is filled with entirely new experiences and \buff is half refreshed.

For \sac, retaining \tsac yields higher mean returns for the actuator fault (\slip) but not for the sensor fault (\sens).
In \sens, every combination of \tsac and \buff hovers around $-23$, showing the four knowledge transfer approaches have little impact on full adaptation performance.
In \slip, retaining \tsac yields higher mean returns; retaining \tsac and discarding \buff results in the highest return of $-0.76\pm0.04$, whereas discarding \tsac drops the performance to roughly $-1.26\pm0.06$. 

For \ppo, the value of retaining the model parameters \tppo depends on the fault.
For the \sens fault, reinitializing \tppo improves performance, increasing the mean return from $-47.32\pm0.24$ to $-19.96\pm1.07$; \mem has little effect.
For the \slip fault, all four knowledge transfer approaches converge at -0.66 to -0.78, with the best performance coming from discarding both \tppo and \mem ($-0.66\pm0.02$). 

At full adaptation, \ppo shows higher mean returns by reinitializing \tppo when the observations are corrupted but not when only the dynamics shift, while \sac shows higher mean returns when retaining \tsac in the actuator fault, with \buff influencing performance marginally.

\subsection{Prior Fault Adaptation Approaches}

\label{sec:prior_fault_adaptation_approaches}

The results presented in Section~\ref{sec:prior_fault_adaptation_approaches} indicate that \ppo and \sac, when combined with appropriate knowledge transfer strategies, achieve strong early performance relative to both model-based meta-\gls{rl} approaches, namely \gls{grbal} and \gls{rebal}~\citep{nagabandi2018learning}, and the policy-dynamics value function approach, \gls{pdvf}~\citep{raileanu2020fast}, on comparable Ant tasks, although these comparisons should be interpreted with caution because the evaluated studies differ in their environments, fault settings, and experimental protocols.

The \gls{pdvf} framework~\citep{raileanu2020fast} is designed to enable rapid adaptation to environments with previously unseen dynamics. The approach learns a representation of both policies and environment dynamics, and at test time infers a dynamics embedding from a small number of interactions in a new environment. This embedding is then used, together with a learned value function, to identify an effective policy representation for the remainder of the episode. Consequently, adaptation in \gls{pdvf} occurs primarily at the beginning of the episode, rather than through continued policy updates. In contrast, \ppo and \sac refine a transferred policy throughout interaction with a fault environment, allowing adaptation to proceed continuously after fault onset.

More broadly, \gls{pdvf} can be viewed as representing a class of approaches that rely on a learned space of policies and environment dynamics to facilitate rapid policy selection in new settings. Such methods can be highly effective when the test environment lies within the variability captured during training. However, their performance is inherently constrained by the coverage of the training environments and the expressiveness of the learned policy representations. This limitation is particularly relevant in fault adaptation scenarios, where faults may vary substantially in type and severity.

\citet{raileanu2020fast} evaluated \gls{pdvf} under variations in physical parameters such as wind direction, fluid current, and limb lengths across MuJoCo domains including Spaceship, Swimmer, Ant-wind, and Ant-legs. In contrast, this study examines more real-world fault types, including range-of-motion restrictions, severed and unsevered limbs, frozen sensors, and actuator slippage. As such, the extent to which \gls{pdvf} generalizes to these more realistic and diverse fault conditions remains an open question. Furthermore, \gls{pdvf} requires training an ensemble of policies across multiple environments, along with separate policy and dynamics embedding models and a value function over these embeddings. Consequently, its training complexity grows with the size and diversity of the policy set and the number of environment variants considered during training.

Model-based meta-\gls{rl} approaches, such as \gls{grbal} and \gls{rebal}, address adaptation through a different mechanism. These methods meta-train a dynamics model to support rapid online adaptation from a small amount of recent experience~\citep{nagabandi2018learning}. This formulation allows the model to quickly adjust to changes in environment dynamics, as demonstrated in settings involving disturbances such as leg failures and terrain variations. However, this efficiency depends on the assumption that the distribution of tasks encountered during meta-training adequately captures the variability present at test time. In fault adaptation settings, where failures may be unpredictable and highly diverse, this assumption may not hold, potentially limiting performance when encountering out-of-distribution faults.

In contrast, the model-free algorithms considered in this study, \ppo and \sac, do not rely on an explicit dynamics model or a predefined task distribution for adaptation. Instead, a policy learned in a normal environment is transferred and subsequently refined through continued interaction with the fault environment. While this approach is generally less sample-efficient than methods explicitly designed for rapid adaptation, it provides greater flexibility in handling unforeseen faults. Empirically, this flexibility is reflected in the results, where transferred policies exhibit strong initial performance and continue to improve with additional experience.

Taken together, these comparisons highlight a key distinguishing feature of the framework studied here: fault adaptation is formulated as a continual \gls{rl} problem. Standard policy gradient algorithms, \ppo and \sac, are used both to learn an initial policy under normal conditions and to update that policy online following fault onset. Unlike approaches that depend on a predefined task distribution or a learned set of policy representations spanning anticipated variations, the present framework adapts directly through interaction with the fault environment. The trade-off is that adaptation may require more real experience than methods specifically designed for rapid adaptation. Nevertheless, the results indicate that this trade-off is favourable in practice, with \sac providing efficient early adaptation and \ppo achieving strong asymptotic performance.

\section{Future Work}
\label{sec:future_work}

Building on these insights, future research will further explore ways to harness the potential of knowledge transfer methods.
The aim is to develop more advanced, generalized knowledge transfer methods by considering the characteristics of environments and faults.
The goal is to investigate how these characteristics can guide and enhance the design of knowledge transfer methods.
Following the methodology proposed by \cite{chen2023adapt}, such methods would selectively transfer the most useful knowledge, determining what to transfer based on similarities between the current environment and those previously encountered. 
For example, if an agent encounters an environment (e.g., fault) similar to one it previously adapted to, transferring the specific knowledge that proved effective may be more beneficial than transferring general knowledge.
This approach may lead to more consistent performance across various scenarios.

Ensuring safety during and after adaptation to a fault environment remains essential.
Current methods generally do not guarantee safe behaviour, especially when encountering unforeseen faults.
Thus, learning in safety-critical environments remains an open research avenue, as highlighted by several works on safe adaptation \citep{zhang2020cautious,peng2021safe}.
While this work focuses on improving state-of-the-art \rl algorithms through knowledge transfer, the importance of safe adaptation in robotic applications is recognized.
Future work will centre on safety-aware fault adaptation in settings involving other agents, fragile objects, and potential self-hazard scenarios.
This work is currently being ported to a six-degrees-of-freedom robotic arm---planned hardware trials will quantify adaptation latency, safety-critical behaviour, and control-bandwidth limits, with results to be reported separately.

Another avenue of future work is integrating \rl with broader \ai control paradigms.
Hybrid approaches that combine policy gradient methods with symbolic reasoning, model-based planning, or neuro-symbolic control may further enhance fault tolerance and generalization across diverse environments.
Such integration could provide mechanisms for embedding domain knowledge, logical constraints, or predictive models alongside adaptive learning, thereby increasing both robustness and interpretability.
Nevertheless, the primary focus of this work remains on policy gradient-based fault adaptation, owing to its compatibility with real-time robotic control and its demonstrated ability to embed fault-tolerant behaviour directly into policies.

\section{Conclusion}
\label{sec:conclusion}

This study demonstrates the viability of \glsentryfull{rl} algorithms as a practical approach to enhancing hardware fault tolerance in real-world machines. By systematically comparing \glsentryfull{ppo} and \glsentryfull{sac} under multiple actuator and sensor fault conditions, the results show that both algorithms can deliver robust and rapid recovery, reducing the need for costly hardware redundancy. This approach can mitigate productivity losses in resource-constrained or time-sensitive domains, where even minor fault-related events are costly.

Quantitatively, \ppo demonstrated strong asymptotic performance in high-dimensional tasks such as Ant-v5, achieving mean returns as high as $6167$ when the model parameters were retained, whereas discarding the model parameters reduced performance to below $-300$. In contrast, \sac achieved lower peak returns but adapted approximately eight times faster in real-time ($1.2$ vs.\ $10.1$ days), revealing a trade-off between speed and stability.

In the low-dimensional FetchReachDense-v3 environment, performance was fault-specific. When the position sensor was corrupted, discarding \ppos model parameters improved performance substantially, with returns roughly doubling compared to retaining the model parameters. Under actuator slip, however, \sac retained a clear speed advantage, converging within $0.2$ hours compared to \ppos $0.6$ hours. Taken together, these findings confirm that optimal knowledge transfer strategies depend on both the fault type and the algorithm.

When benchmarked against prior methods, including meta-\rl algorithms (GrBAL, ReBAL, MAML) and PD-VF, both \ppo and \sac achieved up to 5--13 times higher early adaptation returns, depending on the fault type and stage of adaptation. This demonstrates that integrating knowledge transfer strategies into off-the-shelf \rl algorithms can outperform specialized approaches.

Overall, these findings underscore that \rl offers a quantitative, scalable, and practical path to embedding fault tolerance directly into machine control. \ppo excels in maximizing long-term performance, \sac excels in rapid real-time recovery, and together they provide complementary strategies for future industrial deployment.

\section*{Acknowledgements}

We express our sincere gratitude to Mitsubishi Electric Corporation for their generous support, which made this work possible. Their assistance has been invaluable, and we are deeply grateful for the insightful feedback and constructive input they have provided throughout the project.

\appendix

\section{Evaluation Statistics}
\label{sec:evaluation_statistics}

\begin{table}[!h]
    \centering
    \scriptsize
    \begin{tabular}{>{\centering\arraybackslash}p{2.1cm} >{\centering\arraybackslash}p{1.4cm} >{\centering\arraybackslash}p{3.5cm} ccc}
        \toprule
        \textbf{Fault} & \textbf{Algorithm} & \textbf{Transfer Setting} & \textbf{Mean} & \textbf{SEM} & \textbf{CI [L, U]} \\

        \midrule

        \multirow{8}{*}{\makecell{Ankle ROM\\Restriction}}
        & \multirow{4}{*}{PPO}
        & retain $\theta_{PPO}$, retain $\mathcal{M}$
        & 1200.10
        & 165.28
        & [876.15, 1524.05] \\

        &
        & retain $\theta_{PPO}$, discard $\mathcal{M}$
        & 1200.10
        & 165.28
        & [876.15, 1524.05] \\

        &
        & discard $\theta_{PPO}$, retain $\mathcal{M}$
        & -522.02
        & 93.34
        & [-704.96, -339.08] \\

        &
        & discard $\theta_{PPO}$, discard $\mathcal{M}$
        & -522.02
        & 93.34
        & [-704.96, -339.08] \\

        \cmidrule{2-6}

        & \multirow{4}{*}{SAC}
        & retain $\theta_{SAC}$, retain $\mathcal{B}$
        & 395.49
        & 78.18
        & [242.25, 548.73] \\

        &
        & retain $\theta_{SAC}$, discard $\mathcal{B}$
        & 394.23
        & 91.43
        & [215.03, 573.43] \\

        &
        & discard $\theta_{SAC}$, retain $\mathcal{B}$
        & 738.39
        & 22.52
        & [694.25, 782.53] \\

        &
        & discard $\theta_{SAC}$, discard $\mathcal{B}$
        & 740.32
        & 24.64
        & [692.04, 788.61] \\

        \midrule

        \multirow{8}{*}{\makecell{Hip ROM\\Restriction}}
        & \multirow{4}{*}{PPO}
        & retain $\theta_{PPO}$, retain $\mathcal{M}$
        & 3075.30
        & 520.16
        & [2055.79, 4094.81] \\

        &
        & retain $\theta_{PPO}$, discard $\mathcal{M}$
        & 3075.30
        & 520.16
        & [2055.79, 4094.81] \\

        &
        & discard $\theta_{PPO}$, retain $\mathcal{M}$
        & -338.28
        & 40.52
        & [-417.69, -258.86] \\

        &
        & discard $\theta_{PPO}$, discard $\mathcal{M}$
        & -338.28
        & 40.52
        & [-417.69, -258.86] \\

        \cmidrule{2-6}

        & \multirow{4}{*}{SAC}
        & retain $\theta_{SAC}$, retain $\mathcal{B}$
        & 2830.99
        & 472.29
        & [1905.30, 3756.68] \\

        &
        & retain $\theta_{SAC}$, discard $\mathcal{B}$
        & 2830.99
        & 472.29
        & [1905.30, 3756.68] \\

        &
        & discard $\theta_{SAC}$, retain $\mathcal{B}$
        & 741.74
        & 27.27
        & [688.28, 795.19] \\

        &
        & discard $\theta_{SAC}$, discard $\mathcal{B}$
        & 741.74
        & 27.27
        & [688.28, 795.19] \\

        \midrule

        \multirow{8}{*}{\makecell{Broken,\\Severed\\Limb}}
        & \multirow{4}{*}{PPO}
        & retain $\theta_{PPO}$, retain $\mathcal{M}$
        & 811.28
        & 98.02
        & [619.16, 1003.40] \\

        &
        & retain $\theta_{PPO}$, discard $\mathcal{M}$
        & 811.28
        & 98.02
        & [619.16, 1003.40] \\

        &
        & discard $\theta_{PPO}$, retain $\mathcal{M}$
        & -826.01
        & 88.09
        & [-998.67, -653.35] \\

        &
        & discard $\theta_{PPO}$, discard $\mathcal{M}$
        & -826.01
        & 88.09
        & [-998.67, -653.35] \\

        \cmidrule{2-6}

        & \multirow{4}{*}{SAC}
        & retain $\theta_{SAC}$, retain $\mathcal{B}$
        & 44.24
        & 69.61
        & [-92.19, 180.67] \\

        &
        & retain $\theta_{SAC}$, discard $\mathcal{B}$
        & 44.24
        & 69.61
        & [-92.19, 180.67] \\

        &
        & discard $\theta_{SAC}$, retain $\mathcal{B}$
        & 727.24
        & 25.64
        & [676.98, 777.50] \\

        &
        & discard $\theta_{SAC}$, discard $\mathcal{B}$
        & 727.24
        & 25.64
        & [676.98, 777.50] \\

        \midrule

        \multirow{8}{*}{\makecell{Broken,\\Unsevered\\Limb}}
        & \multirow{4}{*}{PPO}
        & retain $\theta_{PPO}$, retain $\mathcal{M}$
        & 6167.02
        & 128.25
        & [5915.66, 6418.39] \\

        &
        & retain $\theta_{PPO}$, discard $\mathcal{M}$
        & 6167.02
        & 128.25
        & [5915.66, 6418.39] \\

        &
        & discard $\theta_{PPO}$, retain $\mathcal{M}$
        & -333.67
        & 53.97
        & [-439.45, -227.89] \\

        &
        & discard $\theta_{PPO}$, discard $\mathcal{M}$
        & -333.67
        & 53.97
        & [-439.45, -227.89] \\

        \cmidrule{2-6}

        & \multirow{4}{*}{SAC}
        & retain $\theta_{SAC}$, retain $\mathcal{B}$
        & 5642.99
        & 172.58
        & [5304.74, 5981.24] \\

        &
        & retain $\theta_{SAC}$, discard $\mathcal{B}$
        & 5642.99
        & 172.58
        & [5304.74, 5981.24] \\

        &
        & discard $\theta_{SAC}$, retain $\mathcal{B}$
        & 711.42
        & 31.10
        & [650.46, 772.38] \\

        &
        & discard $\theta_{SAC}$, discard $\mathcal{B}$
        & 711.42
        & 31.10
        & [650.46, 772.38] \\

        \bottomrule

    \end{tabular}
    \caption{Mean return $\pm$ standard error of the mean (SEM) for PPO and SAC prior to adaptation (0 time steps) in the four Ant-v5 fault environments. Square-bracketed limits give the 95\% confidence interval for the mean, computed with a Student's t-distribution. Transfer settings indicate whether pre-fault model parameters ($\theta$) and experience stores ($\mathcal{M}$ or $\mathcal{B}$) were retained or discarded.}
    \label{tab:ant_no_adaptation}
\end{table}


\begin{table}[!h]
    \centering
    \scriptsize
    \begin{tabular}{>{\centering\arraybackslash}p{2.1cm} >{\centering\arraybackslash}p{1.4cm} >{\centering\arraybackslash}p{3.5cm} ccc}
        \toprule
        \textbf{Fault} & \textbf{Algorithm} & \textbf{Transfer Setting} & \textbf{Mean} & \textbf{SEM} & \textbf{CI [L, U]} \\

        \midrule

        \multirow{8}{*}{\makecell{Ankle ROM\\Restriction}}
        & \multirow{4}{*}{PPO}
        & retain $\theta_{PPO}$, retain $\mathcal{M}$
        & 2139.18
        & 266.93
        & [1616.01, 2662.36] \\

        &
        & retain $\theta_{PPO}$, discard $\mathcal{M}$
        & 1960.71
        & 265.06
        & [1441.20, 2480.23] \\

        &
        & discard $\theta_{PPO}$, retain $\mathcal{M}$
        & 667.57
        & 12.24
        & [643.57, 691.56] \\

        &
        & discard $\theta_{PPO}$, discard $\mathcal{M}$
        & 648.87
        & 12.99
        & [623.40, 674.34] \\

        \cmidrule{2-6}

        & \multirow{4}{*}{SAC}
        & retain $\theta_{SAC}$, retain $\mathcal{B}$
        & 2096.70
        & 180.05
        & [1743.79, 2449.60] \\

        &
        & retain $\theta_{SAC}$, discard $\mathcal{B}$
        & 2569.49
        & 148.15
        & [2279.11, 2859.87] \\

        &
        & discard $\theta_{SAC}$, retain $\mathcal{B}$
        & 153.14
        & 49.05
        & [57.00, 249.28] \\

        &
        & discard $\theta_{SAC}$, discard $\mathcal{B}$
        & 2197.15
        & 77.31
        & [2045.61, 2348.68] \\

        \midrule

        \multirow{8}{*}{\makecell{Hip ROM\\Restriction}}
        & \multirow{4}{*}{PPO}
        & retain $\theta_{PPO}$, retain $\mathcal{M}$
        & 4158.55
        & 267.46
        & [3634.33, 4682.76] \\

        &
        & retain $\theta_{PPO}$, discard $\mathcal{M}$
        & 4463.93
        & 295.02
        & [3885.68, 5042.18] \\

        &
        & discard $\theta_{PPO}$, retain $\mathcal{M}$
        & 551.09
        & 18.95
        & [513.96, 588.23] \\

        &
        & discard $\theta_{PPO}$, discard $\mathcal{M}$
        & 527.44
        & 21.79
        & [484.73, 570.14] \\

        \cmidrule{2-6}

        & \multirow{4}{*}{SAC}
        & retain $\theta_{SAC}$, retain $\mathcal{B}$
        & 4343.12
        & 267.30
        & [3819.21, 4867.04] \\

        &
        & retain $\theta_{SAC}$, discard $\mathcal{B}$
        & 4271.32
        & 180.97
        & [3916.61, 4626.02] \\

        &
        & discard $\theta_{SAC}$, retain $\mathcal{B}$
        & 1671.89
        & 285.79
        & [1111.74, 2232.04] \\

        &
        & discard $\theta_{SAC}$, discard $\mathcal{B}$
        & 2304.59
        & 133.51
        & [2042.91, 2566.26] \\

        \midrule

        \multirow{8}{*}{\makecell{Broken,\\Severed\\Limb}}
        & \multirow{4}{*}{PPO}
        & retain $\theta_{PPO}$, retain $\mathcal{M}$
        & 2374.34
        & 231.94
        & [1919.74, 2828.94] \\

        &
        & retain $\theta_{PPO}$, discard $\mathcal{M}$
        & 2316.24
        & 206.45
        & [1911.61, 2720.87] \\

        &
        & discard $\theta_{PPO}$, retain $\mathcal{M}$
        & 657.21
        & 13.63
        & [630.49, 683.94] \\

        &
        & discard $\theta_{PPO}$, discard $\mathcal{M}$
        & 664.55
        & 10.68
        & [643.62, 685.48] \\

        \cmidrule{2-6}

        & \multirow{4}{*}{SAC}
        & retain $\theta_{SAC}$, retain $\mathcal{B}$
        & 1751.43
        & 177.38
        & [1403.76, 2099.10] \\

        &
        & retain $\theta_{SAC}$, discard $\mathcal{B}$
        & 2312.05
        & 225.16
        & [1870.75, 2753.36] \\

        &
        & discard $\theta_{SAC}$, retain $\mathcal{B}$
        & 846.47
        & 138.28
        & [575.44, 1117.50] \\

        &
        & discard $\theta_{SAC}$, discard $\mathcal{B}$
        & 1754.01
        & 141.59
        & [1476.49, 2031.52] \\

        \midrule

        \multirow{8}{*}{\makecell{Broken,\\Unsevered\\Limb}}
        & \multirow{4}{*}{PPO}
        & retain $\theta_{PPO}$, retain $\mathcal{M}$
        & 5463.51
        & 158.17
        & [5153.50, 5773.52] \\

        &
        & retain $\theta_{PPO}$, discard $\mathcal{M}$
        & 5450.41
        & 177.82
        & [5101.89, 5798.93] \\

        &
        & discard $\theta_{PPO}$, retain $\mathcal{M}$
        & 576.57
        & 19.93
        & [537.50, 615.63] \\

        &
        & discard $\theta_{PPO}$, discard $\mathcal{M}$
        & 561.17
        & 17.72
        & [526.44, 595.89] \\

        \cmidrule{2-6}

        & \multirow{4}{*}{SAC}
        & retain $\theta_{SAC}$, retain $\mathcal{B}$
        & 5841.98
        & 155.97
        & [5536.28, 6147.69] \\

        &
        & retain $\theta_{SAC}$, discard $\mathcal{B}$
        & 4627.35
        & 181.44
        & [4271.73, 4982.97] \\

        &
        & discard $\theta_{SAC}$, retain $\mathcal{B}$
        & 3860.36
        & 266.95
        & [3337.14, 4383.59] \\

        &
        & discard $\theta_{SAC}$, discard $\mathcal{B}$
        & 2235.33
        & 150.06
        & [1941.21, 2529.45] \\

        \bottomrule

    \end{tabular}
    \caption{Mean return $\pm$ standard error of the mean (SEM) for PPO and SAC after partial adaptation (300,000 time steps) in the four Ant-v5 fault environments. Square-bracketed limits give the 95\% confidence interval for the mean, computed with a Student's t-distribution. Transfer settings indicate whether pre-fault model parameters ($\theta$) and experience stores ($\mathcal{M}$ or $\mathcal{B}$) were retained or discarded.}
    \label{tab:ant_early_adaptation}
\end{table}


\begin{table}[!h]
    \centering
    \scriptsize
    \begin{tabular}{>{\centering\arraybackslash}p{2.1cm} >{\centering\arraybackslash}p{1.4cm} >{\centering\arraybackslash}p{3.5cm} ccc}
        \toprule
        \textbf{Fault} & \textbf{Algorithm} & \textbf{Transfer Setting} & \textbf{Mean} & \textbf{SEM} & \textbf{CI [L, U]} \\

        \midrule

        \multirow{8}{*}{\makecell{Ankle ROM\\Restriction}}
        & \multirow{4}{*}{PPO}
        & retain $\theta_{PPO}$, retain $\mathcal{M}$
        & 5578.17
        & 75.67
        & [5429.86, 5726.48] \\

        &
        & retain $\theta_{PPO}$, discard $\mathcal{M}$
        & 5364.29
        & 117.26
        & [5134.46, 5594.12] \\

        &
        & discard $\theta_{PPO}$, retain $\mathcal{M}$
        & 5157.58
        & 149.86
        & [4863.85, 5451.32] \\

        &
        & discard $\theta_{PPO}$, discard $\mathcal{M}$
        & 5248.90
        & 119.52
        & [5014.65, 5483.15] \\

        \cmidrule{2-6}

        & \multirow{4}{*}{SAC}
        & retain $\theta_{SAC}$, retain $\mathcal{B}$
        & 5105.93
        & 103.59
        & [4902.89, 5308.98] \\

        &
        & retain $\theta_{SAC}$, discard $\mathcal{B}$
        & 5083.92
        & 118.23
        & [4852.20, 5315.65] \\

        &
        & discard $\theta_{SAC}$, retain $\mathcal{B}$
        & 2488.81
        & 486.90
        & [1534.49, 3443.13] \\

        &
        & discard $\theta_{SAC}$, discard $\mathcal{B}$
        & 4468.83
        & 119.29
        & [4235.02, 4702.65] \\

        \midrule

        \multirow{8}{*}{\makecell{Hip ROM\\Restriction}}
        & \multirow{4}{*}{PPO}
        & retain $\theta_{PPO}$, retain $\mathcal{M}$
        & 5396.93
        & 259.20
        & [4888.90, 5904.96] \\

        &
        & retain $\theta_{PPO}$, discard $\mathcal{M}$
        & 5385.84
        & 248.95
        & [4897.89, 5873.79] \\

        &
        & discard $\theta_{PPO}$, retain $\mathcal{M}$
        & 5965.29
        & 238.63
        & [5497.57, 6433.01] \\

        &
        & discard $\theta_{PPO}$, discard $\mathcal{M}$
        & 5847.63
        & 243.34
        & [5370.69, 6324.57] \\

        \cmidrule{2-6}

        & \multirow{4}{*}{SAC}
        & retain $\theta_{SAC}$, retain $\mathcal{B}$
        & 4146.15
        & 516.62
        & [3133.58, 5158.72] \\

        &
        & retain $\theta_{SAC}$, discard $\mathcal{B}$
        & 5679.06
        & 161.71
        & [5362.11, 5996.01] \\

        &
        & discard $\theta_{SAC}$, retain $\mathcal{B}$
        & 1790.97
        & 716.24
        & [387.13, 3194.80] \\

        &
        & discard $\theta_{SAC}$, discard $\mathcal{B}$
        & 5475.11
        & 166.62
        & [5148.54, 5801.68] \\

        \midrule

        \multirow{8}{*}{\makecell{Broken,\\Severed\\Limb}}
        & \multirow{4}{*}{PPO}
        & retain $\theta_{PPO}$, retain $\mathcal{M}$
        & 5178.76
        & 309.41
        & [4572.32, 5785.21] \\

        &
        & retain $\theta_{PPO}$, discard $\mathcal{M}$
        & 5919.98
        & 115.85
        & [5692.92, 6147.04] \\

        &
        & discard $\theta_{PPO}$, retain $\mathcal{M}$
        & 5764.29
        & 162.25
        & [5446.28, 6082.31] \\

        &
        & discard $\theta_{PPO}$, discard $\mathcal{M}$
        & 5487.20
        & 147.78
        & [5197.55, 5776.84] \\

        \cmidrule{2-6}

        & \multirow{4}{*}{SAC}
        & retain $\theta_{SAC}$, retain $\mathcal{B}$
        & 5514.13
        & 103.45
        & [5311.37, 5716.90] \\

        &
        & retain $\theta_{SAC}$, discard $\mathcal{B}$
        & 5393.69
        & 129.19
        & [5140.48, 5646.91] \\

        &
        & discard $\theta_{SAC}$, retain $\mathcal{B}$
        & 5416.13
        & 137.30
        & [5147.02, 5685.24] \\

        &
        & discard $\theta_{SAC}$, discard $\mathcal{B}$
        & 4721.59
        & 148.84
        & [4429.86, 5013.31] \\

        \midrule

        \multirow{8}{*}{\makecell{Broken,\\Unsevered\\Limb}}
        & \multirow{4}{*}{PPO}
        & retain $\theta_{PPO}$, retain $\mathcal{M}$
        & 5964.47
        & 211.19
        & [5550.52, 6378.41] \\

        &
        & retain $\theta_{PPO}$, discard $\mathcal{M}$
        & 5847.15
        & 217.86
        & [5420.15, 6274.15] \\

        &
        & discard $\theta_{PPO}$, retain $\mathcal{M}$
        & 5803.32
        & 310.24
        & [5195.25, 6411.39] \\

        &
        & discard $\theta_{PPO}$, discard $\mathcal{M}$
        & 6103.00
        & 214.55
        & [5682.48, 6523.52] \\

        \cmidrule{2-6}

        & \multirow{4}{*}{SAC}
        & retain $\theta_{SAC}$, retain $\mathcal{B}$
        & 6390.86
        & 121.99
        & [6151.75, 6629.97] \\

        &
        & retain $\theta_{SAC}$, discard $\mathcal{B}$
        & 6212.33
        & 124.05
        & [5969.19, 6455.47] \\

        &
        & discard $\theta_{SAC}$, retain $\mathcal{B}$
        & 6088.35
        & 103.37
        & [5885.74, 6290.96] \\

        &
        & discard $\theta_{SAC}$, discard $\mathcal{B}$
        & 5886.22
        & 104.16
        & [5682.06, 6090.38] \\

        \bottomrule

    \end{tabular}
    \caption{Mean return $\pm$ standard error of the mean (SEM) for PPO and SAC after full adaptation (30,000,000 and 3,000,000 time steps, respectively) in the four Ant-v5 fault environments. Square-bracketed limits give the 95\% confidence interval for the mean, computed with a Student's t-distribution. Transfer settings indicate whether pre-fault model parameters ($\theta$) and experience stores ($\mathcal{M}$ or $\mathcal{B}$) were retained or discarded.}
    \label{tab:ant_full_adaptation}
\end{table}


\begin{table}[!h]
    \centering
    \scriptsize
    \begin{tabular}{>{\centering\arraybackslash}p{2.1cm} >{\centering\arraybackslash}p{1.4cm} >{\centering\arraybackslash}p{3.5cm} ccc}
        \toprule
        \textbf{Fault} & \textbf{Algorithm} & \textbf{Transfer Setting} & \textbf{Mean} & \textbf{SEM} & \textbf{CI [L, U]} \\

        \midrule

        \multirow{8}{*}{\makecell{Frozen\\Shoulder\\Lift Position\\Sensor}}
        & \multirow{4}{*}{PPO}
        & retain $\theta_{PPO}$, retain $\mathcal{M}$
        & -47.75
        & 0.23
        & [-48.19, -47.30] \\

        &
        & retain $\theta_{PPO}$, discard $\mathcal{M}$
        & -47.75
        & 0.23
        & [-48.19, -47.30] \\

        &
        & discard $\theta_{PPO}$, retain $\mathcal{M}$
        & -47.65
        & 0.82
        & [-49.26, -46.03] \\

        &
        & discard $\theta_{PPO}$, discard $\mathcal{M}$
        & -47.65
        & 0.82
        & [-49.26, -46.03] \\

        \cmidrule{2-6}

        & \multirow{4}{*}{SAC}
        & retain $\theta_{SAC}$, retain $\mathcal{B}$
        & -47.88
        & 0.24
        & [-48.34, -47.42] \\

        &
        & retain $\theta_{SAC}$, discard $\mathcal{B}$
        & -47.88
        & 0.24
        & [-48.34, -47.42] \\

        &
        & discard $\theta_{SAC}$, retain $\mathcal{B}$
        & -50.25
        & 0.37
        & [-50.98, -49.53] \\

        &
        & discard $\theta_{SAC}$, discard $\mathcal{B}$
        & -50.25
        & 0.37
        & [-50.98, -49.53] \\

        \midrule

        \multirow{8}{*}{\makecell{Slippery\\Elbow Flex\\Joint}}
        & \multirow{4}{*}{PPO}
        & retain $\theta_{PPO}$, retain $\mathcal{M}$
        & -2.04
        & 0.08
        & [-2.21, -1.88] \\

        &
        & retain $\theta_{PPO}$, discard $\mathcal{M}$
        & -2.04
        & 0.08
        & [-2.21, -1.88] \\

        &
        & discard $\theta_{PPO}$, retain $\mathcal{M}$
        & -29.20
        & 1.51
        & [-32.15, -26.25] \\

        &
        & discard $\theta_{PPO}$, discard $\mathcal{M}$
        & -29.20
        & 1.51
        & [-32.15, -26.25] \\

        \cmidrule{2-6}

        & \multirow{4}{*}{SAC}
        & retain $\theta_{SAC}$, retain $\mathcal{B}$
        & -2.99
        & 0.12
        & [-3.23, -2.75] \\

        &
        & retain $\theta_{SAC}$, discard $\mathcal{B}$
        & -2.99
        & 0.12
        & [-3.23, -2.75] \\

        &
        & discard $\theta_{SAC}$, retain $\mathcal{B}$
        & -24.29
        & 0.41
        & [-25.09, -23.49] \\

        &
        & discard $\theta_{SAC}$, discard $\mathcal{B}$
        & -24.29
        & 0.41
        & [-25.09, -23.49] \\

        \bottomrule

    \end{tabular}
    \caption{Mean return $\pm$ standard error of the mean (SEM) for PPO and SAC prior to adaptation (0 time steps) in the two FetchReachDense-v3 fault environments. Square-bracketed limits give the 95\% confidence interval for the mean, computed with a Student's t-distribution. Transfer settings indicate whether pre-fault model parameters ($\theta$) and experience stores ($\mathcal{M}$ or $\mathcal{B}$) were retained or discarded.}
    \label{tab:fetchreach_no_adaptation}
\end{table}


\begin{table}[!h]
    \centering
    \scriptsize
    \begin{tabular}{>{\centering\arraybackslash}p{2.1cm} >{\centering\arraybackslash}p{1.4cm} >{\centering\arraybackslash}p{3.5cm} ccc}
        \toprule
        \textbf{Fault} & \textbf{Algorithm} & \textbf{Transfer Setting} & \textbf{Mean} & \textbf{SEM} & \textbf{CI [L, U]} \\

        \midrule

        \multirow{8}{*}{\makecell{Frozen\\Shoulder\\Lift Position\\Sensor}}
        & \multirow{4}{*}{PPO}
        & retain $\theta_{PPO}$, retain $\mathcal{M}$
        & -47.50
        & 0.23
        & [-47.95, -47.05] \\

        &
        & retain $\theta_{PPO}$, discard $\mathcal{M}$
        & -47.50
        & 0.22
        & [-47.93, -47.06] \\

        &
        & discard $\theta_{PPO}$, retain $\mathcal{M}$
        & -23.58
        & 1.36
        & [-26.25, -20.91] \\

        &
        & discard $\theta_{PPO}$, discard $\mathcal{M}$
        & -24.65
        & 1.52
        & [-27.63, -21.68] \\

        \cmidrule{2-6}

        & \multirow{4}{*}{SAC}
        & retain $\theta_{SAC}$, retain $\mathcal{B}$
        & -25.37
        & 1.11
        & [-27.55, -23.19] \\

        &
        & retain $\theta_{SAC}$, discard $\mathcal{B}$
        & -23.88
        & 0.87
        & [-25.59, -22.17] \\

        &
        & discard $\theta_{SAC}$, retain $\mathcal{B}$
        & -23.36
        & 0.28
        & [-23.90, -22.81] \\

        &
        & discard $\theta_{SAC}$, discard $\mathcal{B}$
        & -22.98
        & 0.27
        & [-23.51, -22.46] \\

        \midrule

        \multirow{8}{*}{\makecell{Slippery\\Elbow Flex\\Joint}}
        & \multirow{4}{*}{PPO}
        & retain $\theta_{PPO}$, retain $\mathcal{M}$
        & -1.72
        & 0.45
        & [-2.60, -0.84] \\

        &
        & retain $\theta_{PPO}$, discard $\mathcal{M}$
        & -1.14
        & 0.36
        & [-1.84, -0.44] \\

        &
        & discard $\theta_{PPO}$, retain $\mathcal{M}$
        & -0.98
        & 0.04
        & [-1.05, -0.91] \\

        &
        & discard $\theta_{PPO}$, discard $\mathcal{M}$
        & -1.01
        & 0.04
        & [-1.09, -0.93] \\

        \cmidrule{2-6}

        & \multirow{4}{*}{SAC}
        & retain $\theta_{SAC}$, retain $\mathcal{B}$
        & -1.04
        & 0.04
        & [-1.12, -0.96] \\

        &
        & retain $\theta_{SAC}$, discard $\mathcal{B}$
        & -0.97
        & 0.07
        & [-1.10, -0.85] \\

        &
        & discard $\theta_{SAC}$, retain $\mathcal{B}$
        & -1.48
        & 0.07
        & [-1.62, -1.34] \\

        &
        & discard $\theta_{SAC}$, discard $\mathcal{B}$
        & -1.39
        & 0.06
        & [-1.51, -1.28] \\

        \bottomrule

    \end{tabular}
    \caption{Mean return $\pm$ standard error of the mean (SEM) for PPO and SAC after partial adaptation (30,000 time steps) in the two FetchReachDense-v3 fault environments. Square-bracketed limits give the 95\% confidence interval for the mean, computed with a Student's t-distribution. Transfer settings indicate whether pre-fault model parameters ($\theta$) and experience stores ($\mathcal{M}$ or $\mathcal{B}$) were retained or discarded.}
    \label{tab:fetchreach_early_adaptation}
\end{table}


\begin{table}[!h]
    \centering
    \scriptsize
    \begin{tabular}{>{\centering\arraybackslash}p{2.1cm} >{\centering\arraybackslash}p{1.4cm} >{\centering\arraybackslash}p{3.5cm} ccc}
        \toprule
        \textbf{Fault} & \textbf{Algorithm} & \textbf{Transfer Setting} & \textbf{Mean} & \textbf{SEM} & \textbf{CI [L, U]} \\

        \midrule

        \multirow{8}{*}{\makecell{Frozen\\Shoulder\\Lift Position\\Sensor}}
        & \multirow{4}{*}{PPO}
        & retain $\theta_{PPO}$, retain $\mathcal{M}$
        & -47.32
        & 0.24
        & [-47.79, -46.85] \\

        &
        & retain $\theta_{PPO}$, discard $\mathcal{M}$
        & -47.32
        & 0.32
        & [-47.94, -46.70] \\

        &
        & discard $\theta_{PPO}$, retain $\mathcal{M}$
        & -19.96
        & 1.07
        & [-22.05, -17.87] \\

        &
        & discard $\theta_{PPO}$, discard $\mathcal{M}$
        & -20.88
        & 1.15
        & [-23.13, -18.63] \\

        \cmidrule{2-6}

        & \multirow{4}{*}{SAC}
        & retain $\theta_{SAC}$, retain $\mathcal{B}$
        & -22.65
        & 0.37
        & [-23.38, -21.91] \\

        &
        & retain $\theta_{SAC}$, discard $\mathcal{B}$
        & -22.98
        & 0.58
        & [-24.13, -21.84] \\

        &
        & discard $\theta_{SAC}$, retain $\mathcal{B}$
        & -23.08
        & 0.28
        & [-23.62, -22.54] \\

        &
        & discard $\theta_{SAC}$, discard $\mathcal{B}$
        & -22.72
        & 0.27
        & [-23.24, -22.20] \\

        \midrule

        \multirow{8}{*}{\makecell{Slippery\\Elbow Flex\\Joint}}
        & \multirow{4}{*}{PPO}
        & retain $\theta_{PPO}$, retain $\mathcal{M}$
        & -0.78
        & 0.11
        & [-0.99, -0.58] \\

        &
        & retain $\theta_{PPO}$, discard $\mathcal{M}$
        & -0.71
        & 0.13
        & [-0.98, -0.45] \\

        &
        & discard $\theta_{PPO}$, retain $\mathcal{M}$
        & -0.67
        & 0.02
        & [-0.72, -0.63] \\

        &
        & discard $\theta_{PPO}$, discard $\mathcal{M}$
        & -0.66
        & 0.02
        & [-0.70, -0.62] \\

        \cmidrule{2-6}

        & \multirow{4}{*}{SAC}
        & retain $\theta_{SAC}$, retain $\mathcal{B}$
        & -0.85
        & 0.04
        & [-0.92, -0.78] \\

        &
        & retain $\theta_{SAC}$, discard $\mathcal{B}$
        & -0.76
        & 0.04
        & [-0.84, -0.69] \\

        &
        & discard $\theta_{SAC}$, retain $\mathcal{B}$
        & -1.27
        & 0.06
        & [-1.39, -1.15] \\

        &
        & discard $\theta_{SAC}$, discard $\mathcal{B}$
        & -1.25
        & 0.06
        & [-1.37, -1.12] \\

        \bottomrule

    \end{tabular}
    \caption{Mean return $\pm$ standard error of the mean (SEM) for PPO and SAC after full adaptation (50,000 time steps) in the two FetchReachDense-v3 fault environments. Square-bracketed limits give the 95\% confidence interval for the mean, computed with a Student's t-distribution. Transfer settings indicate whether pre-fault model parameters ($\theta$) and experience stores ($\mathcal{M}$ or $\mathcal{B}$) were retained or discarded.}
    \label{tab:fetchreach_full_adaptation}
\end{table}

\clearpage
\bibliographystyle{elsarticle-harv} 
\bibliography{elsarticle}

\end{document}